\pdfoutput=1

\documentclass[8pt]{article}

\usepackage{algorithm}
\usepackage{algorithmic}
\newcommand{\algorithmicbreak}{\textbf{break}}
\newcommand{\BREAK}{\STATE \algorithmicbreak}

\usepackage{subfloat}
\usepackage{subcaption}

\usepackage{xcolor}
\usepackage{xspace}
\usepackage{listings}

\usepackage{subfloat}
\usepackage{url}
\usepackage[export]{adjustbox}
\usepackage{enumitem}
\usepackage{tabularx}
\usepackage{graphicx}
\usepackage{indentfirst}
\usepackage{balance}
\usepackage{booktabs} 

\usepackage{amsmath,amsfonts,amsthm}
\usepackage{mathtools}
\usepackage{commath}
\usepackage{multirow}
\usepackage[numbers]{natbib}
\usepackage{hyperref}

\newcommand{\sys}{AutoFreeze}

\newcommand{\ie}{{\it i.e.}, }
\title{ Title}
\author{ Yuhan Liu, Saurabh Agarwal, Shivaram Venkataraman\\ University of Wisconsin-Madison }
\date{}

\title{AutoFreeze: Automatically Freezing Model Blocks to Accelerate Fine-tuning}

\begin{document}
\maketitle
\begin{abstract}
With the rapid adoption of machine learning (ML), a number of domains now use the approach of fine tuning models which were pre-trained on a large corpus of data. However, our experiments show that even fine-tuning on models like BERT can take many hours even when using modern accelerators like GPUs. While prior work proposes limiting the number of layers that are fine-tuned, e.g., freezing all layers but the last layer, we find that such static approaches lead to reduced accuracy. We propose, \sys{}, a system that uses an adaptive approach to choose which layers are trained and show how this can accelerate model fine-tuning while preserving accuracy. We also develop mechanisms to enable efficient caching of intermediate activations which can reduce the forward computation time when performing fine-tuning. We extend \sys{} to perform distributed fine-tuning and design two execution modes that minimize cost and running time respectively. Our evaluation on ten NLP tasks shows that \sys{}, with caching enabled, can improve fine-tuning on a single GPU by up to 2.55$\times$. On a 64 GPU cluster, for fine-tuning on the AG's news dataset, \sys{} is able to achieve up to 4.38$\times$ speedup when optimizing for end-to-end training time and 5.03$\times$ reduction in total cost when optimizing for efficiency, without affecting model accuracy.

\end{abstract}

\maketitle

\section{Introduction}

Deep Learning based models have been shown to provide extremely competitive performance across a wide range of tasks.
However, building deep learning based models for new tasks requires a large amount of data and compute~\cite{roh2019survey}. To circumvent these requirements practitioners bootstrap new tasks from existing models. To bootstrap from existing models, practitioners typically use transfer learning or fine tuning~\cite{zhuang2020comprehensive}. In case of transfer-learning, the features from a large pre-trained model are directly used on a new task and \emph{only the last one or few} layers are trained to develop a specialized model. Closely related is the practice of fine tuning where weights from a pre-trained model are used to initialise a new task; following initialization \emph{all layers} of the model are trained until convergence. 

In the case of language models~\cite{howard2018universal},
fine-tuning has become a standard part of the two stage training process. In first stage, which is pre-training, complex models (e.g., BERT~\cite{devlin2018bert}) are trained with a large corpus of unlabeled data. In the second stage, which is fine tuning, the pre-trained model is fine-tuned for a specific task such as sentiment analysis~\cite{maas-etal-2011-learning} or topic classification~\cite{zhang2016characterlevel} etc.

While fine-tuning is cheaper than pre-training a model, it is important to note that pre-training is usually performed very infrequently compared to fine tuning. For example in the case of language models, a vast majority of practitioners take a pre-trained BERT model and perform fine tuning on their data sets~\cite{maas-etal-2011-learning,rajpurkar2018know,zellers2018swag,hermann2015teaching,zhang2016characterlevel,sun2019fine} or for new tasks~\cite{sun2019fine,wolf-etal-2020-transformers,zhu2020incorporating,liu2019text}.

Prior approaches in developing tools for improving ML model training exploit ways to reduce training time by maximizing throughput and better utilizing resources. A number of works including PyTorch DistributedDataParallel (DDP)~\cite{li2020pytorch} and BytePS~\cite{258953} provide support for data-parallel distributed training and improve utilization of heterogeneous resources such as GPUs, CPUs, and network bandwidth. More recent works like Cerebro~\cite{cerebro} and PipeDream~\cite{pipedream} utilize hybrid parallelism to improve model selection throughput and minimize synchronization bottlenecks. 
 
Even when using Pytorch DDP~\citep{li2020pytorch}, we find that fine-tuning BERT for the sentiment classification task with the Yelp dataset~\cite{zhang2016characterlevel} can take around 27 hours \footnote{Measured on CloudLab} in a four P100 GPU cluster. 

The primary reason fine tuning is so computationally expensive is the large size of the pre-trained model.  Each layer of the pre-trained model is computationally intensive (Section~\ref{sec:bert_decsription}) and requires a significant amount of time for forward and backward pass.  Further, existing training systems are generic and hence oblivious to the convergence properties of the fine-tuning workloads; given that fine tuning already starts from a pre-trained model, training systems need to be aware of the opportunities that arise from rapid convergence to achieve high throughput.

\begin{figure}[!t]
\centering
\includegraphics[width=.7\linewidth]{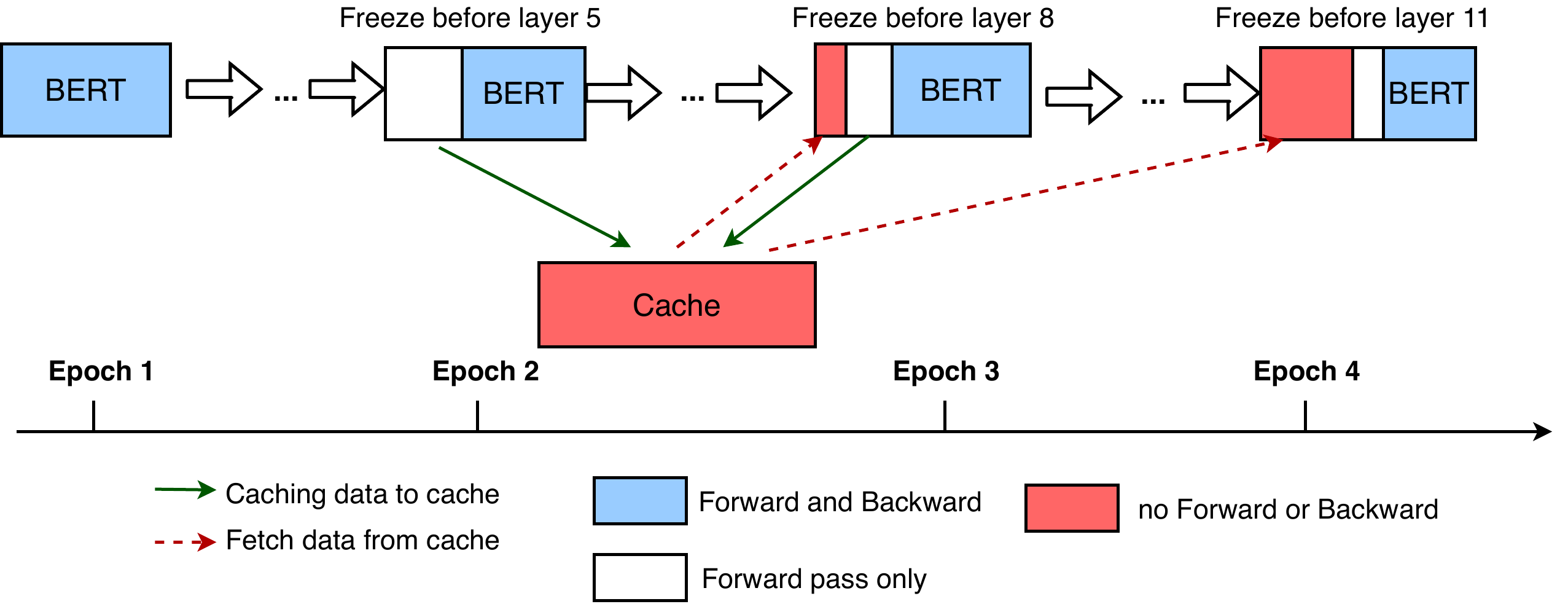}
  \caption{We present a high level design of \sys{}. During fine-tuning, \sys{} adaptively determines layers which can be frozen. Once layers are frozen, the backward computation for those layers can be avoided. At later epochs, intermediate outputs are also cached leading to further gains. }
  \label{fig:timeline}
\end{figure}

A natural approach to improve fine-tuning performance is to limit the number of layers of the model that are updated, thus making it similar to transfer learning. 
For example, if we consider BERT$_{BASE}$ which has 12 encoding blocks, prior approach by ~\citet{lee2019would} trains a fixed number of blocks (e.g., the last 3 blocks) and \emph{freezes} the weights for the remaining blocks. However, this approach can affect the final model accuracy. For example with the MRPC dataset~\cite{dolan-brockett-2005-automatically}  this approach can reduce the time for an epoch by 2$\times$, but we find that the accuracy of the fine-tuned model suffers, dropping from 87\% to 76.5\%.

Another approach used in prior work by ~\citet{chen2020lottery} is to apply the ``Lottery Ticket Hypothesis'' to identify matching subnetworks in pre-trained BERT models to enforce sparsity in models trained for different downstream tasks. While this approach retains accuracy and can lead to sparser models, it does not lead to improvements in training speed without dedicated hardware or libraries~\cite{liu2018rethinking}.

In this paper, we propose a novel approach where the number of model blocks that are updated and resources used during fine-tuning are adaptively chosen during the fine-tuning process. Our work is inspired by recent work of ~\citet{Raghu2017SVCCASV} who developed SVCCA, a new metric that captures how different layers of model change over the course of training. The SVCCA score for a layer, as proposed by ~\citet{Raghu2017SVCCASV}, is computed by comparing the intermediate model weights with the final weights and can thus be used for post-hoc analysis.  Applying that approach to model fine-tuning we observe that the initial layers of the model converge rapidly and thus we can \emph{freeze} such layers. Freezing the initial layers means that the backward pass for those layers can be skipped, thereby reducing the computation and communication required. While SVCCA scores show the promise of freezing layers early, we still need an online algorithm that can decide which layers should be frozen and when. We develop a gradient-norm based test that ranks layers by their rate of change and based on it, selects the slowest changing layers for freezing. We show that our method is effective at detecting when layers should be frozen without affecting accuracy across multiple datasets.

Beyond just reducing the time for backward pass freezing  early layers of the model provided several advantages. One major advantage is that freezing layers will also reduce the amount of communication required for distributed training. For example freezing layers of BERT can save around 27MB per layer frozen~\footnote{Size of one BERT encoder layer.}, thus reducing the synchronization required. 

\begin{figure}[!t]
    \centering
    \includegraphics[width=0.4\textwidth]{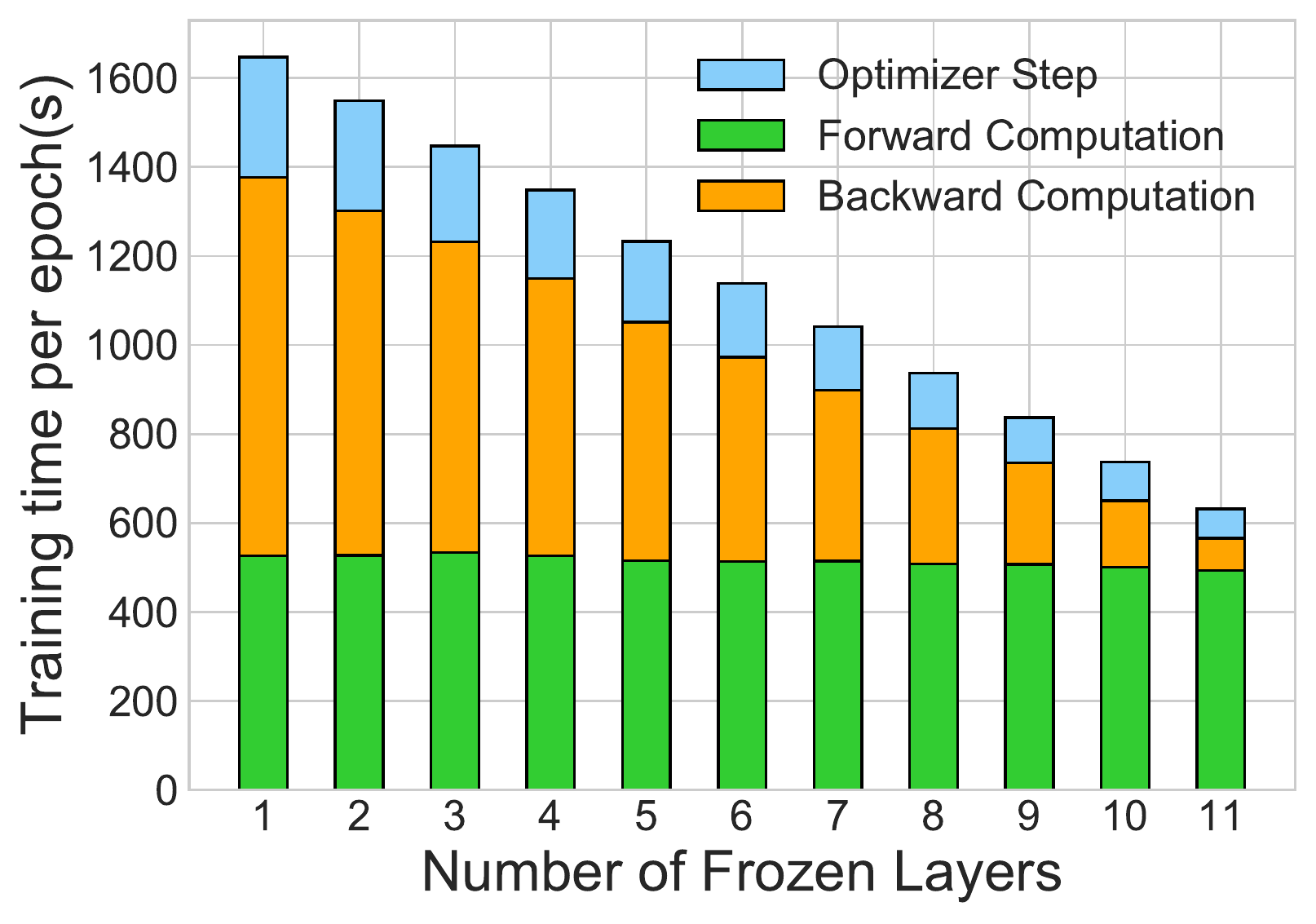}
     \vspace{-0.15in}
    \caption{\small{\textbf{Potential Benefits of Freezing:} For IMDb dataset we show the potential savings in time when performing freezing. Time for forward pass is constant since we need to perform a full forward pass before calculating gradients. On the other hand timing for backward pass reduces as we increase number of layers frozen.}}
    \label{fig:freeze_potential_benefits_imdb}
\end{figure}

Another potential benefit of freezing is that it can reduce the time for forward pass. Since the frozen layers don't change during subsequent training iterations, the output for a given data point will be constant for the frozen layers. Applying this insight we can cache the output of the forward pass up to the layer that has been frozen. Once the same data point is selected to be used again for training, we can load the pre-computed intermediate values from our cache and continue training.
Figure~\ref{fig:freeze_potential_benefits_imdb} shows the potential saving our freezing module can provide by caching the intermediate outputs.

We design \sys{}, a system for automatically freezing layers to accelerate fine-tuning. 
Our system consists of two main modules on each GPU: a \emph{freezing module} that has a pluggable decision engine that can make decisions on which layers should be frozen based on the aggregated gradients as training progresses. The freezing module also includes a decision engine that dynamically picks the amount of resources to use during fine-tuning. We design two modes that the user can choose from: Performance Packing for minimizing training time or Efficiency Packing for minimizing cost. We also design a \emph{storage manager} module to implement the caching functions described above and the storage manager handles a number of common concerns in caching, including selecting the appropriate backend (CPU memory / SSD etc.) and deciding when to evict data from the cache.

We evaluate \sys{} using a wide range of fine-tuning tasks.
including topic classification on the AG's News dataset~\cite{zhang2016characterlevel} and Sogou News dataset\cite{sun2019fine}, sentiment analysis on Yelp Full dataset\cite{zhang2016characterlevel} and IMDb dataset\cite{maas-etal-2011-learning}, question answering on SQuAD2.0 dataset\cite{rajpurkar2018know}, multiple choice task on SWAG dataset\cite{zellers2018swag}, and text summarization on CNN/DailyMail dataset \cite{hermann2015teaching}. 
We find that for a single machine fine tuning \sys{} can improve training time by up to 2.55$\times$ while affecting accuracy by less than 0.1\%. We also show that \sys{} is especially effective for large datasets like Yelp where freezing layers reduces fine-tuning time from 52.5 hours to 27 hours and caching further reduces this to 24.6 hours. In distributed fine-tuning \sys{} can significantly improve training time and efficiency, reducing the end-to-end training time by $4.4\times$ with the performance packing mode and reducing total cost by 5.03$\times$ when using the efficiency packing mode for fine tuning on the AG's News dataset in a 64 GPU cluster. 

\section{Motivation and Background}
\label{sec:motivation}

In this section we first provide background on model fine-tuning and transfer learning and detail why fine-tuning is expensive.
Following that we motivate how freezing or limiting the number of layers of a model trained can lead to significant savings. Finally we show
how static schemes that freeze a constant number of layers are ineffective.

\begin{figure*}[!t]
    \centering
    \includegraphics[width=0.8\textwidth]{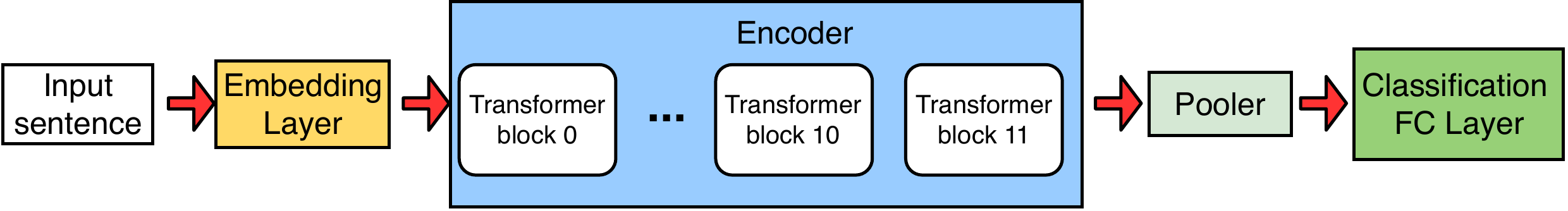}
    \caption{\small{\textbf{BERT model architecture}}}
    \label{fig:bert_arch}
\end{figure*}

\subsection{Model Fine-tuning}
Transfer learning and fine tuning of large pre-trained models has enabled easy use of deep models for new tasks and new datasets~\cite{sun2019fine,wolf-etal-2020-transformers,zhu2020incorporating,liu2019text}. In transfer learning we use the features from a large pre-trained model and train only the last few layers to specialize on a new task, while in case of fine tuning the whole pre-trained model is trained on a new task. Both transfer learning and fine tuning have several advantages over training models from initialization including i) enabling use of deep models when training data is scarce, ii) transferring common features among related tasks iii) significantly faster convergence that also reduces the computation time. 

Further, periodic fine-tuning is a necessity for models deployed in production, since in real world the data distribution changes quite frequently~\cite{suprem2020odin}. When data distribution drifts further from the training distribution, 
models typically observe increased error rates from out of distribution(OOD) data points~\cite{liang2020enhancing, madras2019detecting}.
Periodically fine-tuning models with newly collected data is one of the most common methods to keep error rates low~\cite{suprem2020odin,liang2020enhancing, zisselman2020deep}, making fine-tuning an extremely important workload for ML deployments in production.

Although feature based transfer learning is very popular in computer vision tasks~\cite{sharif2014cnn, huang2017zero,bansal2018zero, ying2018transfer, chen2018lstd}, recent works~\cite{howard2018universal, peters2018deep} show that language models enjoy significantly better performance when using fine tuning. 
However, even when performing fine tuning, large models like BERT~\cite{devlin2018bert} require a significant amount of time. For example, fine tuning BERT on the relatively small IMDB dataset~\cite{maas-etal-2011-learning}, containing 25K points, takes around 3 hours on a single P100 GPU. On larger datasets like Yelp (Table~\ref{tab:dataset_stat}) we see that fine-tuning can take more than two days on single P100 GPU.  Distributed training provides limited benefits given the size of the models~\cite{reddi2020mlperf}, with four P100 GPUs only reducing the training time to 27 hours.
Even on the latest A100 GPU, fine-tuning $BERT_{LARGE}$ on the Yelp dataset can take around two hours. Thus, the exorbitant cost of fine tuning becomes a limiting factor for data scientists in developing new models. 


\subsection{BERT Model Architecture and Timing}
\label{sec:bert_decsription}
To get a deeper understanding of the performance of fine-tuning on BERT, we first discuss the model architecture and then present a breakdown of fine-tuning time. We primarily focus on $BERT_{BASE}$ which is depicted in Figure~\ref{fig:bert_arch}. $BERT_{BASE}$ has 12 encoder layers, which are also called  Transformer blocks. Each Transformer block is identical and comprises of a self-attention layer with 12 attention heads and a fully connected layer of size 768. In this work we refer to layer and transformer block interchangeably, therefore by freezing a layer we mean freezing the entire transformer block.

As discussed in prior work~\cite{sun2019fine}, fine-tuning for text classification tasks typically only requires around 4 epochs to achieve state-of-the-art accuracy. But the time taken per epoch is high because of the following reasons:

\noindent\textbf{Memory constraints:} With the BERT model being large (model weights around 420 MB) and the intermediate activations of each layer also being large, the batch size that can be used for fine tuning is limited by GPU memory. With an NVIDIA P100 GPU, having 12GB of memory, we observe that we are limited to a batch size of 6 \footnote{Measured using PyTorch 1.0.1}. Even on newer hardware like A100 GPU having 40GB of memory, the maximum batch size that can be supported is limited to 32. 

\noindent\textbf{Computation needs:} The transformer blocks discussed above are also compute intensive and we observe that the gradient calculation time, especially in the backward pass can be significant. For example, when fine-tuning with the IMDb dataset, we see that doing one iteration takes around 435ms of which more than 50\% is taken by the backward pass. 

\noindent\textbf{Communication intensive:} The large model size also imposes significant communication overheads when performing distributed data-parallel training. For example, when scaling from one p3.2xlarge to 64 p3.2xlarge GPUs on Amazon EC2, while having a fixed batch size per GPU, leads to an inflation in per-iteration time from 0.29 seconds to 8.9 seconds.

\subsection{Existing Approaches}
\label{sec:static}
One direct approach to reduce the computational cost is to only fine-tune a subset of the layers~\cite{lee2019would}. For example, as shown in Figure~\ref{fig:freeze_potential_benefits_imdb}, only updating the last $k$ layers of the BERT leads  to an almost linear decrease in time per iteration. This hints that avoiding gradient computation for certain layers, i.e., \emph{freezing}, can significantly reduce training time. 

We note that in order to realize the gain from freezing, the layers should be frozen in order, \textit{e.g.} freezing layers 3 and 4 before freezing earlier layers 1 and 2 is not going to provide any speedup. This is due to the usage of automatic differentiation~\cite{autograd} in popular deep learning libraries for backpropagation. Automatic differentiation uses the gradient of later layers to calculate the gradients of earlier layers, \textit{i.e.}, to calculate gradients of Layer 1 automatic differentiation requires gradient of Layer 2 to be calculated. However, this leads to the question of which layers should be frozen and how does freezing of layers effect the model accuracy. 

\begin{figure*}[t!]
 	\begin{subfigure}[t]{.3\linewidth}%
 	\center
 	\includegraphics[width=\linewidth]{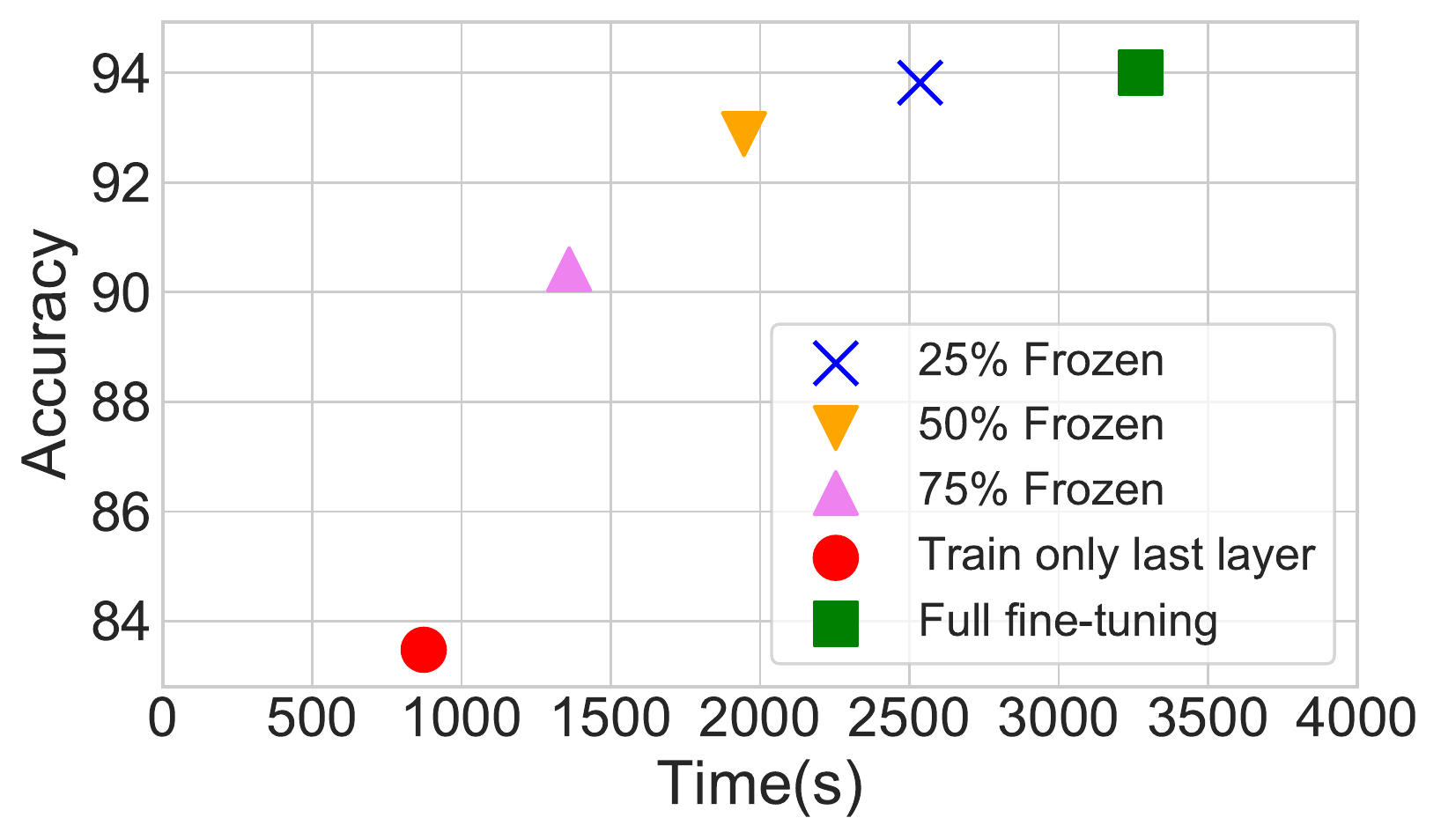}
 	\caption{CINIC}
 	\label{fig:cinic_transfer}
 	\end{subfigure}
 	\begin{subfigure}[t]{.3\linewidth}%
 	\center
 	\includegraphics[width=\linewidth]{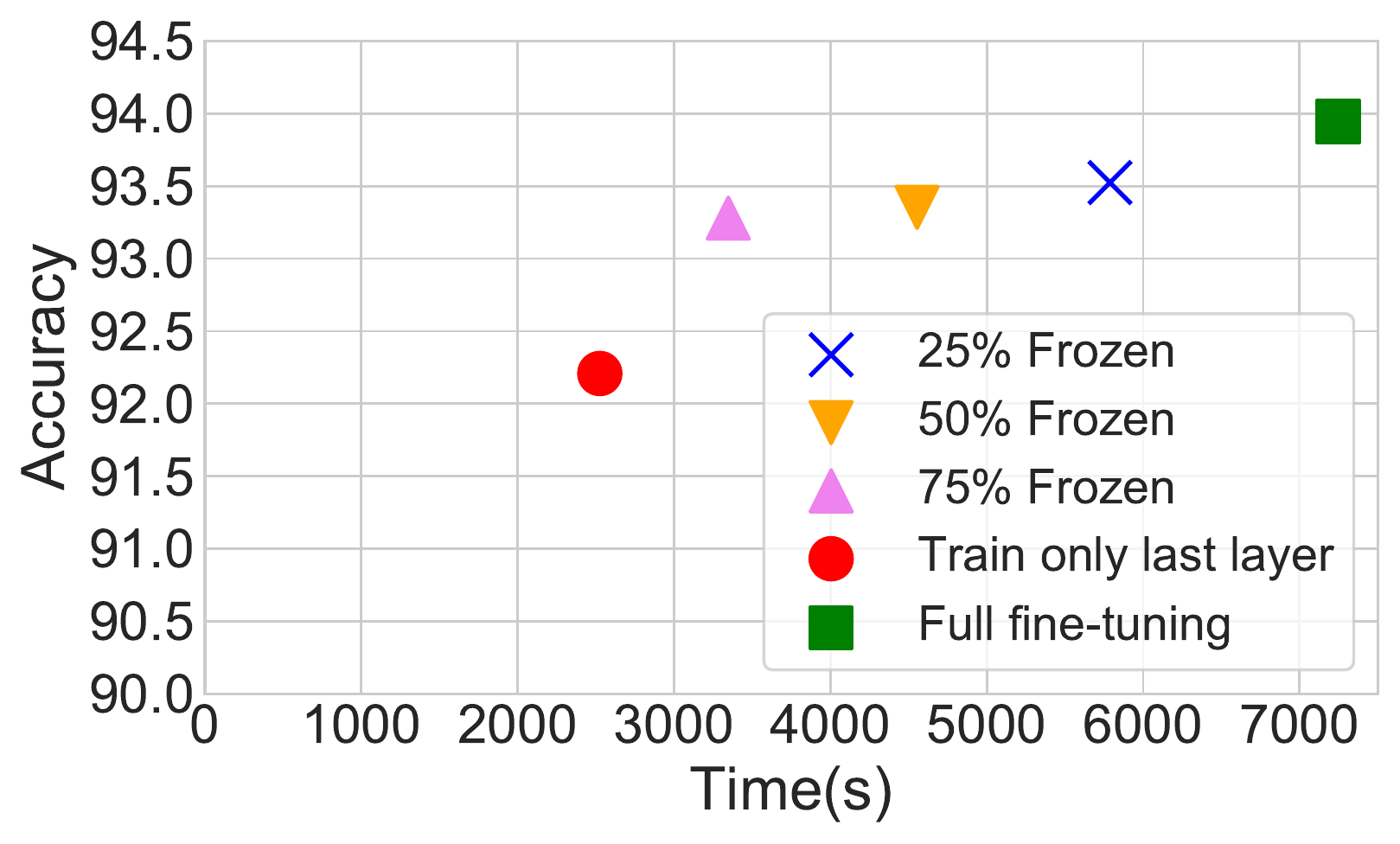}
 	\caption{IMDb}
 	\label{fig:imdb_fixed}
 	\end{subfigure}
 	\begin{subfigure}[t]{.3\linewidth}%
 	\center
 	\includegraphics[width=\linewidth]{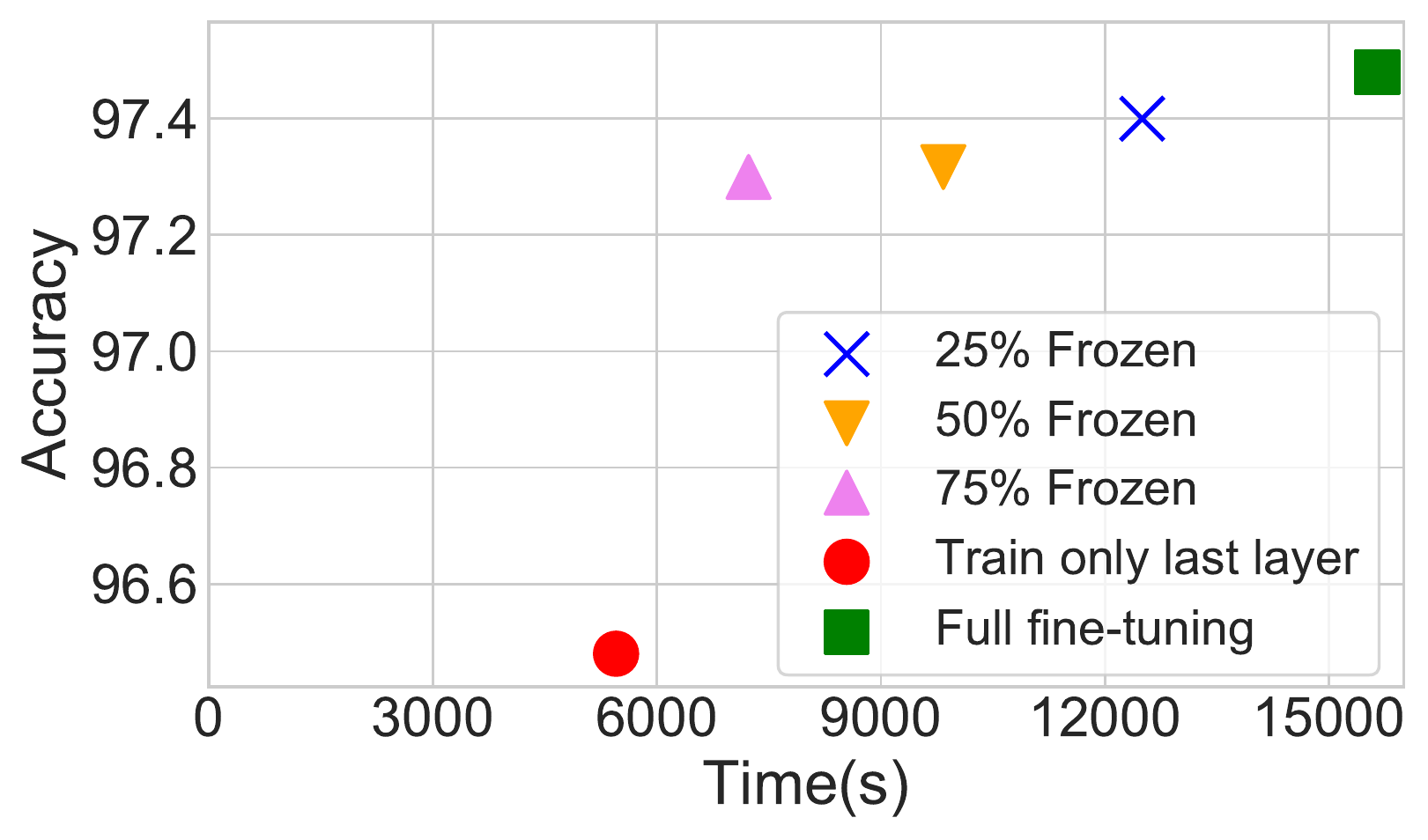}
 	\caption{Sogou News}
 	\label{fig:sogou_fixed}
 	\end{subfigure}
    \caption{\textbf{Evaluating Static Freezing:} We evaluate performance of static freezing (freezing a fixed number of layers) on pre-trained ResNet-18 and BERT. In (a), we use ResNet-18 pre-trained on CINIC-10 for fine-tuning on CIFAR-10. In (b) and (c), we fine-tune ${\rm BERT_{BASE}}$ on IMDb and Sogou News datasets. Static schemes improve fine-tuning time but often lead to loss in accuracy.
    }
    \label{fig:fixed_scheme}
\end{figure*}

First, we consider simple static freezing schemes where a fixed number of layers are chosen to be updated during training as presented in~\cite{lee2019would}. In Figure~\ref{fig:fixed_scheme} we compare static freezing schemes when fine-tuning $BERT_{BASE}$ with IMDb and Sogou dataset. We compare training the last 25\%, 50\%, 75\% of the layers, or only the last layer, to full fine-tuning. We see that such static freezing schemes lead to 0.5\% to 1.7\% accuracy drop for IMDb. On the other hand for Sogou, we observe that while some static freezing schemes only suffer 0.2\% accuracy loss, training only the last layer still leads to significant accuracy loss. We also see similar results for an image classification workload where we fine-tune ResNet-18 model~\cite{he2016deep} pre-trained on CINIC-10~\cite{darlow2018cinic} with CIFAR-10 dataset. In Figure~\ref{fig:cinic_transfer} we again see that training only the last layer leads to around 10\% reduction in accuracy, while training 50\% of the layers leads to 1.12\% reduction in accuracy.

Other approaches to improve fine-tuning~\cite{chen2020lottery} use the ``Lottery Ticket Hypothesis" to identify matching subnetworks such that a specific sparsity can be achieved. However, finding the sparsified subnetwork does not provide direct speedup because these approaches use unstructured pruning that prunes individual weights. EarlyBERT~\cite{chen2020earlybert} on the other hand identifies lottery-tickets in the early stage of BERT training, which provides training speedup compared to baselines, but results in accuracy degradation.

Overall, our results show that existing schemes either affect the accuracy of fine-tuning or provide limited speedups on existing hardware. Next, we propose developing a novel adaptive method that can select appropriate layers for freezing, thereby improving performance without affecting accuracy. 


\section{AutoFreeze Design}
We next describe the design of \sys{}, a system for automatically freezing model layers during fine tuning. We first discuss the scheme used by \sys{} to decide which layers to freeze and following that discuss how \sys{} can automatically cache intermediate outputs to further improve performance. Finally, we describe how \sys{} can improve resource utilization in distributed settings.

\subsection{Adaptive Freezing for Fine-tuning}
As described in Section~\ref{sec:static}, statically determining which layers to freeze can lead to reduced accuracy. Our insight is that layers which are closer to convergence are good candidates for freezing and by periodically inspecting the progress of each layer we can determine when a layer can be safely frozen.

\begin{figure*}[t!]
\centering
 	\begin{subfigure}[t]{.4\linewidth}%
 	\includegraphics[width=\linewidth]{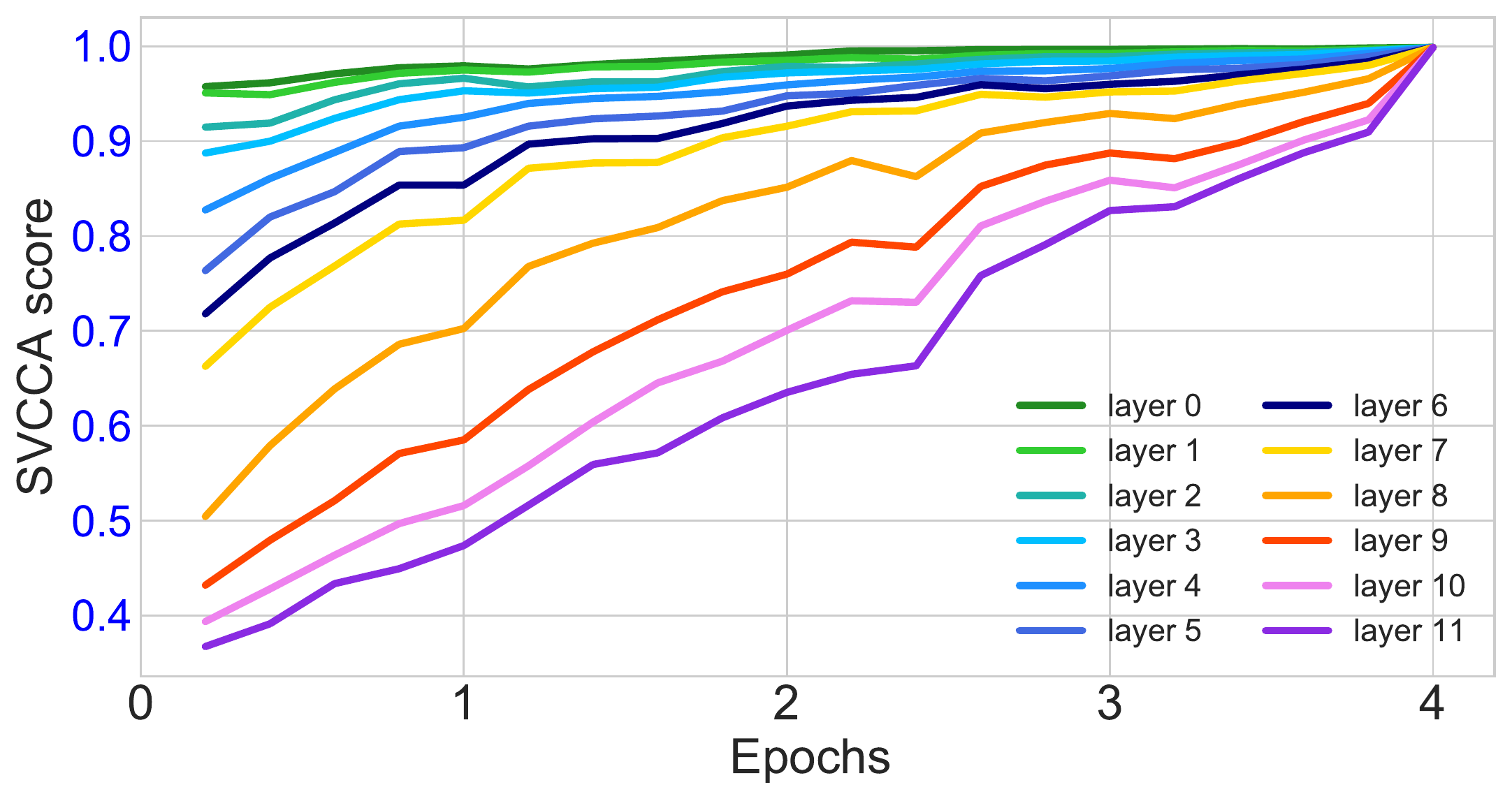}
 	\caption{IMDb}
 	\end{subfigure}
 	\begin{subfigure}[t]{.4\linewidth}%
 	\includegraphics[width=\linewidth]{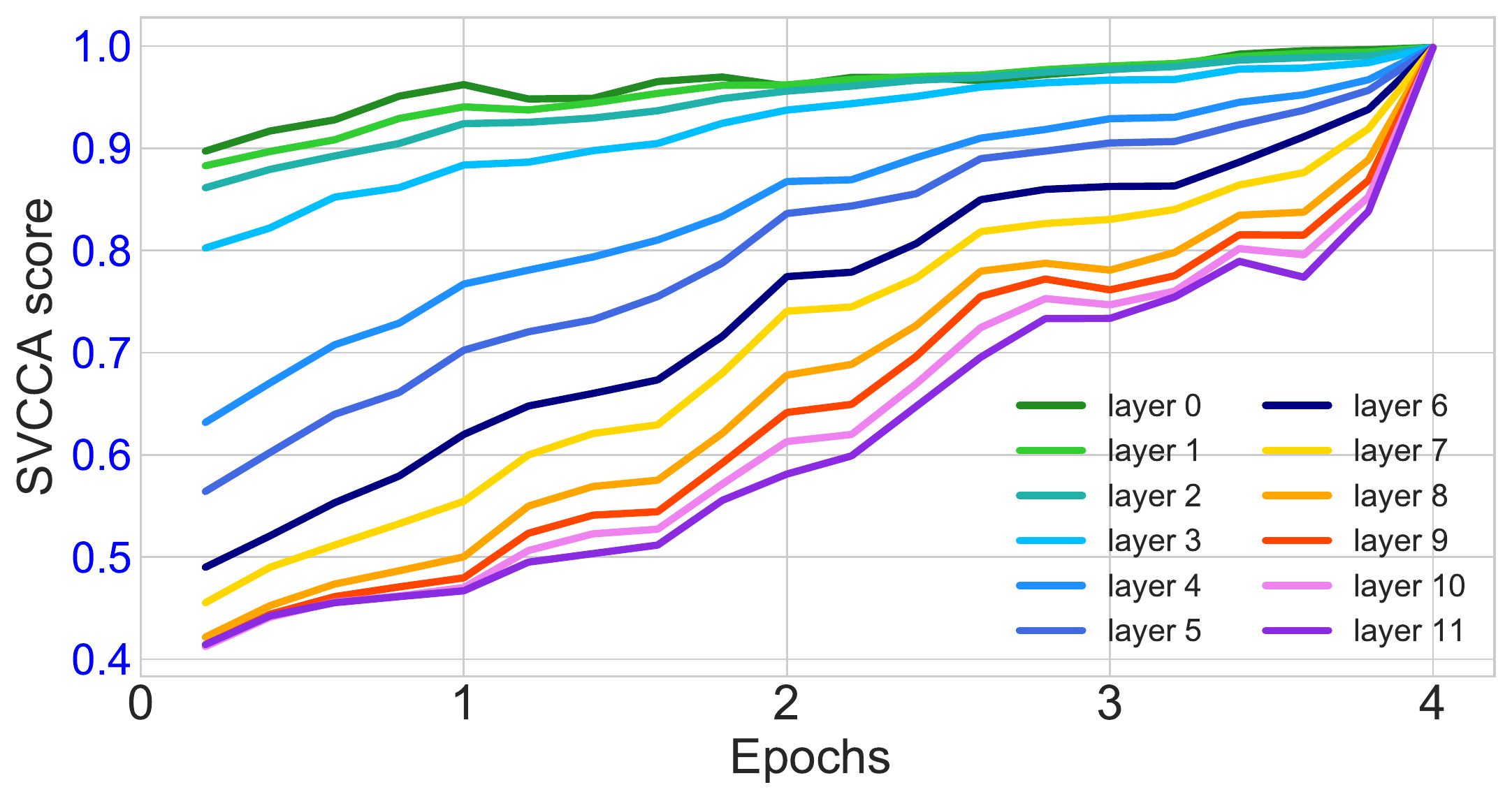}
 	\caption{Sogou News}
 	\end{subfigure}
 	\vspace{-0.15in}
 	\caption{\small{\textbf{SVCCA scores:} We calculate SVCCA score five times in each epoch for $BERT_{BASE}$ on two datasets (a) IMDb (b) Sogou. This shows that layers of a neural network converge bottom-up, \textit{i.e} earlier layers converge faster than later layers. This bottom-up convergence allows us to freeze earlier layers first allowing us to obtain computational benefits from freezing.}}
 \label{fig:svcca_score_compare}
 \vspace{-0.05in}
\end{figure*}

We validate our intuition by using SVCCA~\cite{Raghu2017SVCCASV}, a recently proposed metric for understanding convergence of neural networks. The SVCCA score is a metric which evaluates the similarity between two layers of neural network.
To understand convergence of each layer individually ~\citet{Raghu2017SVCCASV} perform a post-hoc analysis. They calculate SVCCA score by comparing the layers of the model during training with the layers of the already converged model. We use the same IMDB dataset as before and compute the SVCCA scores by comparing each layer's weights periodically (5 times every epoch in this case) with the final weights of the model. SVCCA scores range from $0$ to $1$ with $1$ indicating an exact match (i.e., that the intermediate weights match the final model weights). The results of this experiment are depicted in Figure~\ref{fig:svcca_score_compare}. We observe two main takeaways from this experiment in Figure~\ref{fig:svcca_score_compare}. First, we see that layers of the model converge in order with earlier layers (e.g. layers 0-4) reaching high SVCCA scores within one epoch.
Second, while some layers converge fast, others take significantly long time.
This indicates that an adaptive freezing scheme can provide performance benefits by freezing layers as they converge. 

The above data validates our intuition about the benefits of adaptively freezing model layers. It also shows that SVCCA score will be an ideal metric for freezing since it can track convergence of a layer and freeze it once the layer reaches convergence. However calculating the SVCCA scores shown in Figure~\ref{fig:svcca_score_compare} requires knowledge of the final model weights, making it inapplicable in practice. Thus we need an online method that can estimate if a layer can be frozen without knowing the final model weights. We next describe how we can use the gradient values at each layer to estimate this.

\begin{algorithm}[!tb]
   \caption{Freezing Module }
   \label{alg:freezing}
\begin{algorithmic}
   \STATE {\bfseries Input:} List of layers that are not frozen $activeLayers$, Percentile for freezing $N$
   \STATE {\bfseries Input:} accumulated gradients for current interval $\Delta_{T_{l}}$ and previous interval $\Delta_{T-1_{l}}$
   
   \FOR{$layer_l$ in $activeLayers$}
   
    \STATE $\eta_l = \left |  \norm{\Delta_{T-1_{l}}} - \norm{\Delta_{T_{l}}}  \right|/ \norm{\Delta_{T-1_{l}}}$ 
   \ENDFOR
   
   \FOR{ $layer_l$ in $activeLayers$}
   
   \IF{$\eta_l <$  $N^{th}$ percentile($\eta$)}
    \STATE freeze $layer_l$
   \ELSE
    \BREAK
   \ENDIF
   
   \ENDFOR
\end{algorithmic}
\end{algorithm}

\subsubsection{Gradient Norm Test} 
\label{sec:grad_norm}
We next present an online test to determine if a layer should be frozen. Our intuition in designing this test is that the rate of change of the gradient values for a layer can be used to determine how fast the model weights are being updated for a particular layer. Consider that we accumulate gradients for each layer in the model ($\Delta$) and perform our test at fixed intervals ($T$). Then we define the gradient norm change for layer $l$, $\eta_{l}$, as 
\begin{equation}
    \eta_l = \left| \norm{\Delta_{T-1_{l}}} - \norm{\Delta_{T_{l}}}  \right|/ \norm{\Delta_{T-1_{l}}}
\end{equation}

We next rank the layers in the order of $\eta_{l}$ to determine the layer that is changing slowest. Given our earlier observation about how layers converge in order, we can designate a layer to be frozen if all the layers preceding it are frozen and it is the slowest changing layer. However, this assumes a strict order in the rate of change of gradient norms and we can thus further relax this by designating a layer to be frozen if all the layers preceding it are frozen and if its rate of change is in the bottom $N^{th}$ percentile, where $N$ is a tunable parameter. Algorithm~\ref{alg:freezing} describes the above procedure. 

\begin{figure*}[t!]
\centering
 	\begin{subfigure}[t]{.4\linewidth}%
 	\center
 	\includegraphics[width=\linewidth]{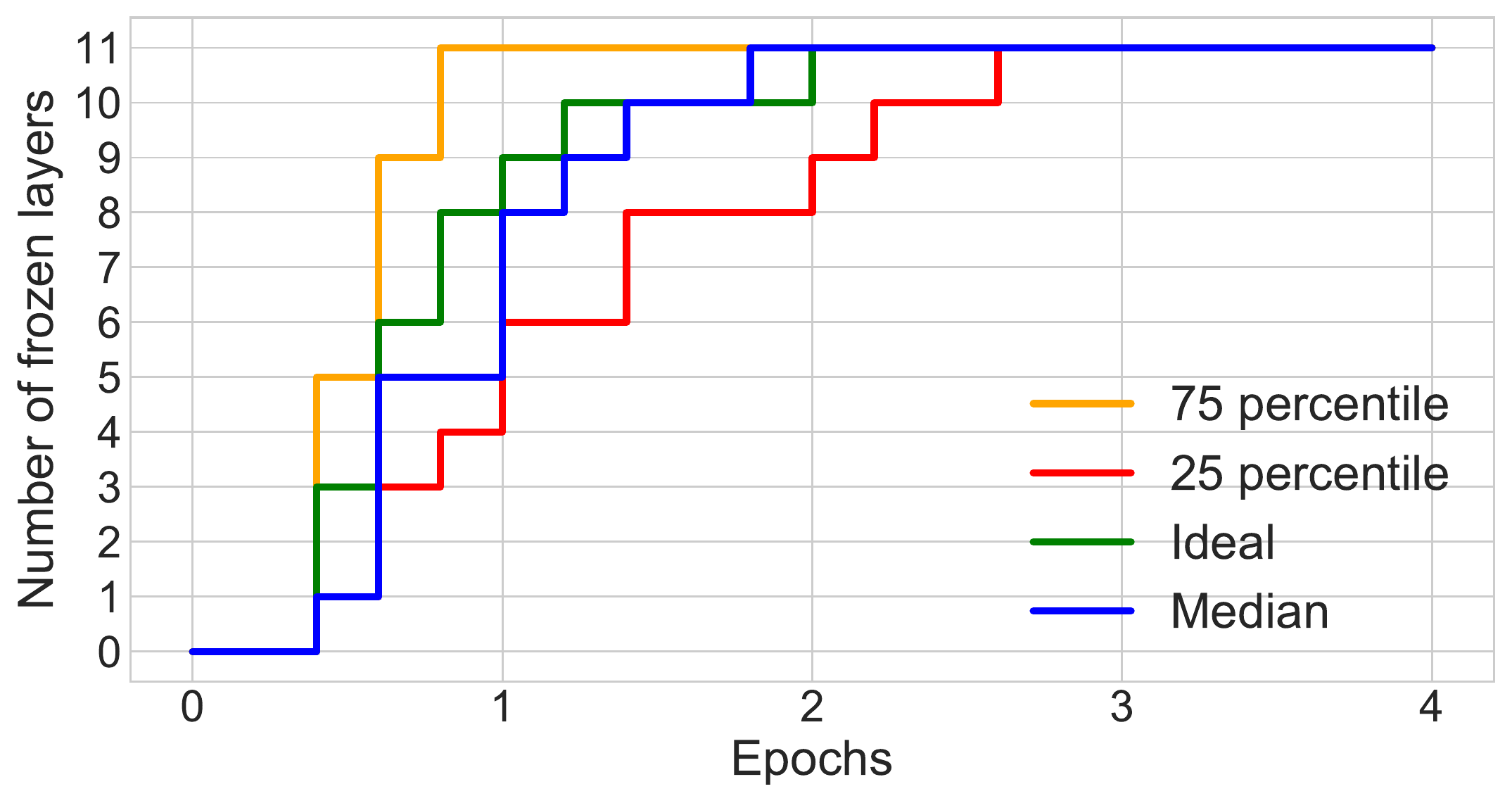}
 	\caption{IMDb}
 	\label{fig:freezing_layer_imdb}
 	\end{subfigure}
 	\begin{subfigure}[t]{.4\linewidth}%
 	\center
 	\includegraphics[width=\linewidth]{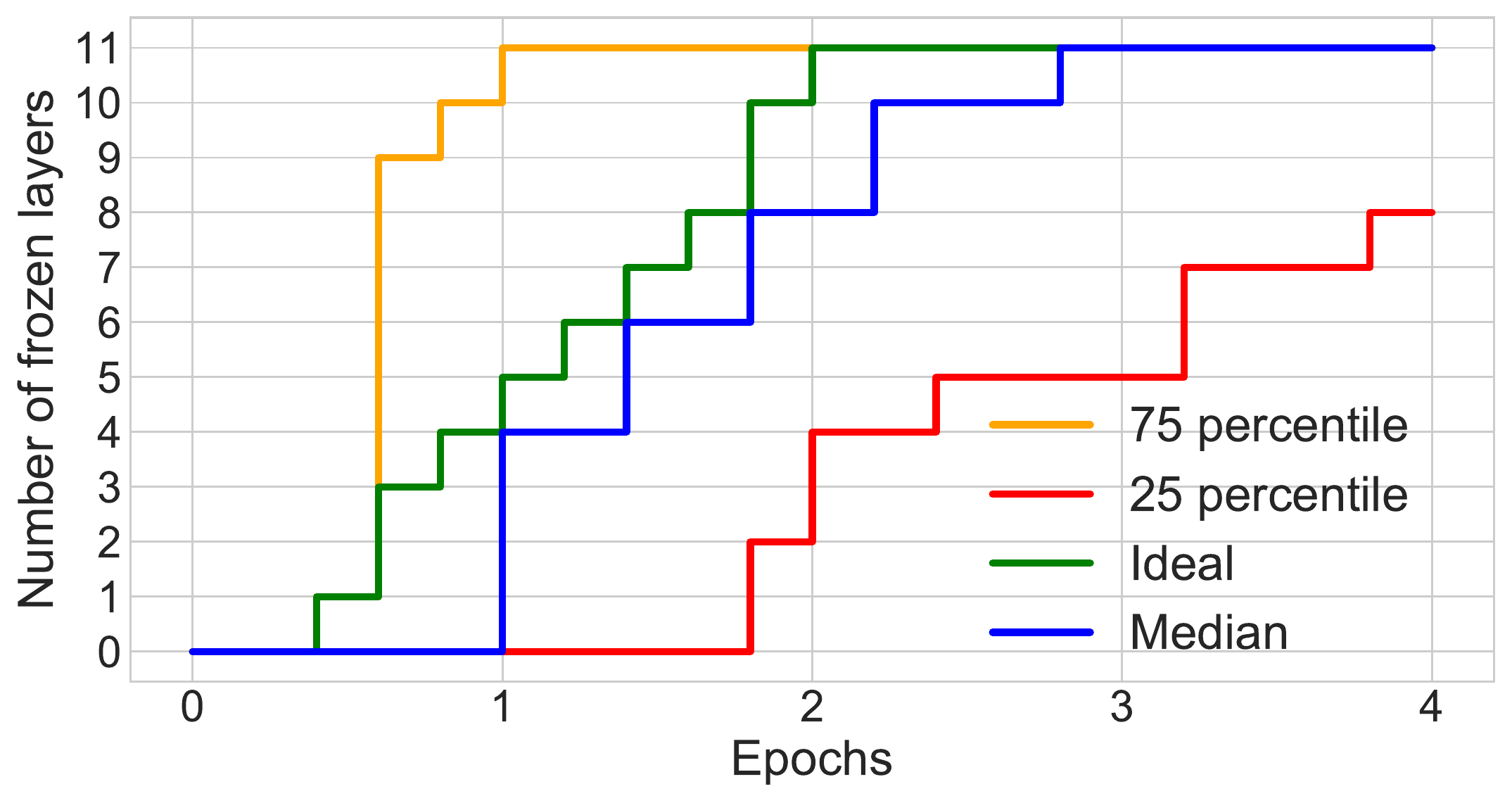}
 	\caption{Sogou News}
 	\label{fig:freezing_layer_sogou}
 	\end{subfigure}
 	\caption{\small{\textbf{Comparing Gradinet Norm Test with Ideal:}} We compare our gradient norm test for $\eta_l$= 25, 50 and 75 percentile, with an ideal SVCCA score based scheme on two datasets (a) IMDb (b) Sogou. We define the ideal scheme as freezing layers with SVCCA scores over 0.9 at each evaluation interval. We observe that gradient norm test with $\eta_l=$ 50 percentile (median) closely matches the ideal scheme.   }
 	\label{fig:svcca_grad_test_compare}
\end{figure*}

\noindent{\textbf{Comparison of Gradient Norm Test to SVCCA score}} 
 In Figure ~\ref{fig:svcca_grad_test_compare}, we evaluate the performance of our proposed gradient norm test by comparing it with the ideal SVCCA score based freezing scheme that has access to the final model weights. In the ideal scheme, we denote a layer as frozen if its SVCCA score compared to the final model weights is above a fixed threshold of 0.9.
In Figures~\ref{fig:freezing_layer_imdb} and~\ref{fig:freezing_layer_sogou}, we vary the percentile value used in Algorithm~\ref{alg:freezing} and see that using too low a percentile value (e.g., 25th percentile) can make the test too conservative resulting in fewer frozen layers compared to the ideal. We also see that using too high a percentile value can lead to the test being too aggressive resulting in loss of accuracy. Finally we see that the median closely tracks the ideal freezing scheme. We perform further evaluation of the effect of varying $N$ in  Section~\ref{sec:evaluation_freezing}.

\subsection{Caching Frozen Layers}
\label{sec:storage_manager}
Freezing a prefix of the model layers can help us avoid running the backward pass on those layers while fine-tuning. However, given that the layer weights are fixed once they are frozen, we can also avoid the forward pass if we are able to materialize and cache the intermediate output in CPU memory/disk. For example, consider a case where 50\% of the model layers are frozen after the first epoch. In this case if we can materialize the output of applying the first 50\% of the layers and save it to disk, then for the following epochs we can directly load this intermediate data and thus also avoid the corresponding 50\% of the forward pass. 

However, there are two main considerations in implementing the caching functionality. First, as model intermediate outputs can be large and can take some time for reading data from the cache, therefore we should only use caching when it will be faster than performing the forward pass. Second, the adaptive freezing algorithm described above, the number of layers frozen could be updated within an epoch making it challenging to determine which outputs should be saved and when. 

To solve the first consideration, we measure the size and time taken for reading intermediate outputs when fine-tuning the Yelp dataset. For every example, the intermediate output is around 1.57MB and this remains the same across all layers as all the transformer blocks in BERT have the same output size. 

However, given that we are limited to small batch sizes (around 6 examples on a P100), we only need to read around 10MB of data for one iteration and this takes around 25ms when using an SSD. On the other hand, doing a forward pass of \textbf{one layer} of $BERT_{BASE}$ takes around \textbf{11ms}. Thus, in this case, we can see that loading data from SSD should provide a speed-up when more than 2 layers are frozen. 
In general, the trade-off between caching and repeating the forward pass depends on the disk bandwidth, batch size and computation speed of the GPU. Evaluating this trade-off is not expensive in practice as few iterations of training can indicate how many layers of a model need to be frozen before caching becomes advantageous. 

To resolve the second consideration about which layers to store, we design the storage manager which we describe next.

\begin{figure*}[t!]
 	\begin{subfigure}[t]{.48\linewidth}%
 	\includegraphics[width=\linewidth]{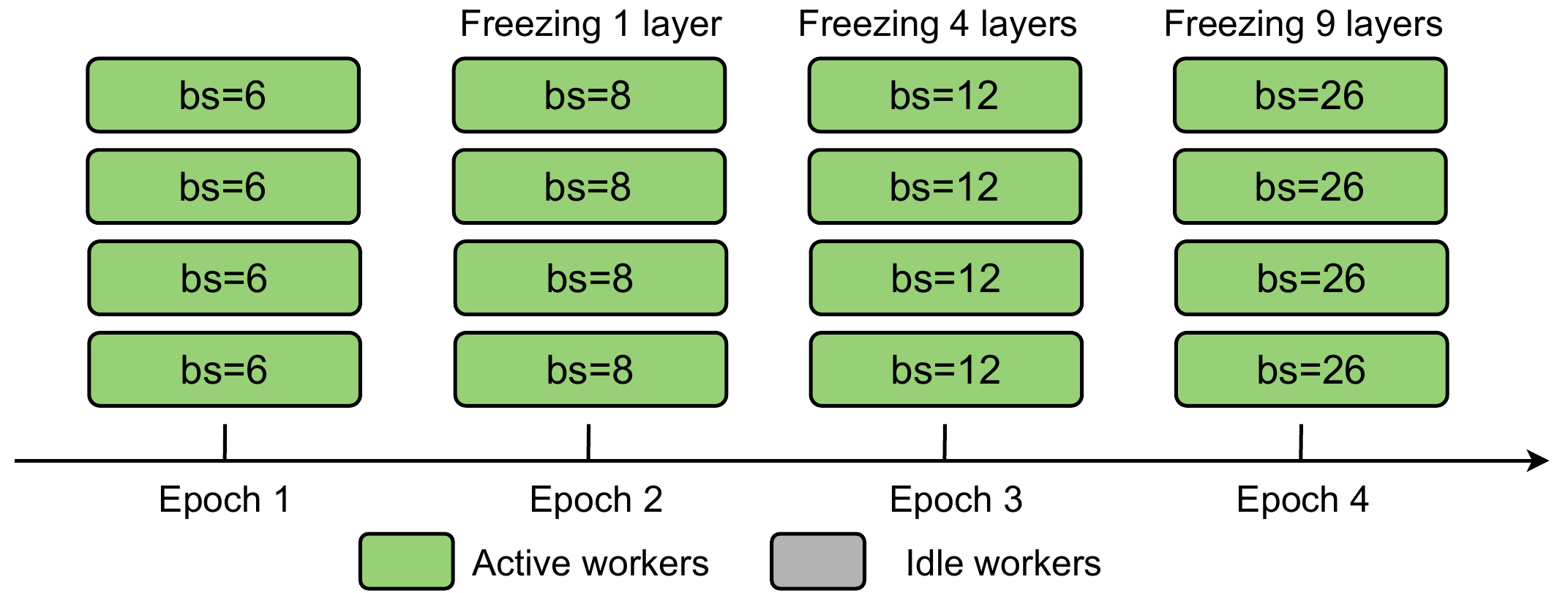}
 	\caption{Performance Packing}
 	\label{fig:perf_mode}
 	\end{subfigure}
 	\begin{subfigure}[t]{.48\linewidth}%
 	\includegraphics[width=\linewidth]{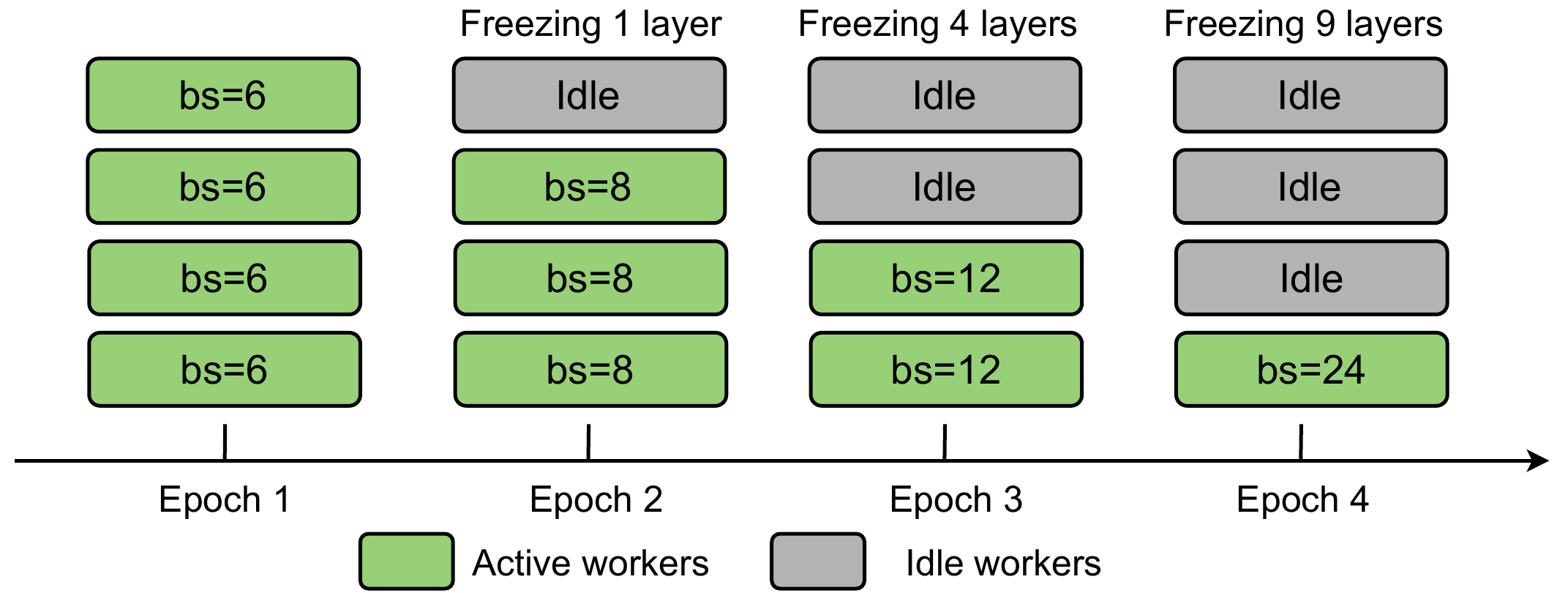}
 	\caption{Efficiency Packing}
 	\label{fig:eff_mode}
 	\end{subfigure}
 	\caption{Two modes for distributed fine-tuning enabled by \sys{}: (a) Performance Packing: We keep number of GPUs during training constant, and as memory consumption decreases due to freezing we increase per GPU batch size to the max batch size which fits on a single GPU. (b) Efficiency Packing: We reduce the number of GPUs during training to the minimum number of GPUs that can maintain the original total effective batch size (e.g. 24).  }
 \label{fig:}
\end{figure*}

\begin{figure}[!t]
\centering
    \includegraphics[width=0.45\textwidth]{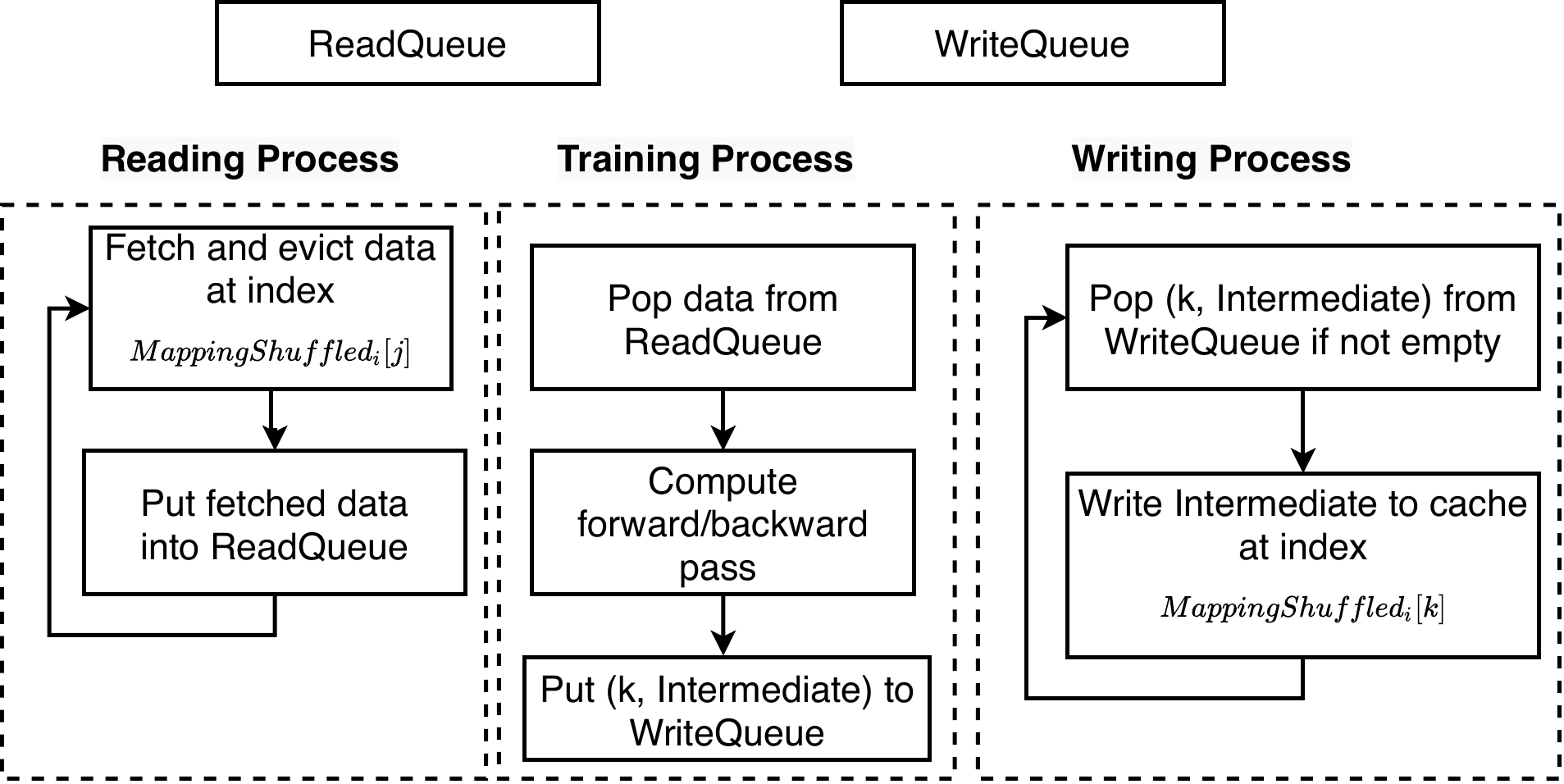}
    \caption[]{\textbf{Storage Manager Design:} In the Storage Manager, reading and writing processes run in parallel to the main training process. The reading process fetches data from cache, while the main training process consumes data from the read queue and produces intermediate outputs.  The writing process saves intermediate outputs to cache. }
    \label{fig:storage_manager_design}
    \hspace{0.1in}
\end{figure}

\begin{figure*}[!t]
\centering
    \includegraphics[width=0.8\textwidth]{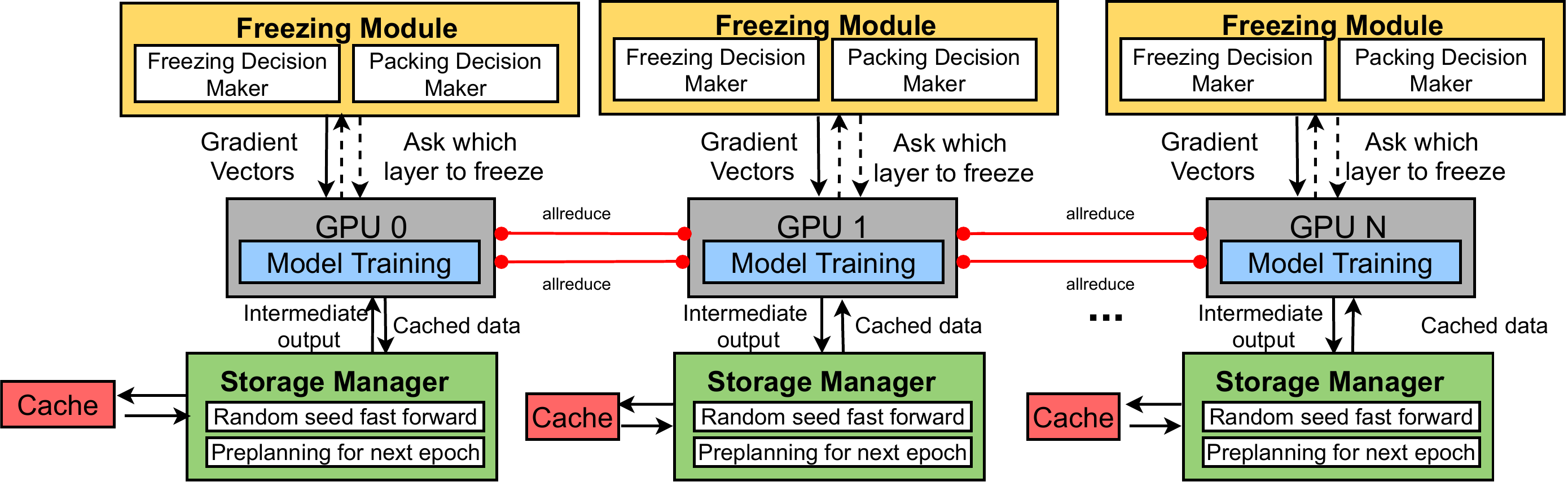}
    \caption[]{\textbf{\sys{} System Design:} In the distributed fine-tuning setting, each worker consists of: (1) Freezing Module that accumulates gradient vectors that are aggregated and decides which layers to freeze based on gradient norm test. (2) Storage Manager caches intermediate activations to disk when the overhead of caching is smaller than forward computation. It also fetches data from disk to save forward pass computation for later epochs.}
    \label{fig:system}
\end{figure*}

\noindent\textbf{Storage Manager}
To handle caching of layers we design a storage manager (Figure~\ref{fig:storage_manager_design}).
The storage manager is responsible for managing where data is cached and up to what layer should the forward pass be executed before saving to cache.
The storage manager notes down the layer $L$ up to which the model was frozen. In that epoch for all data points that are processed, the output of the forward pass up to layer $L$ is written to cache.

We store the intermediate output to disk when it no longer fits in CPU memory. When the dataset ($D$ points) is larger than the disk space available, we save $I$ points to disk ($I < D$ and $I$ is the maximum number of points that fit on disk). During the next epoch, if the number of layers currently frozen is greater than the number of layers of forward pass that were completed before data was saved to cache, 
then the storage manager also evicts the data points once they are read. Based on the fact that the number of data points to be written will never surpass the number of data points read from disk, we will never exceed the disk space. 

Finally as model training typically shuffles data across epochs, the storage manager also ensures that cached data points can be transparently accessed even if only part of the dataset has been cached. We do this by maintaining $MappingShuffled_{i}$ a mapping from the shuffled indices to the original indices for each epoch $i$. When writing the intermediate outputs to disk at epoch $i$, we write at the original indices retrieved from $MappingShuffled_{i}$. To read the data at index $k$ at epoch $i+1$ required for training, we read $MappingShuffled_{i+1}[k]$. While this approach incurs random reads/writes to the cache, since each data item is relatively large ($\sim$1.5MB) we have not found this to be an issue in practice. Finally, as shown in Figure~\ref{fig:storage_manager_design}, we perform read, write and gradient computation in separate processes thereby pipelining I/O with compute.

\DeclarePairedDelimiter\ceil{\lceil}{\rceil}

\subsection{Distributed Fine-tuning}
\label{sec:distr_description}
We next extend our design to consider how \sys{} can help when multiple workers are used for fine-tuning. A common approach to speeding up ML model training (and correspondingly fine-tuning) is to use multiple workers in parallel. In this scenario, the most widely used training mode is a ``data-parallel'' mode where each worker calculates the gradient on a batch of data and the gradient values are then aggregated using AllReduce to compute the updated model. Adaptively freezing the layers of a model that are being fine-tuned can lead performance and cost-efficiency improvements in a distributed setup. To understand how, we first construct a performance model that captures the computation and communication savings from freezing layers and then explain two distributed execution modes in \sys{}.

    We consider a distributed training scenario that is similar to DistributedDataParallel (DDP) in Pytorch~\cite{li2020pytorch}. We associate $T_{comp}$ as the time to compute gradients on each worker. Assuming that a model can be partitioned into $k$ buckets, where the first $k-1$ buckets are of size $b$ while the last bucket is of size $\hat{b}$, we denote the time for communicating $k$ gradient buckets across $p$ machines using a bandwidth of $BW$ as $T_{comm}$. In DDP training, the gradient communication is overlapped with the gradient communication and gradients are aggregated (e.g., by calling AllReduce) whenever a bucket becomes ready. Accordingly, using the standard communication model~\cite{alphabetamodel,thakur2005optimization}, the communication time for a single iteration can be modeled as:
\[
 T_{comm}(k,b,p,BW) = k \times (\alpha \times (p-1) + 2 \times b \times \frac{(p-1)}{p \times BW} ) 
\]
where $\alpha$ represents the latency cost (i.e., cost per message sent). 
The overall time taken to run one epoch for a dataset consisting of $N$ examples with batch size $BS$ becomes the number of iterations times the time per iteration.
\[
\frac{N}{BS} \times ( max(T_{comp}, T_{comm}) ) 
\]

In addition to computation savings describe before, in the distributed fine-tuning setup, \sys{} provides the following additional savings:

\noindent\textbf{Communication Savings} As explained before, our approach in \sys{} can lead to skipping gradient computation for layers. Correspondingly we only need to communicate the gradients that have been computed leading to a reduction in the number of bytes sent. 

\noindent\textbf{Memory Savings} 
The peak memory used during is dominated by following three components (i) model weights, (ii) gradients and intermediate activations (iii) number of data points (batch size) in the active mini-batch.

By freezing layers \sys{} skips the gradient computation of frozen layers thus reducing memory used for storing gradients and intermediate activations. This reduction in memory can be used to increase the batch size or reduce the number workers to improve performance and efficiency as we describe next.

Based on the above savings, we next design two modes that can maximize the efficiency and performance of distributed fine-tuning when freezing model layers: (1) \emph{Efficiency Packing mode} (2) \emph{Performance Packing mode}. 

\noindent\textbf{Efficiency Packing mode } 
Our goal in the efficiency packing mode is to \emph{minimize the cost} of distributed fine-tuning. Reducing the cost implies reducing the number of
worker-hours used for fine-tuning. Because the time per iteration increases with the number of machines used (latency term $\alpha$ in the performance model), the \emph{most cost efficient configuration is to use the least number of workers} for a given batch size. Thus, in the efficiency packing mode, we increase the per-worker batch size when layers are frozen and correspondingly reduce the number of workers used in training to maintain the total batch size constant as shown in Figure~\ref{fig:eff_mode}. Denoting the initial total effective batch size as $b$ and initial batch size per worker as $b_{0}$, when freezing $l$ layers reduces memory required, we can increase the batch size in a worker to $b_{l}$ without violating memory constraint. Correspondingly we reduce the number of workers by $\frac{b_{l}}{b_{0}}$, thus maintaining the total effective batch size $b$. 

\noindent\textbf{Performance Packing mode }
In contrast to Efficiency Packing mode, the goal of Performance packing is to \emph{minimize wall clock time} for fine tuning. 

From our performance model we observe that for a given cluster of $p$ workers, the wall clock time is minimized when  we reduce the number of iterations per epoch. Reducing number of iterations, reduces number of gradient synchronization steps which in turn reduces the total communication overhead. The memory savings from freezing model layers allows us to use larger batch sizes, thus reducing the number of iterations in a single epoch.

More concretely, in the Performance Packing mode, we keep the number of workers used in training $p$ constant  and increase the per worker batch size $b_{i}$ as memory becomes available (Figure~\ref{fig:perf_mode}) because of freezing layers. Thus, the total effective batch size is increased, reducing frequency of gradient synchronization while using the maximum parallelism available, and thus minimizing the end-to-end training time.

\noindent\textbf{Example:}
Consider a dataset that contains 120K data points being fine tuned on 64 GPUs. Initially we start with batch size of 384 (6 on each GPU) with no layers frozen.  If \sys{} decides to freeze the first 11 BERT layers at some point, Performance Packing increases the batch size to 3456 (54 each machine), while Efficiency Packing keeps the total effective batch size as 384 and reduces number of GPU machines to 8 keeping the batch size 384 (48 per machine). For the next epoch after freezing, the time taken with Performance Packing and Efficiency Packing comes out to 36 seconds and 131 seconds respectively. Assuming that cost/second of a GPU is $c$, the total cost of this epoch with Performance Packing is $36\times64\times c = 2304\times c$, while that of Efficiency Packing is $131\times8\times c = 1048\times c$. We can see that Efficiency Packing has 2.2$\times$ lower cost but Performance Packing is 3.3$\times$ faster.

It is easy for users of \sys{} to configure which mode to choose based on their needs. Finally, utilizing Efficiency Packing mode maintains a fixed total effective batch size which we show in Section~\ref{sec:dist_eval} ensures minimum accuracy loss. While using Performance Packing mode users can also configure a maximum batch size and \sys{} will ensure that total effective batch size does not go beyond the configured threshold.

\subsection{Overall Design, Implementation}
Putting the above subsections together, the design of our system, \sys{}, is shown in Figure~\ref{fig:system}. There are two modules on every GPU (e.g. GPU$0$ in Figure~\ref{fig:system}): (1) Freezing Module that makes decision on the set of layers to freeze at different intervals of the fine-tuning procedure. (2) Storage Manager that caches intermediate outputs of the BERT encoder to disk in parallel to training when necessary. 
We implement \sys{} in Python and design it to work with PyTorch models~\cite{paszke2019pytorch}. 
In the distributed fine-tuning setting, the Freezing Module is called on the synchronized gradients on each worker, producing the same freezing decision. When Caching is enabled for the distributed training setting, each GPU manages its own cache to avoid any data movement across machines. 

\section{Evaluation}
\label{sec:eval}
We next evaluate \sys{} on a number of NLP datasets and tasks, and measure the performance benefits and model accuracy. We compare \sys{} to existing baselines and also study scalability by using up to 64 GPUs.

\begin{table*}[]
\centering
\begin{footnotesize}
\begin{tabular}{@{}lllllllllllllll@{}}
\toprule
\multirow{2}{*}{} & \multicolumn{2}{l}{\begin{tabular}[c]{@{}l@{}}Full\\ fine-tuning\end{tabular}}                           & \multicolumn{3}{l}{\begin{tabular}[c]{@{}l@{}}Frozen \\ up to 9th\end{tabular}}                                     & \multicolumn{3}{l}{\begin{tabular}[c]{@{}l@{}}Frozen\\ up to 12th\end{tabular}}                                    & \multicolumn{3}{l}{AutoFreeze}                                                                                     & \multicolumn{3}{l}{Pruning}                                                                                         \\ \cmidrule(l){2-15} 
                  & \begin{tabular}[c]{@{}l@{}}Acc/\\ Corr\end{tabular} & \begin{tabular}[c]{@{}l@{}}Time\\ (s)\end{tabular} & \begin{tabular}[c]{@{}l@{}}Acc/\\ Corr\end{tabular} & \begin{tabular}[c]{@{}l@{}}Time \\ (s)\end{tabular} & spd & \begin{tabular}[c]{@{}l@{}}Acc/\\ Corr\end{tabular} & \begin{tabular}[c]{@{}l@{}}Time\\ (s)\end{tabular} & spd & \begin{tabular}[c]{@{}l@{}}Acc/\\ Corr\end{tabular} & \begin{tabular}[c]{@{}l@{}}Time\\ (s)\end{tabular} & spd & \begin{tabular}[c]{@{}l@{}}Acc/\\ Corr\end{tabular} & \begin{tabular}[c]{@{}l@{}}Time\\ (s)\end{tabular} & Sparsity \\ \midrule
MRPC~\cite{dolan-brockett-2005-automatically}             & 87.01                                               & 132                                                & 76.47                                               & 66                                                  & 2x      & 69.36                                               & 43                                                 & 3.07x   & 86.27                                               & 112                                                & 1.18x   & 85.04                                               & 133                                                & 50\%     \\
SST-2~\cite{socher-etal-2013-recursive}            & 92.54                                               & 2476                                               & 91.5                                                & 1307                                                & 1.89x   & 86.01                                               & 907                                                & 2.73x   & 91.6                                                & 1665                                               & 1.49x   & 91.7                                                & 2517                                               & 60\%     \\
CoLA~\cite{warstadt2019neural}              & 56.65                                               & 309                                                & 51.53                                               & 150                                                 & 2.06x   & 32.98                                               & 99                                                 & 3.12x   & 56.05                                               & 176                                                & 1.74x   & 52.75                                               & 318                                                & 50\%     \\ \bottomrule
\end{tabular}
\end{footnotesize}
\caption{Performance achieved by full fine-tuning, \sys{}, static freezing, and Lottery Ticket Hypothesis on MRPC, SST-2, and CoLA datasets. For CoLA dataset, we report the Matthew’s Correlation metric. For MRPC and SST-2 datasets, we report the accuracy. }
\label{tab:baselines}
\end{table*}

\begin{table}
\centering
  \begin{tabular}{c|c|c|c}
  \textbf{Dataset} & \textbf{Num Train} & \textbf{Num Test} & \textbf{Type} \\
  \hline
  Yelp F.~\cite{zhang2016characterlevel} & 650,000 & 50,000 & Sentiment\\
  \hline
  Sogou News~\cite{sun2019fine} & 54,000 &  6,000 & Topic \\
  \hline
  AG's News~\cite{zhang2016characterlevel} & 120,000 &  7,600 & Topic \\
  \hline
  IMDb~\cite{maas-etal-2011-learning} & 25,000 & 25,000 & Sentiment\\
  \hline
  SQuAD2.0~\cite{rajpurkar2018know}  & 131,944                                                  & 12,232 & Question \\
  \hline
  SWAG~\cite{zellers2018swag} & 73,546 & 20,006 & Multiple Choice \\
  \hline
  CNN~\cite{hermann2015teaching} & 90,266 & 1,093 & Text Summary \\
  \hline
  DailyMail~\cite{hermann2015teaching} & 196,961 & 12,148 & Text Summary \\
  \end{tabular}
 
 \caption{Statistics and types of datasets used. }
  \label{tab:dataset_stat}
 \end{table}
\subsection{Datasets, hyper-parameters}
 In our experimental study we evaluate \sys{} on- (i) four text classification datasets, (ii) one question answering dataset, (iii) one multiple choice dataset,  (iv) one combined text summarization dataset. The details of the dataset can be found in Table ~\ref{tab:dataset_stat}. We also use three datasets from the GLUE benchmark, which are all classification tasks, to compare \sys{} against Lottery Ticket Hypothesis \cite{chen2020lottery} the details of datasets can be found in Table~\ref{tab:baselines}. 

Across all the above tasks from (i) to (iii), we set the per epoch evaluation intervals during fine-tuning to 5, and we perform the gradient norm test every $\frac{k}{5}$ iterations (where k = total number of iterations per epoch). As in prior work~\cite{howard2018universal}, we used stepped learning rate schedule that decays to slanted triangular learning rate at 0.3 and 0.6 proportions of total iterations from initial learning rate of 1e-5. We set the percentile value for our adaptive freezing algorithm to be the 50th percentile by default unless specified. We run \sys{} for three runs with different random seeds for each dataset. In general, when transformer blocks of the BERT Encoder are frozen, we also freeze the Embedding layer as Autograd does not allow backward gradient flow ~\cite{NoceWrig06} when earlier layers are frozen, \ie when earlier blocks of the BERT encoder are frozen, the gradients for them are not available, so calculating the gradients for the Embedding layer before the Encoder cannot be achieved.  All single GPU experiments are performed on a Azure P100 VM unless otherwise specified.

\subsection{Comparison with baselines}
\label{subsec:baselines}
We compare \sys{} with two baseline - (i)Lottery Ticket Hypothesis~\cite{chen2020lottery} (LTH) and (ii) static freezing (Table~\ref{tab:baselines}).

\noindent{\bf Lottery Ticket Hypothesis:} For comparison we used the same datasets as in prior work~\cite{chen2018lstd}. For LTH, we report the fine-tuning time after finding the winning subnetworks by running Iterative Magnitude Pruning. Although it provides comparable accuracy with full fine-tuning, it does not provide training speedup because it utilizes unstructured magnitude pruning where weights are pruned individually instead of a group. Thus, while the winning subnetwork found by the Lottery Ticket Hypotheses is sparse (i.e. has some fraction of weights as 0), it does not provide actual speedup because there is no computation saving due to lack of support for sparse operations in modern accelerators like GPUs and TPUs~\cite{You2020Drawing}. Overall we find that \sys{} can improve performance by up to $1.74\times$ (Table~\ref{tab:baselines}) with minimal degradation in accuracy when compared to existing baselines.

\noindent{\bf Static Freezing:}
As shown in the Table~\ref{tab:baselines}, static freezing provides significant training speedup but leads to poor accuracy on several different downstream tasks. For example, we observe that freezing up to the $9^{th}$ layer of the BERT encoder results in around 11\% accuracy loss for MRPC, while it only results in around 1\% accuracy loss for the SST-2 dataset. 

\subsection{Freezing, Caching Benefits}
\label{sec:evaluation_freezing}

To further verify the effectiveness of our freezing scheme in terms of training speedup and model accuracy, we apply \sys{} on a number of NLP tasks, and compare it with full fine-tuning of $BERT_{BASE}$. We use a single Azure NC6 VM for these experiments.

\begin{figure*}[!t]
    \centering
    \includegraphics[width=0.8\textwidth]{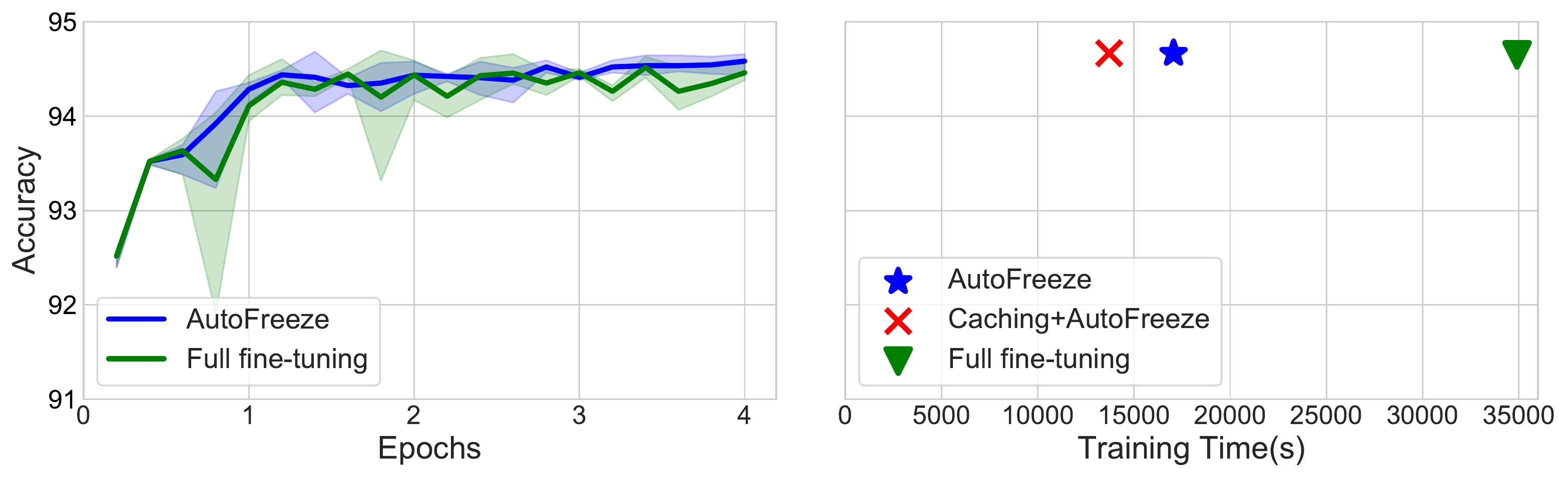}
        \caption{[\textbf{AG} (4 epochs)] \textbf{Left}: Accuracy achieved by \sys{} and fine-tuning for four epochs. Mean and range of accuracy are plotted. \textbf{Right}: Average end-to-end training time for \sys{}, \sys{} with caching, and full fine-tuning across three runs. \sys{} has 2.05$\times$ average improvement in training time when only using the freezing module, and  2.55$\times$ when also using caching. }
    \label{fig:ag_res}
\end{figure*}

\begin{figure*}[!t]
    \centering
    \includegraphics[width=0.8\textwidth]{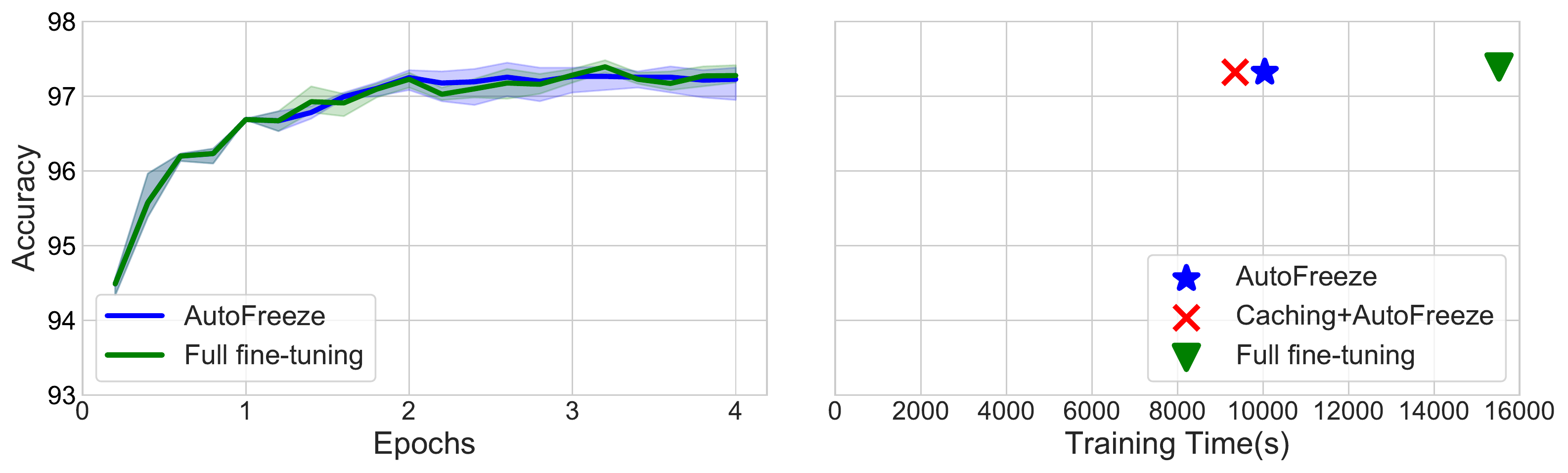}
        \caption{[\textbf{Sogou} (4 epochs)] \textbf{Left}: Accuracy achieved by \sys{} and fine-tuning for four epochs. Mean and range of accuracy are plotted. \textbf{Right}: Average end-to-end training time for \sys{}, \sys{} with Caching, and full fine-tuning across three runs. \sys{} has 1.55$\times$ improvement in training time when only using the freezing module, and 1.66$\times$ when also using caching. }
        \vspace{-0.1in}
    \label{fig:sogou_res}
\end{figure*}

\begin{figure*}[ht]
    \centering
    
    \includegraphics[width=0.8\textwidth]{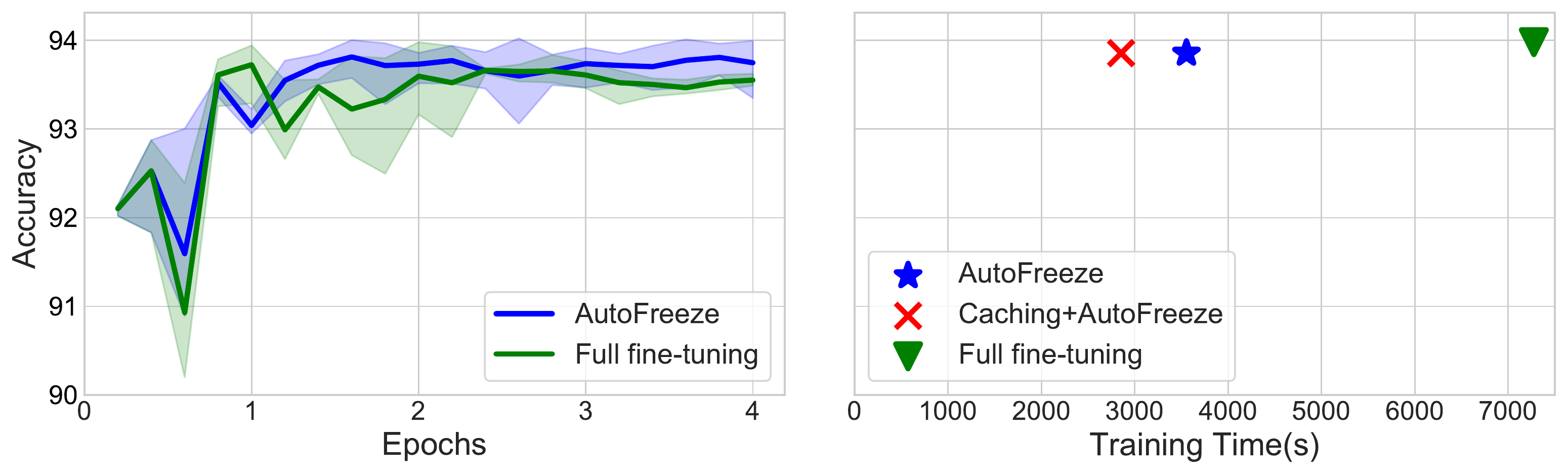}
        \caption{[\textbf{IMDb} (4 epochs)] \textbf{Left}: Accuracy achieved by \sys{} and fine-tuning for four epochs. Mean and range of accuracy are plotted. \textbf{Right}: Average end-to-end training time for \sys{}, \sys{} with Caching enabled, and full fine-tuning across three runs. \sys{} has 2.05$\times$
        improvement in training time when only using the freezing module, and 2.55$\times$ when also using caching.}
    \label{fig:imdb_res}
\end{figure*}

\begin{figure*}[ht]
    \centering
    
    \includegraphics[width=0.8\textwidth]{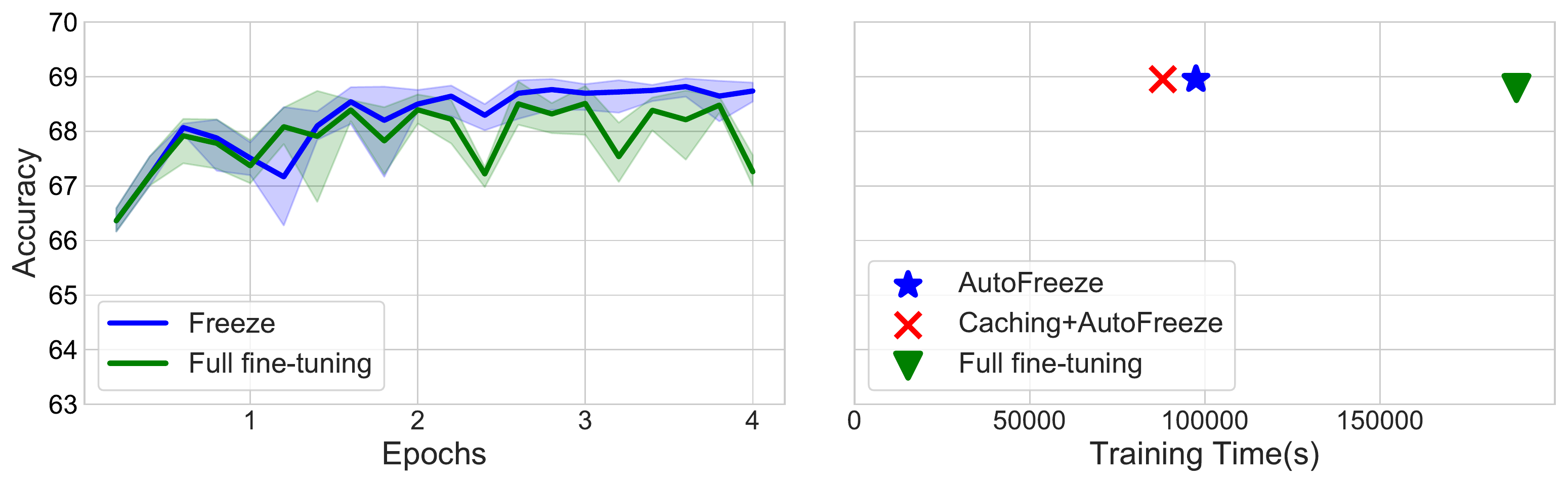}
        \caption{[\textbf{Yelp F.} (4 epochs)] \textbf{Left}: Accuracy achieved by \sys{} and fine-tuning for four epochs. Mean and range of accuracy are plotted. \textbf{Right}: Average end-to-end training time for \sys{}, \sys{} with Caching enabled, and full fine-tuning across three runs. \sys{} has 1.94$\times$ improvement training time when only using the freezing module and 2.15$\times$ when also using caching. }
        \vspace{-0.1in}
    \label{fig:yelp_res}
\end{figure*}

\begin{figure*}[!t]
    \centering
    \includegraphics[width=0.8\textwidth]{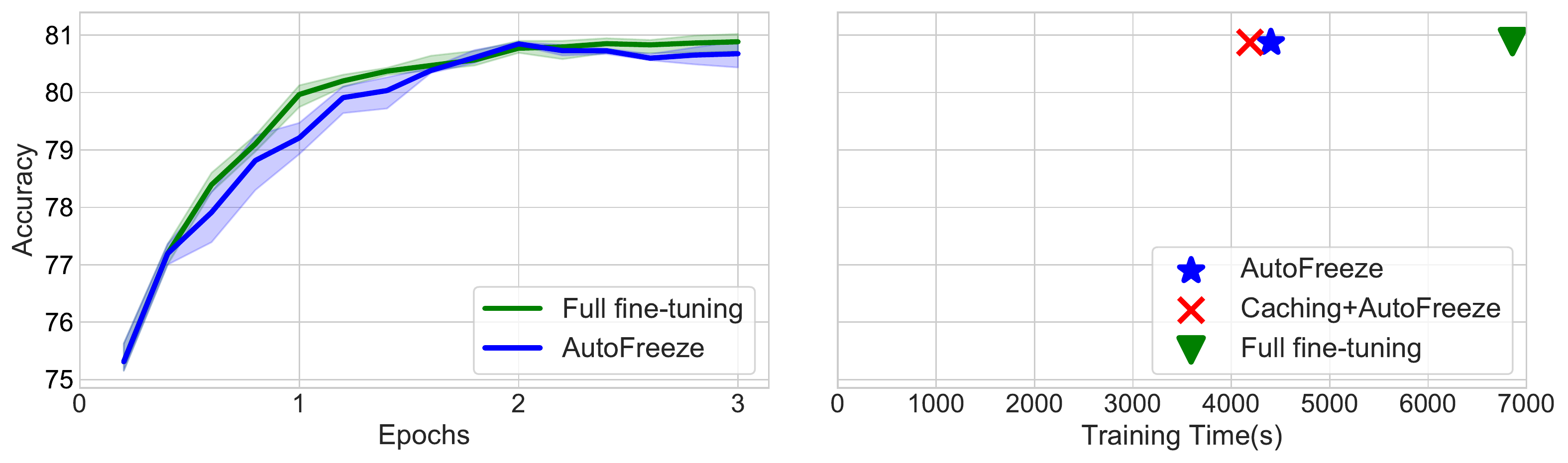}
        \caption{[\textbf{SWAG} (3 epochs)] \textbf{Left}: Accuracy achieved by \sys{} and fine-tuning for four epochs. Mean and range of accuracy are plotted. \textbf{Right}: Average end-to-end training time for \sys{}, \sys{} with caching, and full fine-tuning across three runs. \sys{} has 1.56$\times$ average improvement in training time when only using the freezing module, and  1.64$\times$ when also using caching. }
    \label{fig:swag_res}
\end{figure*}

\begin{figure*}[!t]
    \centering
    \includegraphics[width=0.8\textwidth]{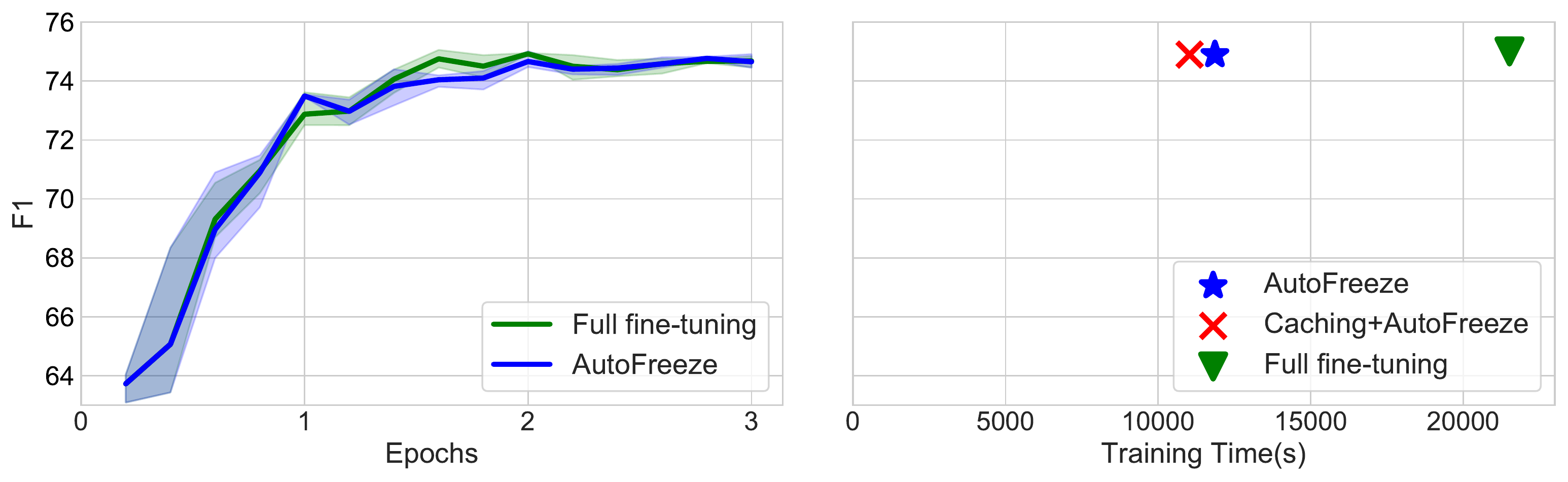}
        \caption{[\textbf{SQuADv2.0} (3 epochs)] \textbf{Left}: F1 achieved by \sys{} and fine-tuning for four epochs. Mean and range of accuracy are plotted. \textbf{Right}: Average end-to-end training time for \sys{}, \sys{} with caching, and full fine-tuning across three runs. \sys{} has 1.81$\times$ average improvement in training time when only using the freezing module, and  1.95$\times$ when also using caching. }
    \label{fig:squad_res}
\end{figure*}

\noindent {\bf Accuracy/F1: } 
In left side of Figures~\ref{fig:ag_res}, \ref{fig:sogou_res}, \ref{fig:imdb_res}, \ref{fig:yelp_res}, \ref{fig:swag_res}, and \ref{fig:squad_res} we plot the mean and the range (max,min) of accuracy values obtained by \sys{} as compared to the baseline.
We see that the ranges for \sys{} overlap with the full fine-tuning line indicating that \sys{} is able to achieve comparable accuracy/F1.  Similar to prior work~\cite{qiao2018a}, we also list the best accuracy/F1 reached across trials for the baseline and \sys{} on the right side of Figures~\ref{fig:sogou_res},~\ref{fig:ag_res},~\ref{fig:imdb_res}, ~\ref{fig:yelp_res}, \ref{fig:squad_res}, and \ref{fig:swag_res}. We include complete numbers in the Appendix in Table~\ref{tab:accuracy_num}. 

From the figures we see that for the  Sogou and IMDb datasets, we observe 0.07\% and 0.1\% reduction in mean of the best accuracy, while for AG News and Yelp F. datasets, we do not see any accuracy loss. From Figure ~\ref{fig:squad_res}, we see an loss of 0.11 in average F1 score for SQuAD v2.0 across three runs. As for SWAG, we observe an accuracy loss of 0.01\% in average accuracy.


\noindent {\bf Training Speedup: } As shown in Figure~\ref{fig:ag_res}, Figure~\ref{fig:sogou_res}, Figure~\ref{fig:imdb_res}, Figure~\ref{fig:yelp_res}, Figure~\ref{fig:swag_res}, and Figure~\ref{fig:squad_res}, across all tasks for three independent runs we are able to achieve an average speedup between 1.55$\times$-2.05$\times$ with respect to full fine-tuning of $BERT_{BASE}$. We observe \sys{} is particularly helpful on large datasets, on AG's News dataset, a large dataset with 120K samples, \sys{} is able to save around 5 hours fine-tuning time. For Yelp F. an even larger dataset with 650K data points, we are able to significantly reduce the fine-tuning time by around 25 hours.


\noindent\textbf{Caching Benefits}
Next, we evaluate the speedup gains achieved by switching on both the freezing and caching modules. 
We see an average speedup of 2.08$\times$ across all evaluated tasks compared to the full fine-tuning, which is a 1.14$\times$ improvement with respect to average speedup when compared to only using the Freezing module. By enabling caching, we are able to achieve upto 1.25$\times$ additional speedup compared to freezing. Generally, we obtain more speedup starting from the third epoch as we start to save the forward pass computation for the frozen layers. For fine-tuning workloads that run for more epochs, the benefits of caching will be more pronounced as shown in our technical report~\cite{liu2021autofreeze}.

\begin{figure*}[ht]
    \centering
    \begin{subfigure}[b]{0.38\textwidth}
    \includegraphics[width=\textwidth]{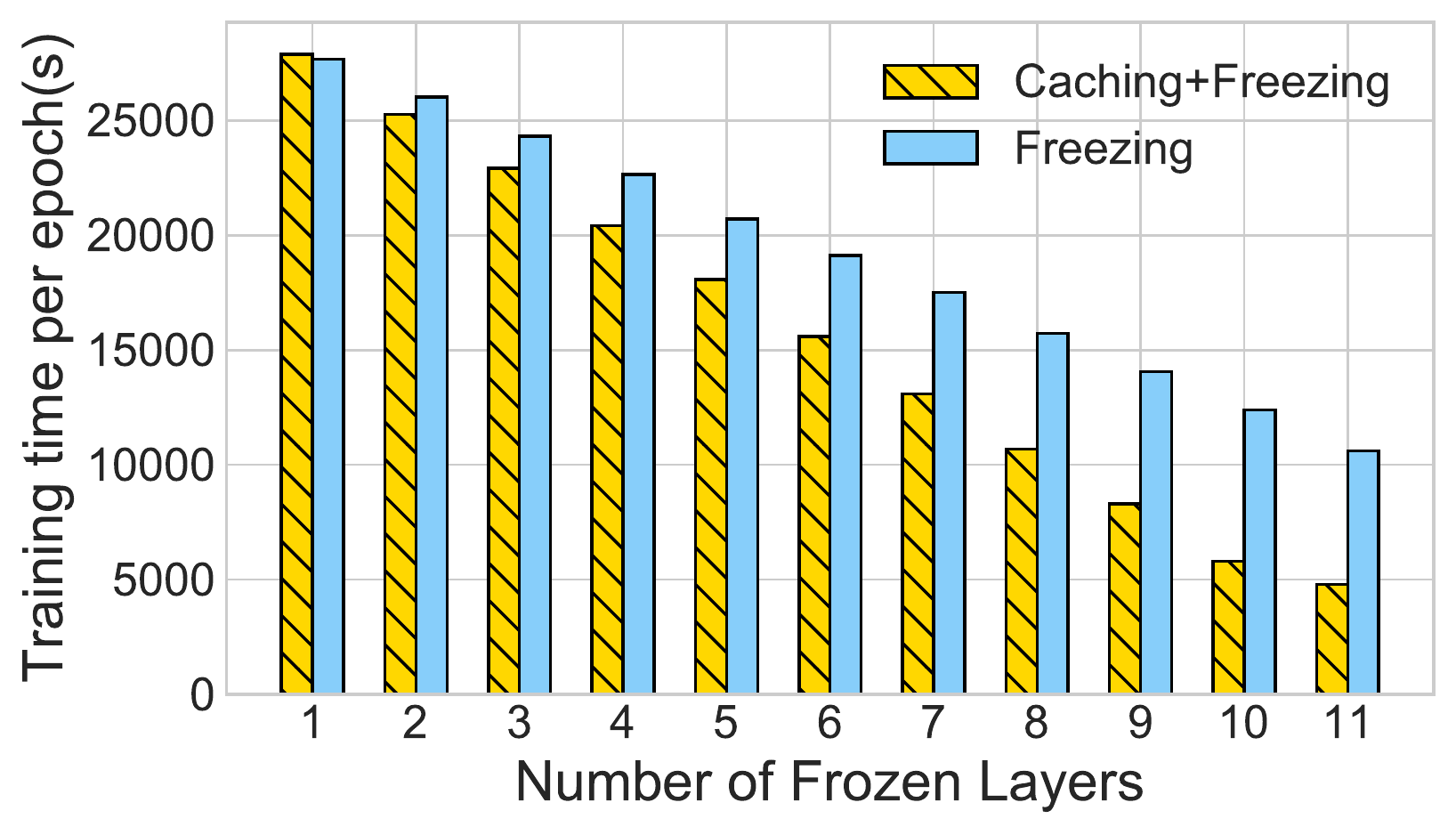}
    \subcaption{[Yelp F.] We only cache 420K (out of 650K) data points as we reach our disk capacity. }
    \label{fig:yelp_f_tradeoff}
    \end{subfigure}
    \hspace{0.1in}
    \begin{subfigure}[b]{0.38\textwidth}
    \includegraphics[width=\textwidth]{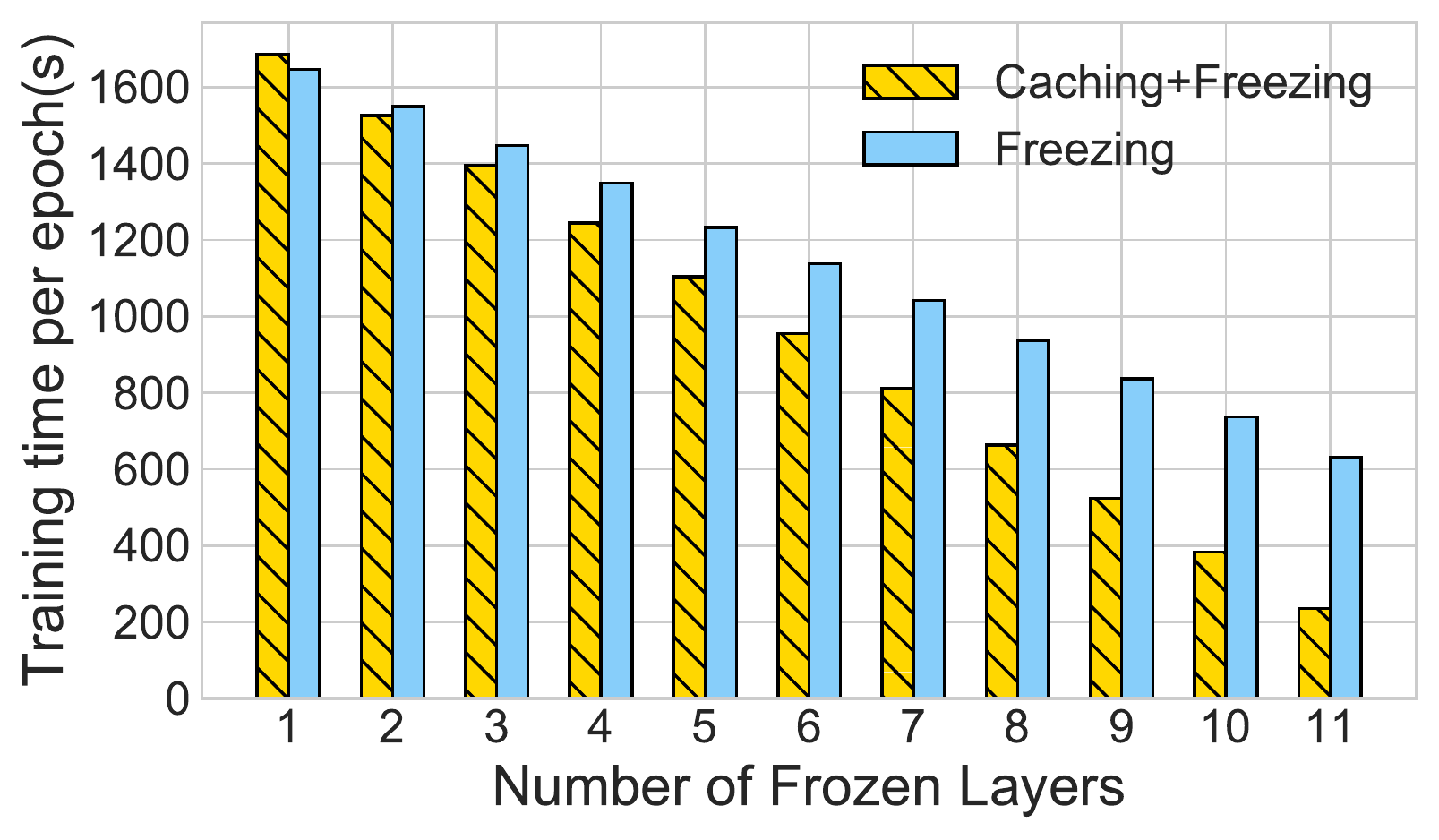}
    \subcaption{[IMDb] We cache the whole dataset, containing 25K points, to CPU memory.}
    \label{fig:imdb_tradeoff}
    \end{subfigure}
    \label{fig:imgdb_time_25}
    \vspace{-0.1in}
    \caption{Trade-off between caching and running the full forward pass as we vary the number of frozen layers. }
    \vspace{0.05in}
\end{figure*}

\noindent\textbf{Caching vs Computation trade-off}
As described in Section~\ref{sec:storage_manager}, when $L$ layers are frozen, the training process consumes data at a rate that corresponds to performing forward and backward computation after layer $L$ while the input reading process operates in parallel to fetch data for the next batch from the cache. Additionally, training process also needs to move data that needs to be written out from GPU to CPU and our measurements show that this adds at most 7\% runtime overhead. Thus the balance between caching vs. redoing computation depends on the dataset size, number of layers frozen and computation speed, our storage manager automatically keeps track of these and chooses the best setup. For \textit{e.g.}, as shown in Figure~\ref{fig:yelp_f_tradeoff}, the overhead from caching fails to be balanced out by the computational savings of skipping part of the forward pass if only the first layer of BERT encoder is frozen. As a result, if the freezing module decides to freeze only the first layer, the storage manager does not activate the caching module.

\subsection{Distributed Fine-tuning}
\label{sec:dist_eval}

In Section~\ref{sec:evaluation_freezing} we show that \sys{} achieves
significant speedups with minuscule degradation in model accuracy on a single GPU. We next evaluate the benefits of \sys{} when performing distributed fine-tuning.\\
\noindent{\bf Setup:} For running all our distributed experiments, we used \textit{p3.2xlarge} instances on AWS. 
We evaluate \sys{} in both Performance Packing and Efficiency Packing modes (Section~\ref{sec:distr_description}) on up to 64 GPUs. We observed that the network bandwidth on AWS suffers from random bursts. To minimize the effects of these bursts we used Wondershaper~\cite{wondershaper} and set the bandwidth to 2Gbps (close to the steady bandwidth).

\begin{figure*}[t!]

 	\begin{subfigure}[t]{.3\linewidth}%
     	\includegraphics[width=\linewidth]{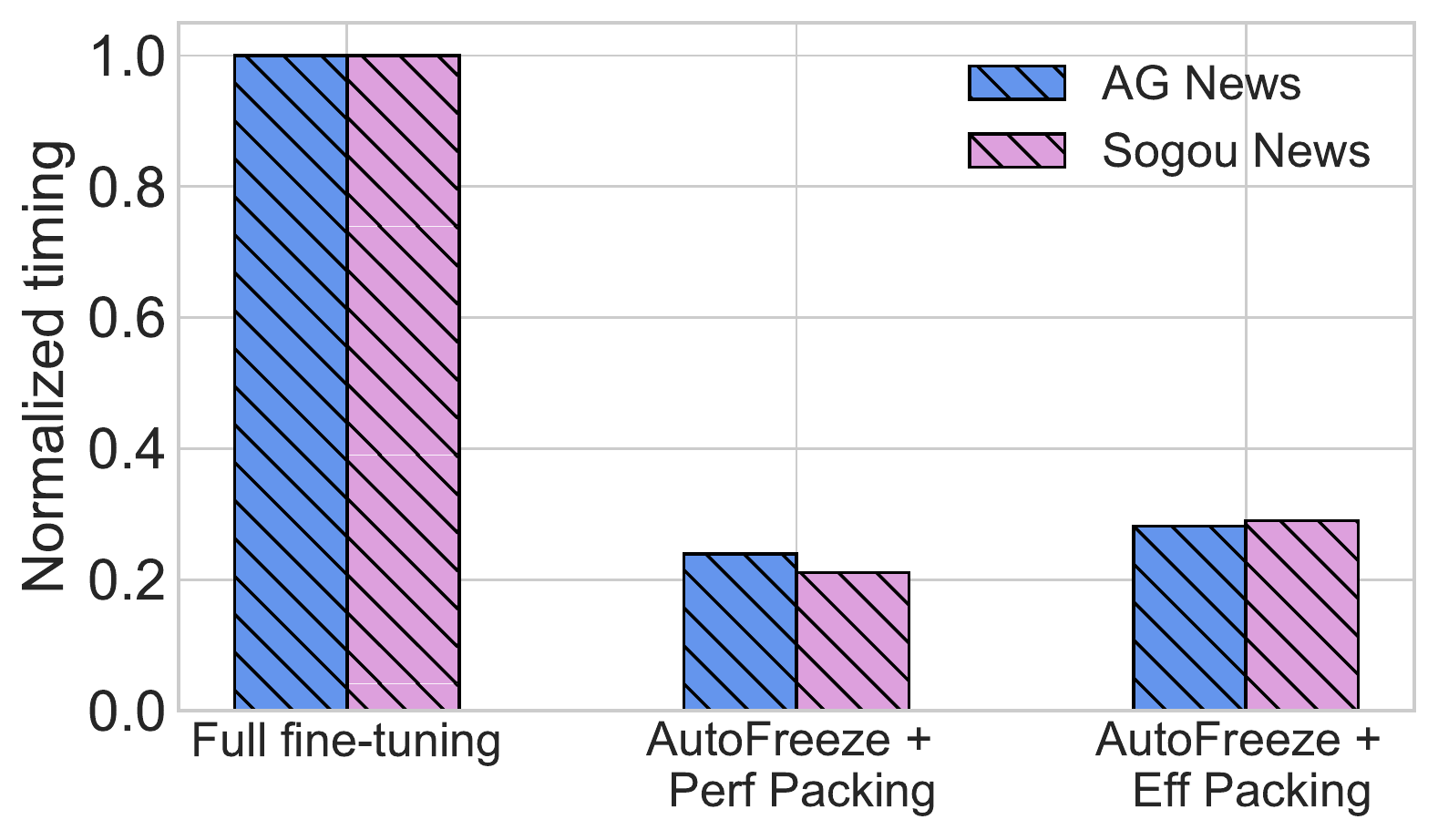}
     	\caption{End-to-end normalized timing for full fine-tuning, \sys{} with performance packing, \sys{} with efficiency packing. }
     	\label{fig:distr_timing}
    
 	\end{subfigure}
 	\hspace{0.1in}
 	\begin{subfigure}[t]{.3\linewidth}%
 	\includegraphics[width=\linewidth]{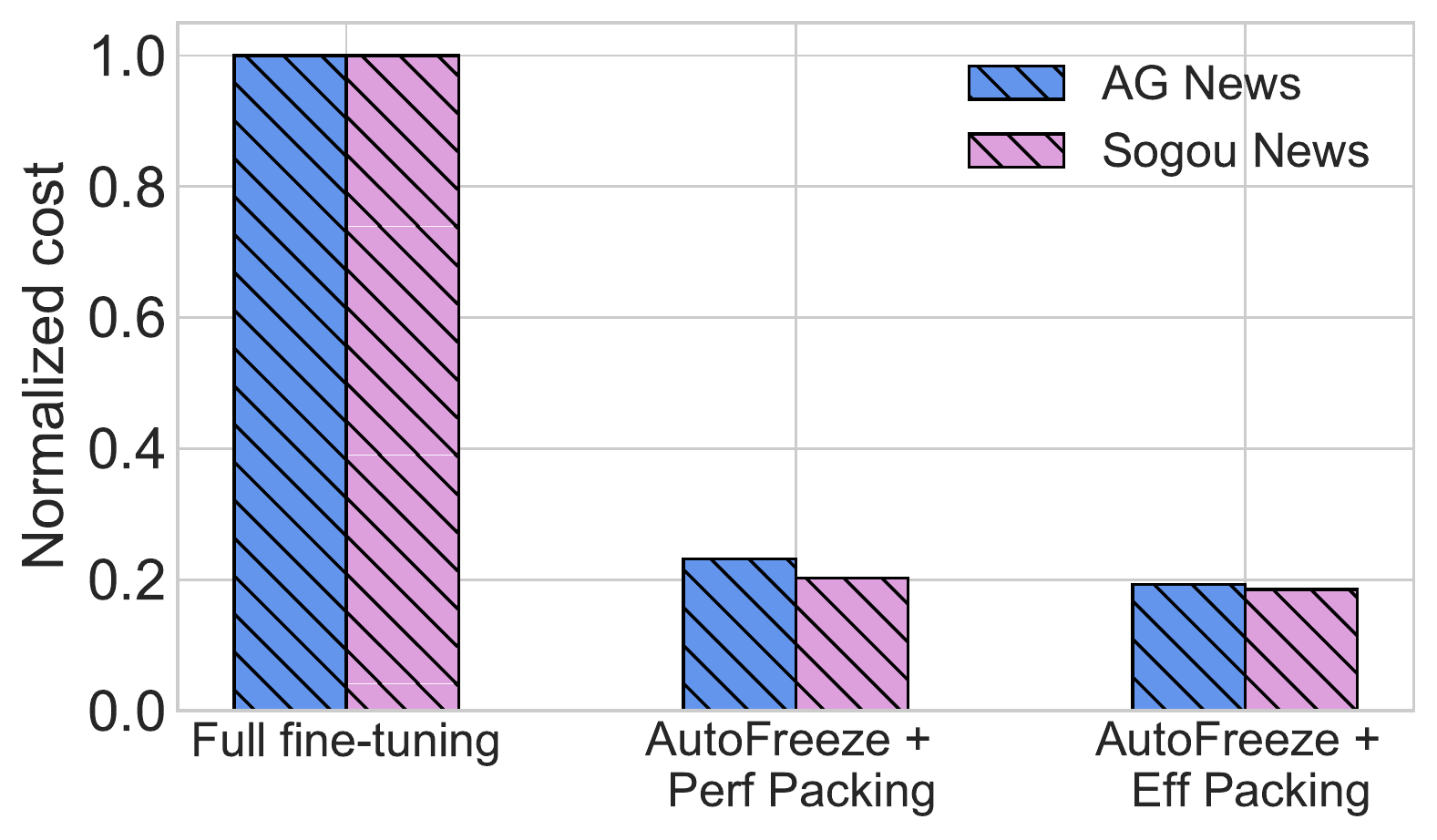}
 	\caption{End-to-end normalized total cost for full fine-tuning, \sys{} with performance packing, \sys{} with efficiency packing.}
 	\label{fig:distr_cost}

 	\end{subfigure}
 	 	\hspace{0.1in}
 	\begin{subfigure}[t]{.3\linewidth}%
 	\includegraphics[width=\linewidth]{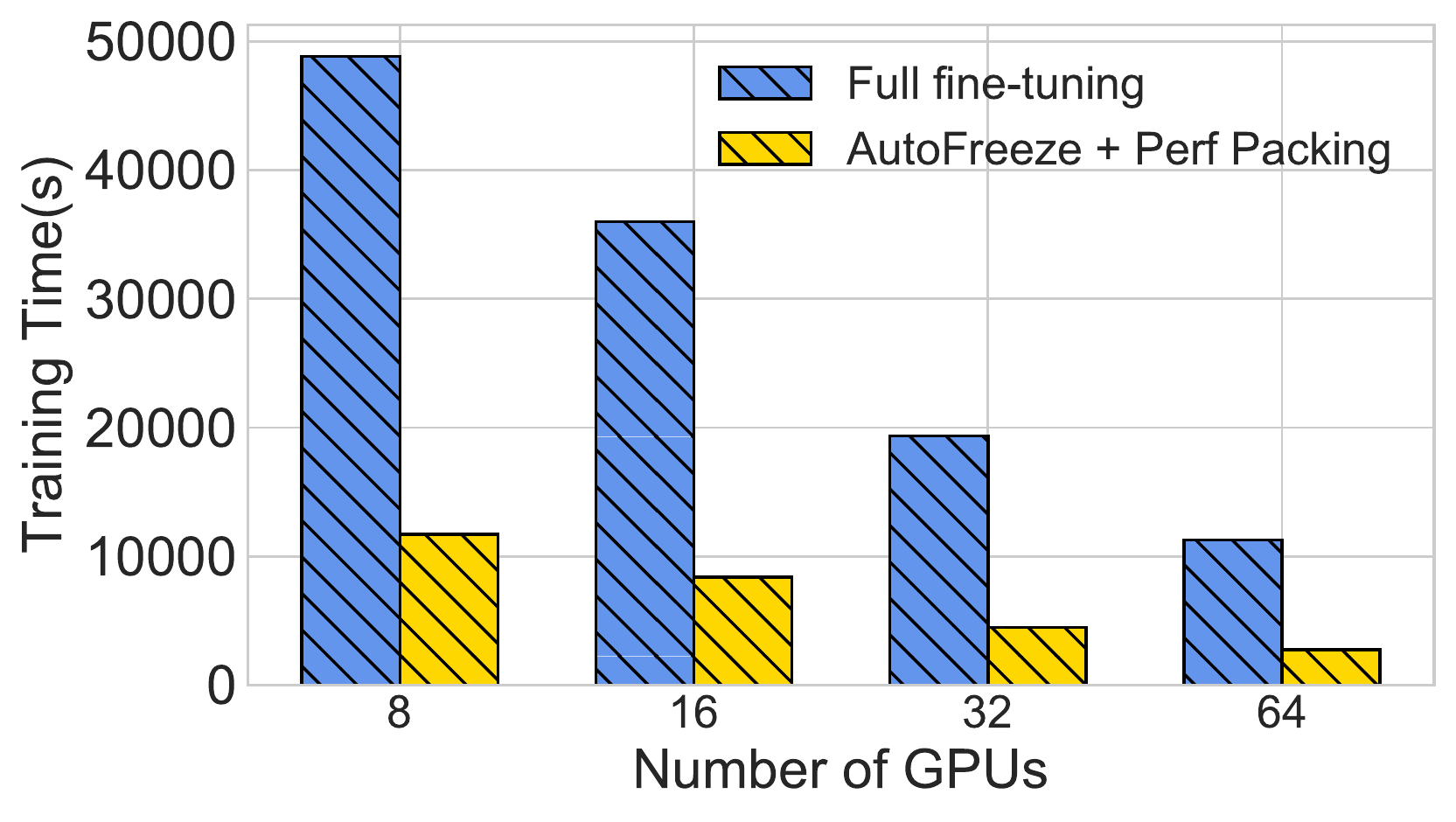}
 	\caption{[AG's News] End-to-end training time for a fixed freezing pattern over fine-tuning for 4 epochs using 8, 16, 32, and 64 GPUs.}
 	\label{fig:scalability_distr}
 	\end{subfigure}
 	\hspace{0.1in}
 	\caption{}
 \label{fig:elastic_gpu}
\end{figure*}

\noindent{\bf Results:}  In Figure~\ref{fig:distr_timing}, we show that Performance Packing is able to reduce the end-to-end training time by 4.38$\times$ for AG's News and 4.74$\times$ for Sogou News datasets when running on 64 GPUs. Efficiency Packing can reduce the end-to-end training time by 3.55$\times$ for AG's News and 3.44$\times$ for Sogou News when compared to performing full fine tuning. As for the total cost, Efficiency Packing reduces the total cost by 5.03$\times$ and 5.21$\times$ for AG's News and Sogou News datasets, while Performance Packing reduces the total cost by 4.38$\times$ and 4.74$\times$ for AG's News and Sogou News datasets when compared against full fine tuning. 

Breaking down the benefits, when we freeze 11 BERT layers for an epoch with the AG's News dataset, the average iteration time for Performance Packing with 64 machines is 1.05 seconds and there are 35 mini-batches in an epoch. On the other hand, the average iteration time for Efficiency Packing on 8 machines is 0.42 seconds but runs 313 mini-batches in an epoch.

We also measured the accuracy for both Efficiency Packing and Performance Packing and compare them to full fine-tuning without changing number of GPUs or per GPU batch size on AG's News and Sogou datasets (Table~\ref{tab:distr_acc}). \sys{} with Efficiency Packing results in negligible or no accuracy loss compared to full fine-tuning, while \sys{} with Performance Packing can result in at most 0.5\% accuracy loss. 
Performance Packing is more prone to accuracy loss as adaptively increasing batch size on freezing of layers can lead to extremely large batches which has been shown to cause accuracy drops~\cite{you2019large}. As mentioned in Section~\ref{sec:distr_description}, users of \sys{} can configure a maximum batch size to avoid accuracy drops. 

Finally, we also study how the benefits of \sys{} changes as we scale from 8 to 64 GPUs. As shown in Figure~\ref{fig:scalability_distr}, we can see that increasing the number of GPUs used in fine-tuning can lead to almost linear decrease in end-to-end fine-tuning time. We also observe that \sys{} can achieve similar speedup (varying from 4.10$\times$-4.31$\times$) across different number of GPUs when compared to full fine-tuning.

\subsection{Ablation Studies}
We next study how the benefits from \sys{} change as we consider more complex training tasks and newer hardware. We also study the sensitivity of \sys{} to various parameters and the overheads from the gradient norm test.

\begin{figure*}[t!]

 	\begin{minipage}[t]{.3\linewidth}%
 	\center
 	\includegraphics[width=\linewidth]{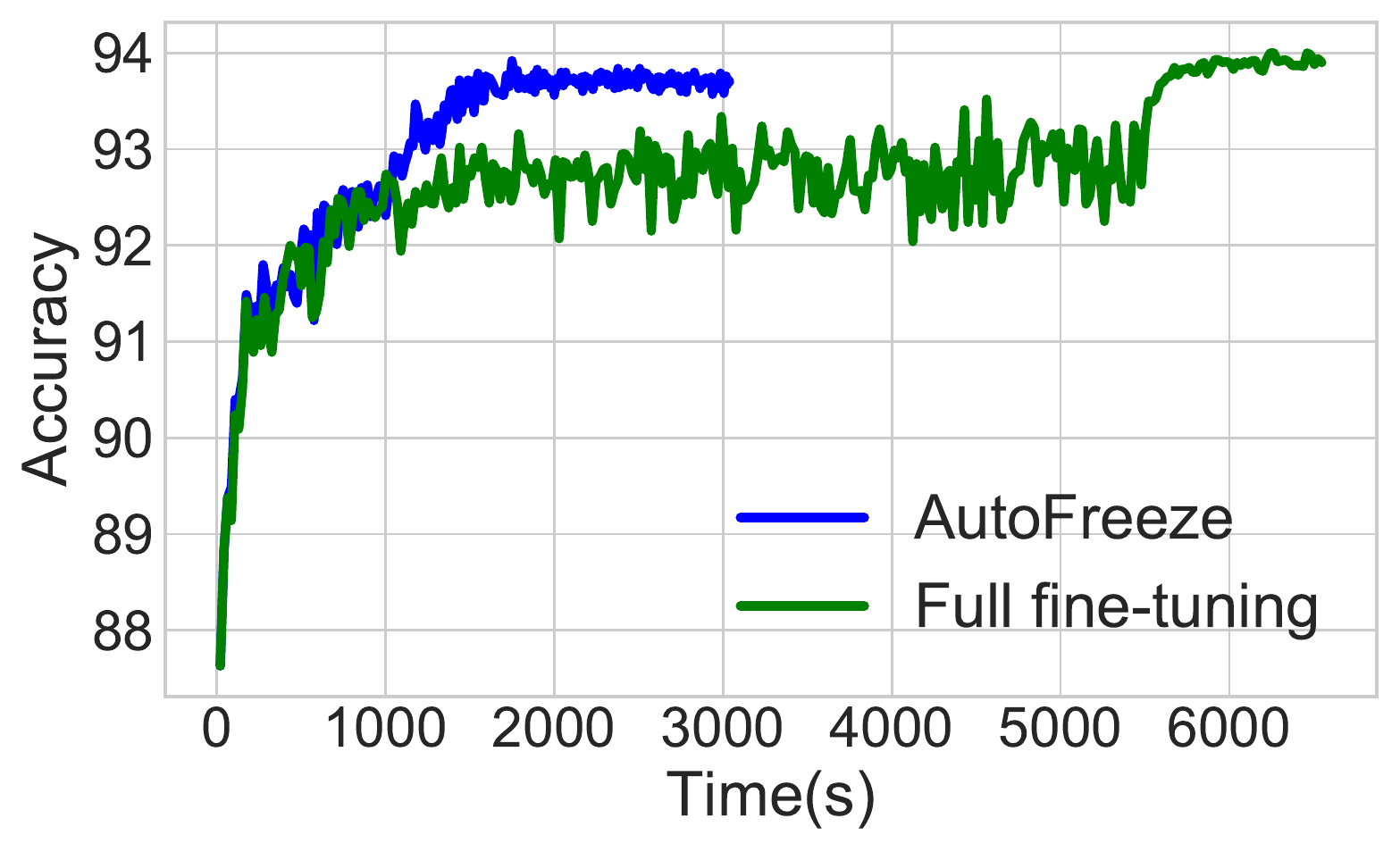}
 	\vspace{-0.25in}
 	\caption{\textbf{[Vision Models]} Performance of full fine tuning and \sys{} when fine tuning ResNet-18 trained on CINIC-10 with CIFAR-10 dataset. \sys{} achieves similar accuracy while being 2.15$\times$ faster.}
 	\label{fig:cinic_res}
 	\end{minipage}
 	\hspace{.1in}
 	\begin{minipage}[t]{.3\linewidth}%
 	\center
 	\includegraphics[width=0.9\linewidth]{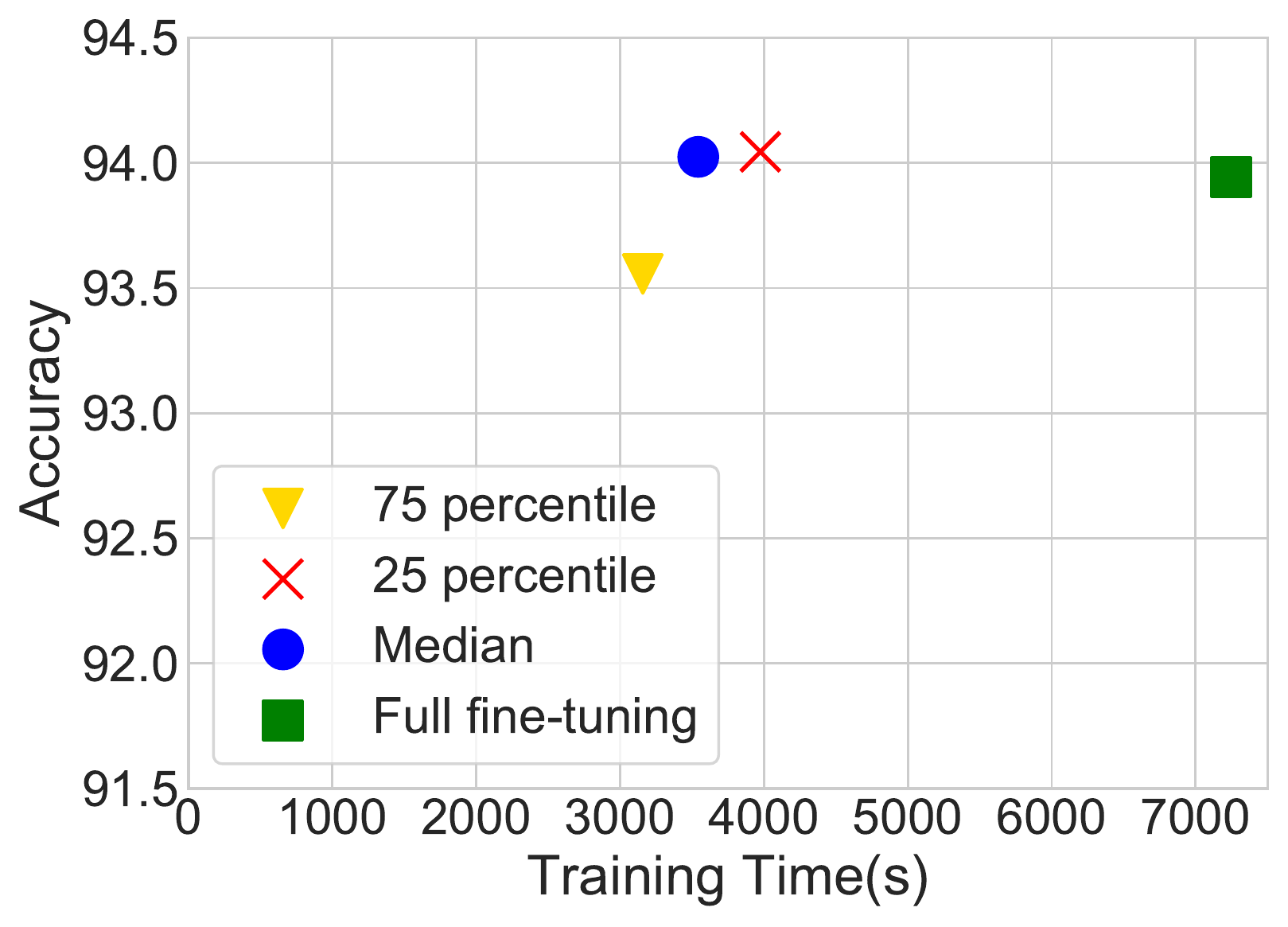}
 	\vspace{-0.1in}
 	\caption{[\textbf{IMDb}] Comparison between freezing schemes when using $75^{th}$ percentile, $25^{th}$ percentile and median in Algorithm~\ref{alg:freezing}.  }
 	\label{fig:pc_compare}
 	\end{minipage}
 	 	\hspace{.1in}
 	\begin{minipage}[t]{.3\linewidth}%
 	\center
 	\includegraphics[width=0.9\linewidth]{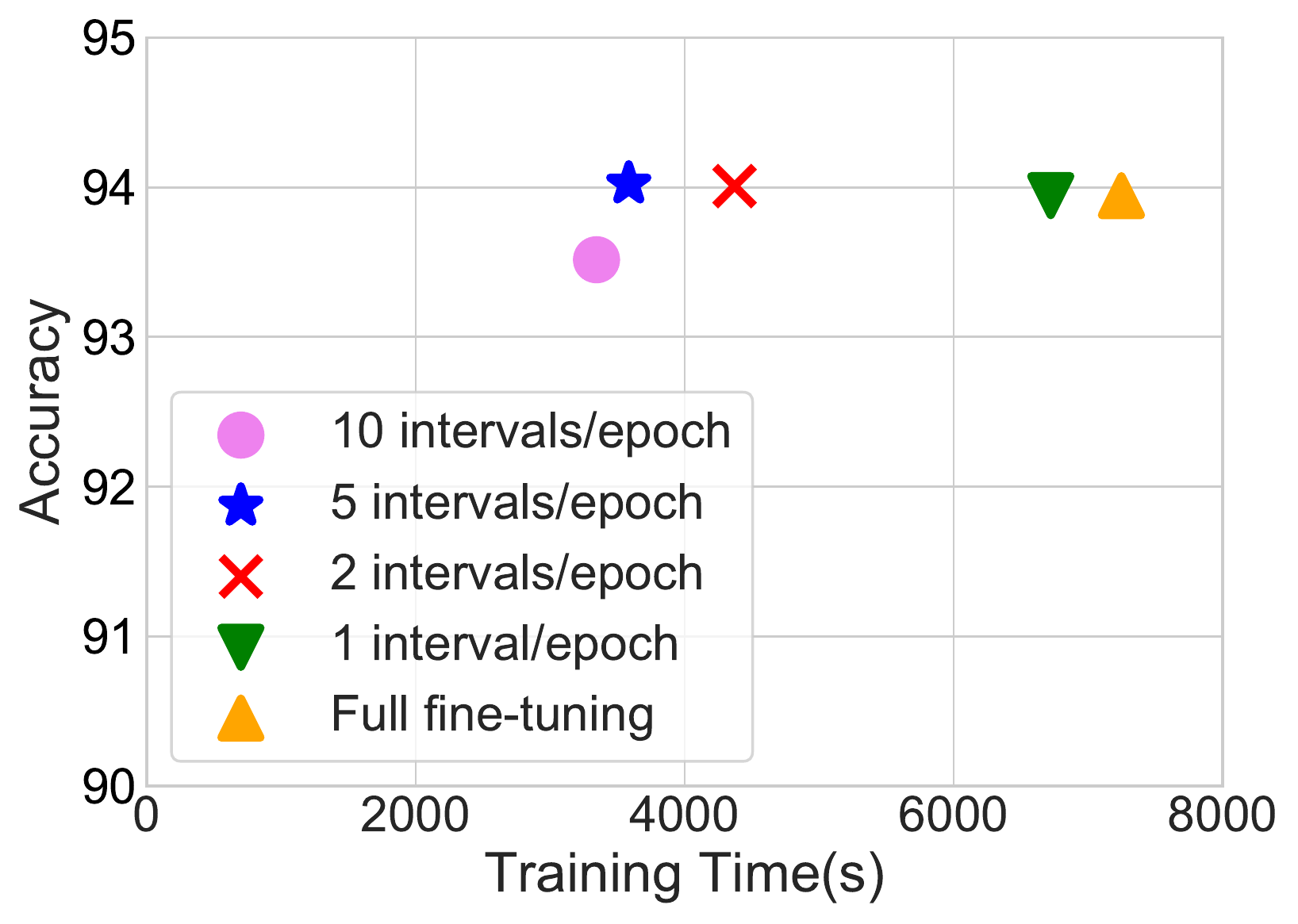}
 	\vspace{-0.1in}
 	\caption{[\textbf{IMDb}] Comparison between freezing schemes by varying number of evaluation intervals for each epoch from 1 interval/epoch to 10 intervals/epoch.}
 	\label{fig:interval_compare}
 	\end{minipage}
 	 	\hspace{.1in}
 	\vspace{-0.05in}
\end{figure*}

\begin{table*}[t]
\centering
\begin{tabular}{@{}llll@{}}
\toprule
           & Full fine-tuning & Eff packing & Perf packing \\ \midrule
AG's News  & 94.43            & 94.54                    & 94.42                     \\
Sogou News & 97.43            & 97.23                    & 96.9                      \\ \bottomrule
\end{tabular}
\caption{[64 GPUs] Accuracy for full fine-tuning, \sys{} with Efficiency Packing, and \sys{} with Performance Packing.}

\label{tab:distr_acc}
\end{table*}

\begin{table*}
\centering
\begin{tabular}{@{}lllll@{}}
\toprule
\multicolumn{5}{l}{\hfil Text Summarization}                                                            \\ \midrule
           & \hfil R1  & \hfil R2  & \hfil RL    & \begin{tabular}[c]{@{}l@{}}Training \\ Time (s)\end{tabular} \\ \midrule
AutoFreeze & 41.31 & 18.9 & 38.46 & 116528                                                       \\ \midrule
Full fine-tuning  & 41.45 & 19.02 & 38.55 & 180450                                                       \\ \bottomrule
\end{tabular}
\caption{\textbf{[Text Summarization]} ROUGE F score results on CNN/DailyMail test set (R1 and R2 stand for unigram and bigram overlap; RL, the longest common subsequence). 
We took top-3 checkpoints based on the performance on the validation set and report the average results on the test set. }
\label{tab:text_summ}
\end{table*}


\noindent{\bf \sys{} on Text summarization: } We test how well \sys{} works on more complex tasks by considering a text summarization problem. We follow the experimental setup and hyper-parameters in~\cite{liu2019text} and  train the \textsc{BertSumAbs} model for 200000 steps on four NVIDIA V100 GPUs. We set the total number of evaluation intervals to 20 and evaluate every 10000 steps during fine-tuning. In Table~\ref{tab:text_summ} we show the Recall-Oriented Understudy for Gisting Evaluation (ROUGE) F score, which is a measurement for the similarity between a candidate document and reference documents. As shown in Table~\ref{tab:text_summ}  we observe negligible loss (less than 0.15) in ROGUE F1 score when comparing \sys{} and full fine-tuning. Further, \sys{} is able to achieve a speedup of 1.55$\times$ comparing to full fine-tuning. 

\noindent {\bf CINIC-10 transfer learning: }
Although BERT fine tuning is our primary focus, we also evaluate \sys{} on the vision task described in Section~\ref{sec:motivation}. We take a ResNet-18 model which is pre-trained with CINIC-10 and fine-tune this model for the CIFAR-10 dataset. In Figure~\ref{fig:cinic_res} we observe that our freezing scheme is able to reduce the fine-tuning time by more than $2\times$ and still reaches a very similar accuracy as of full fine tuning. This demonstrates that our freezing scheme is applicable on tasks other than BERT and we plan to explore these domains in detail the future.

\noindent {\bf Using A100 GPUs: }
To show \sys{} can provide the benefits even on advanced hardware, we tested \sys{} on the latest NVIDIA A100 GPU. We set up the experiments on a p4d.24xlarge instance with eight A100s on AWS, and ran \sys{} on the AG's News dataset. To match the total batch size of 24 \cite{sun2019fine} for classification tasks, we set the per GPU batch size to 3. While full fine-tuning took 3030 seconds for four epochs, \sys{} only took 1670 seconds and reached similar accuracy. Thus, we see a speed-up of 1.81$\times$ on A100s which is comparable with the speedup achieved on P100s and shows that the benefits of \sys{} remain across hardware generations.

Finally, we also ran the experiment to fully utilize the memory of the A100 GPUs by setting the per GPU batch size to 16, and fine-tune $BERT_{LARGE}$ on AG News. \sys{} is able to finish fine-tuning in 1304 seconds compared to 2020 seconds for full fine-tuning (1.54$\times$ speed-up).

\noindent {\bf Using different percentiles: } As shown in Figure \ref{fig:pc_compare}, we observe a trade-off between training speedup gains and model accuracy as we vary the percentile threshold used in Algorithm~\ref{alg:freezing}. We observe, the more aggressive our freezing policy, the greater the accuracy loss will be. For example, if we set $N$ to be 75th percentile which results in a more aggressive policy, we get 0.39\% accuracy loss for the IMDb dataset. However, we do not see accuracy loss when we set $N$ to be 25th or 50th percentile. On the other hand, increasing $N$ gives us more training speedup. A policy with $N=75$ achieves 20\% more speedup compared with the $N=25$ policy with 0.49\% difference in max accuracies achieved. For all experiments in this paper we chose 50th percentile and have not fine tuned this parameter.

\noindent {\bf Varying number of evaluation intervals each epoch: } We also vary how frequently the freezing module is invoked and Figure~\ref{fig:interval_compare} shows the results with the IMDb dataset. We see that if the frequency is too low (e.g., 1 interval/epoch) then the speedup obtained is limited. On the other hand, using 10 intervals/epoch results in gradient vectors that are not fully representative, thus leading to a drop in accuracy. However, we find this trade-off is balanced for a range of values (2 to 5 intervals/epoch).

\noindent {\bf Gradient Norm Test Overheads: } As stated in Section \ref{sec:grad_norm}, the gradient norm test has minimal overhead. The overhead mainly comes from accumulating the gradient vectors within an evaluation interval $T$. For example, for IMDb, the overhead for gradient accumulation in terms of time is 13 seconds for every interval, which is less than 1\% of the execution time of an interval. The memory overhead for storing the gradients is 453MB for an epoch.

\section{Related Work}
Previous works have largely focused on reducing the size of pre-trained models or improving the accuracy/stability for fine-tuning.

\noindent\textbf{Improving Fine-Tuning} Several methods~\cite{miao2020you, murty2020expbert, gururangan2020don} have been developed in NLP literature to achieve good accuracy when fine tuning. ~\cite{howard2018universal} introduce ULFiT, a technique which has enabled state of the art performance when doing fine tuning.  Similarly \cite{sun2019fine} investigates fine-tuning methods of BERT on text classification tasks including layer selection, layerwise learning rate, and multi-task Learning. \cite{sun2019fine} shows lower layers contain more general information, and using features from the last layer of BERT gives the best fine-tuning accuracy. ~\cite{jiang2019smart} proposes a smoothness inducing regularizer for improving robustness during training. Similarly other works have proposed different techniques ~\cite{peters2018deep, houlsby2019parameter, stickland2019bert} aiming to improve accuracy while reducing parameters of the final trained model. In this work, our aim is to speed up time for fine tuning without losing the accuracy gains provided by techniques proposed in~\cite{howard2018universal}. We show that \sys{} achieves the same accuracy as these methods but reduces time for fine-tuning by adaptively freezing layers at run time. \\

\noindent\textbf{Model Compression} The primary goal of several previous works is to reduce the size of the fine tuned model to enable fast inference.~\cite{lan2019albert, ma2019tensorized} perform low rank approximations to reduce the size and compute requirements of the model. ~\cite{sanh2019distilbert, jiao2019tinybert, sun2019patient} use knowledge distillation to train a smaller model.~\cite{yang2019model} uses multiple teacher models to train a student for multi-task learning. Similarly~\cite{qiao2018a, bie2019fully, zafrir2019q8bert} perform quantization to enable fast compression. On the other hand ~\cite{cui2019fine, mccarley2019pruning, michel2019sixteen} reduce the model size using pruning. Similarly~\cite{fan2019reducing} use structured dropout to introduce sparsity.  The goal of these methods is to reduce the time for inference which sometimes lead to more time spent in training. On the other hand in this work we aim to to reduce time for fine tuning of the BERT model for new tasks.\\

\noindent{\textbf{Adaptive ML:}} Another line of work shows that certain layers of network can be skipped dynamically during inference to reduce inference time. ~\cite{xin2020deebert, liu2020fastbert, schwartz2020right} propose techniques for adaptively skipping layers at inference. Another recent work~\cite{agarwal2020accordion} uses gradient norms to adaptively tune communication in distributed learning. On other hand our work is focused on speeding up fine-tuning with adaptive freezing.\\

\noindent\textbf{Speeding up Pre-training} \citet{gong19efficient} present a method to speed up BERT pre-training by progressively increasing the size of the model by stacking layers. \citet{zhang2020accelerating} propose speeding up of BERT pre-training by progressively dropping the layers during pre-training. \citet{chen2020earlybert} introduces EarlyBERT~\cite{chen2020earlybert} which extends the work done on finding lottery-tickets in CNNs~\cite{You2020Drawing} to speedup both pre-training and fine-tuning for BERT models. Experimental evaluation of EarlyBERT~\cite{chen2020earlybert} shows some degradation in accuracy for fine-tuning unlike our work where we have almost no accuracy loss, while providing almost similar speedups. 

\section{Conclusion and Future Work}
Fine-tuning pre-trained models has become a popular and accurate method for developing ML models for new tasks. However there are a number of performance challenges in fine-tuning. In this paper, we proposed \sys{}, a scheme to adapatively freeze parts of the model that are closest to convergence during fine-tuning. We show that using \sys{} on NLP tasks can give up to 2.55x speed up  on a single GPU and 4.38$\times$ in a 64 GPU cluster without affecting accuracy. While this paper mainly focused on BERT due to its popularity, we plan to study if similar approaches also help in other domains like image classification or speech recognition. 
Our implementation is available at \url{https://github.com/uw-mad-dash/AutoFreeze}.

\balance
\clearpage
{
\bibliographystyle{abbrvnat}
\bibliography{example_paper.bib}
}
\newpage
\appendix

\section{Appendix}
\subsection{Complete Results}
\label{sec:complete_results}
Table~\ref{tab:accuracy_num} shows the max accuracy, iterations at which the max accuracy is achieved, and the end-to-end fine-tuning time for \sys{} and full fine-tuning across three random trials for the classification tasks. Table~\ref{tab:squad_summ} shows the max F1, iterations at which the max F1 score is reached, and the end-to-end fine-tuning time for \sys{} and full fine-tuning for SQuAD2.0 dataset. Table~\ref{tab:swag_summ} shows similar measurements for the SWAG dataset. We observe that we gain similar speedup and max accuracy for all the runs for each dataset. 

\begin{figure*}[!t]
    \begin{subfigure}[b]{0.3\textwidth}
    \includegraphics[width=\textwidth]{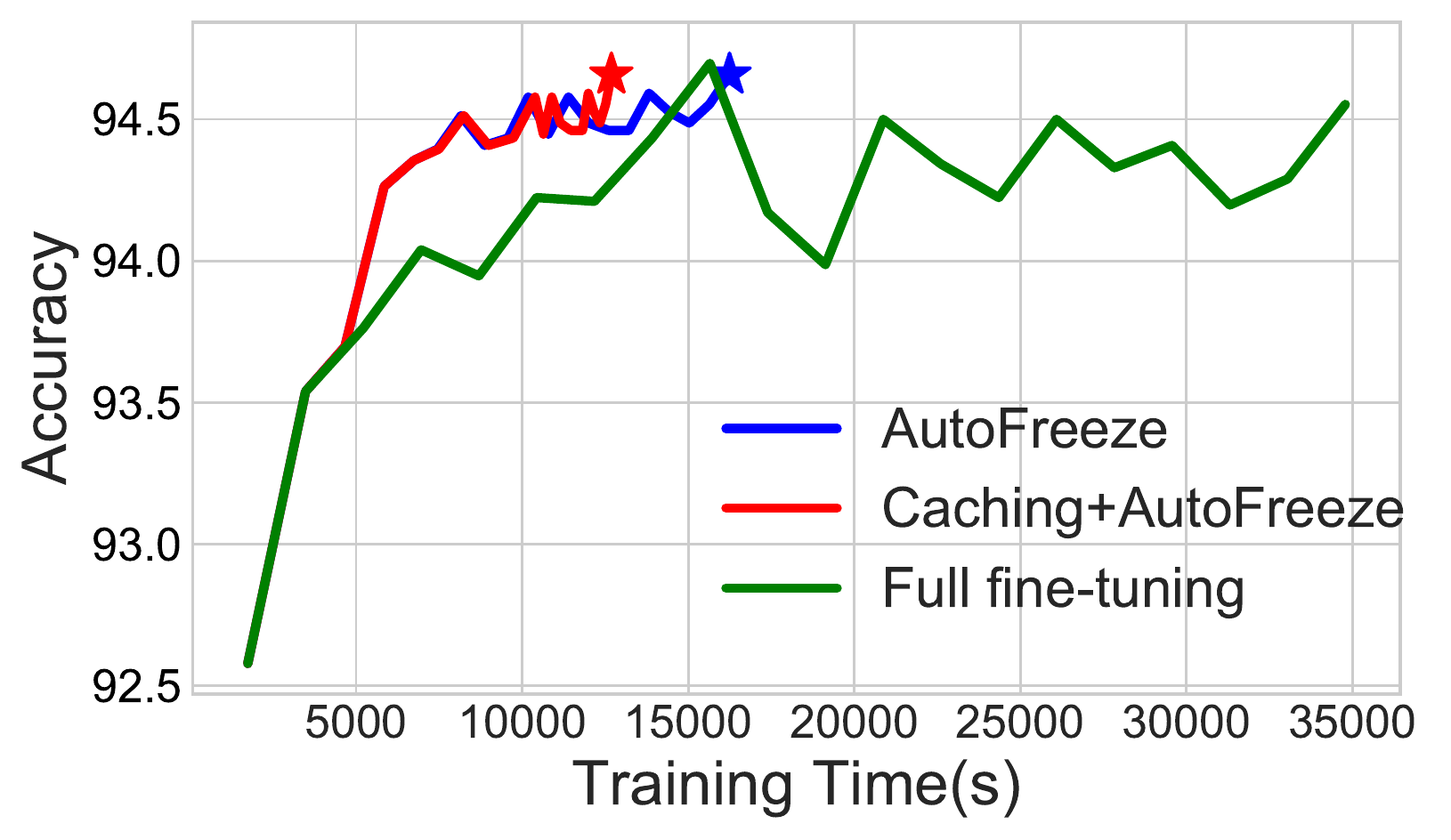}
    \end{subfigure}
    \begin{subfigure}[b]{0.3\textwidth}
    \includegraphics[width=\textwidth]{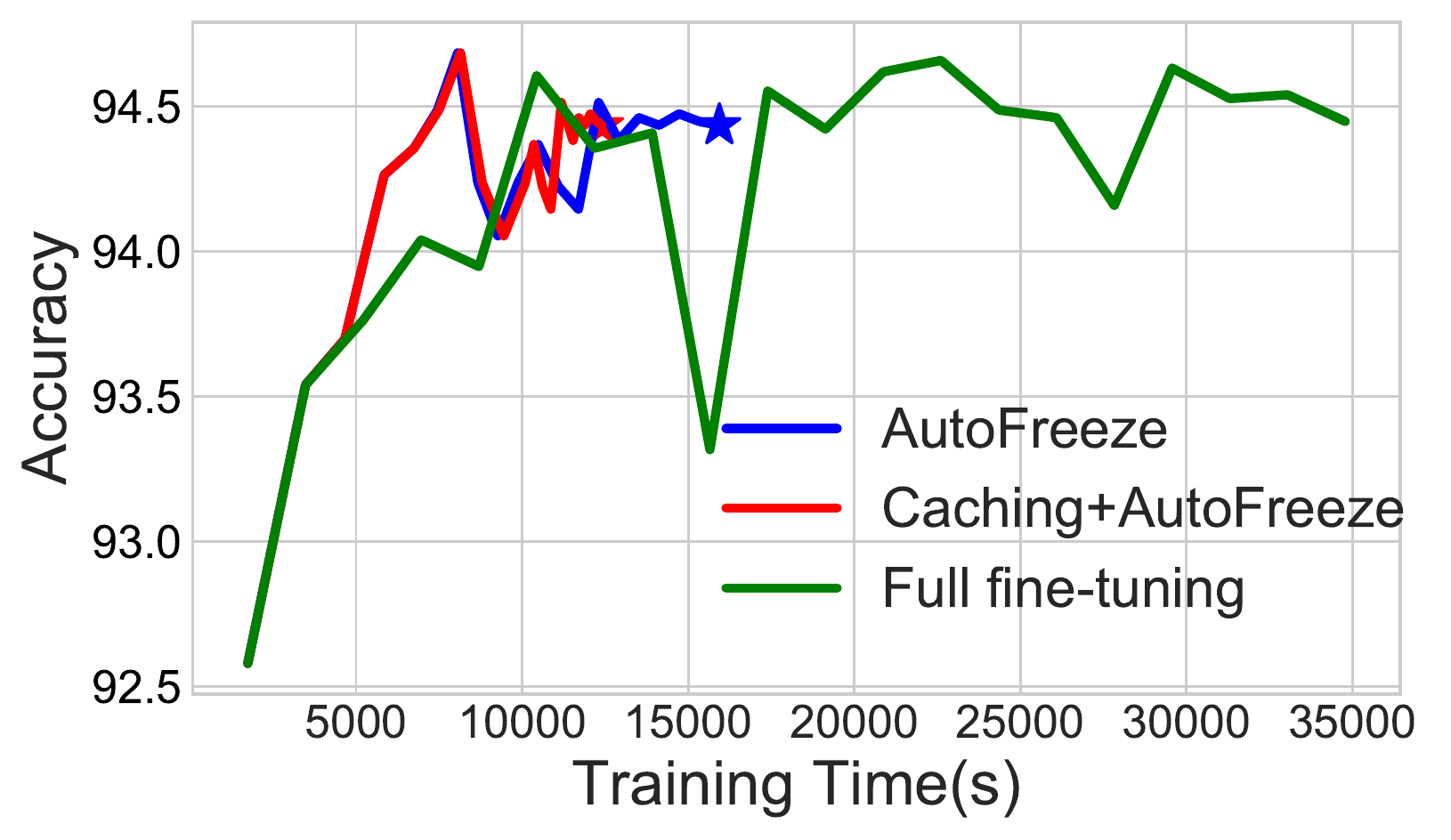}
    \end{subfigure}
    \begin{subfigure}[b]{0.3\textwidth}
    \includegraphics[width=\textwidth]{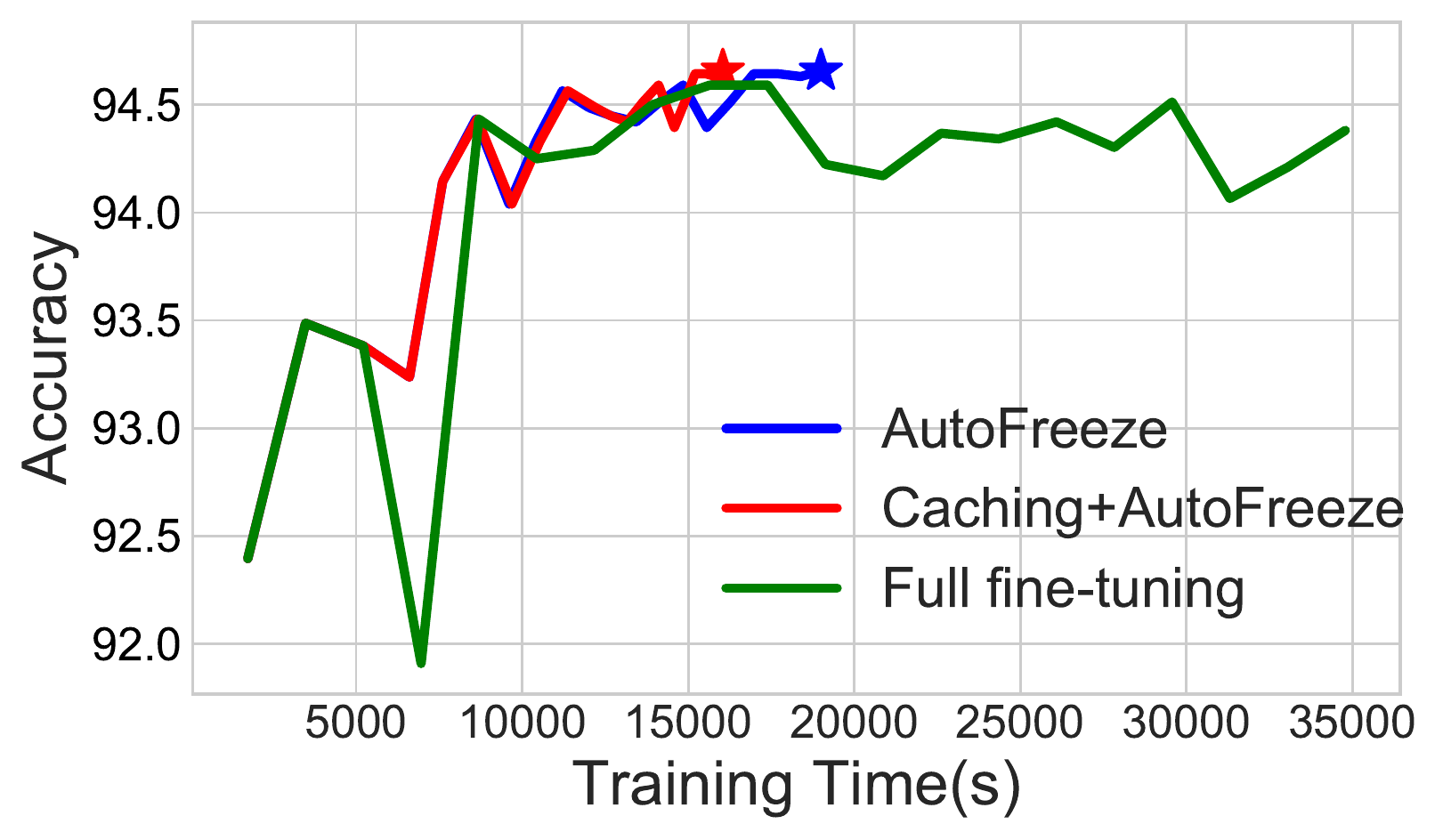}
    \end{subfigure}
    \caption{[\textbf{AG News}] Test accuracy curve for each trial with respect to end-to-end training time for \sys{}, \sys{} with Caching turned on, and full fine-tuning. }
    \label{fig:ag_curve1}
\end{figure*}

\begin{figure*}[!t]
    \begin{subfigure}[b]{0.3\textwidth}
    \includegraphics[width=\textwidth]{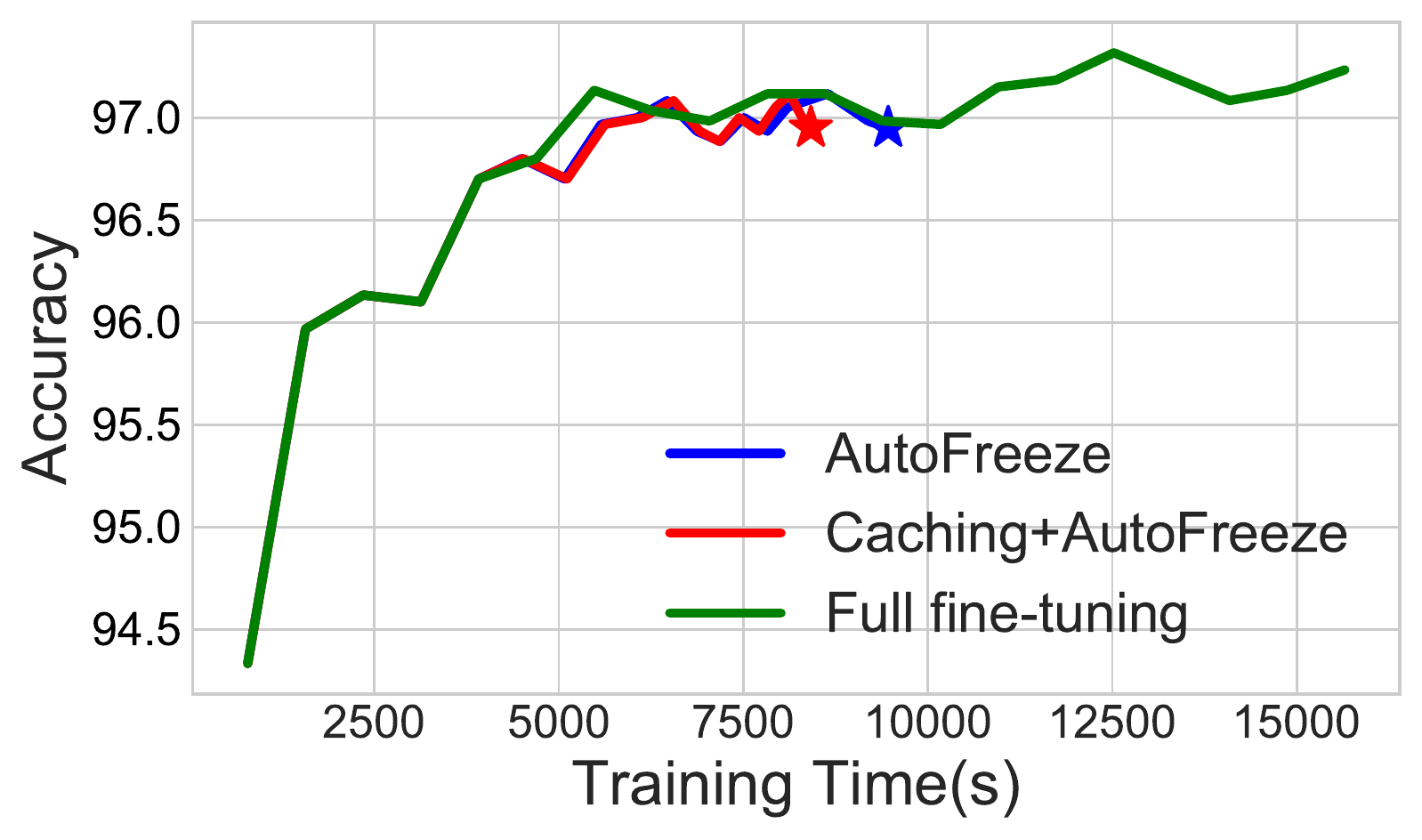}
    \end{subfigure}
    \begin{subfigure}[b]{0.3\textwidth}
    \includegraphics[width=\textwidth]{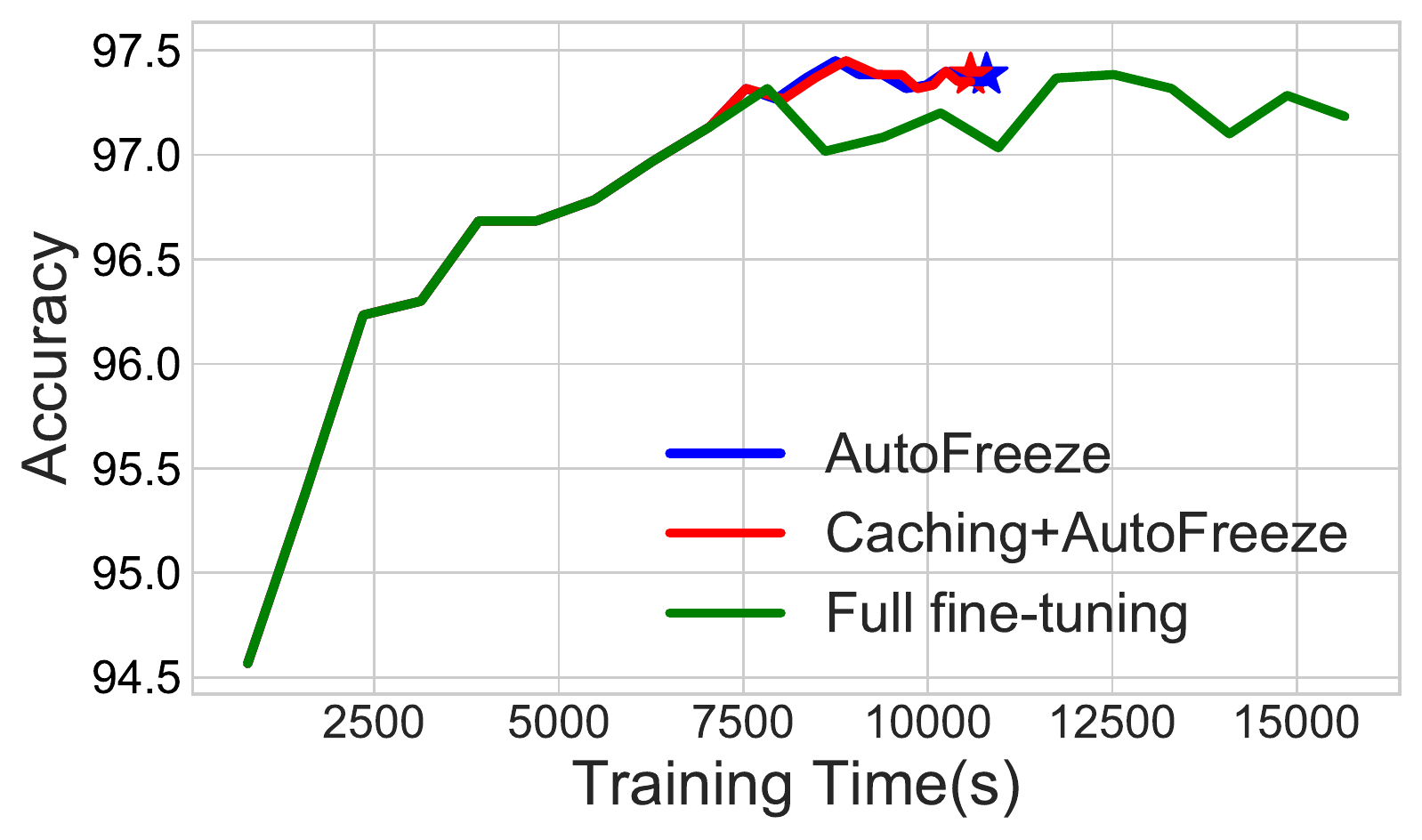}
    \end{subfigure}
        \begin{subfigure}[b]{0.3\textwidth}
    \includegraphics[width=\textwidth]{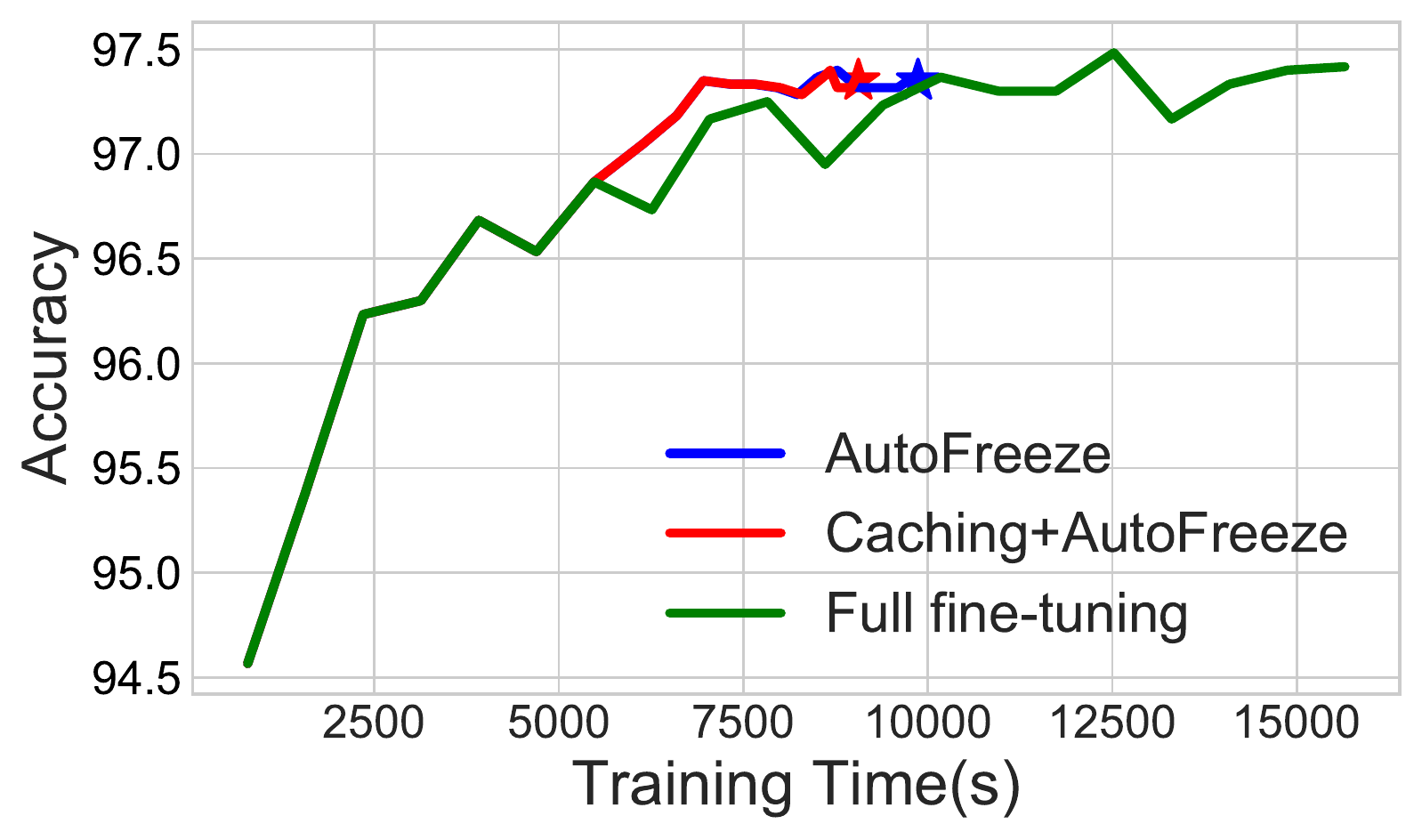}
    \end{subfigure}
    \caption{[\textbf{Sogou News}] Test accuracy curve for each trial with respect to end-to-end training time for \sys{}, \sys{} with Caching turned on, and full fine-tuning. }
     \label{fig:sogou_curve1}
\end{figure*}

\begin{figure*}[!t]
    \centering
    \begin{subfigure}[b]{0.3\textwidth}
    \includegraphics[width=\textwidth]{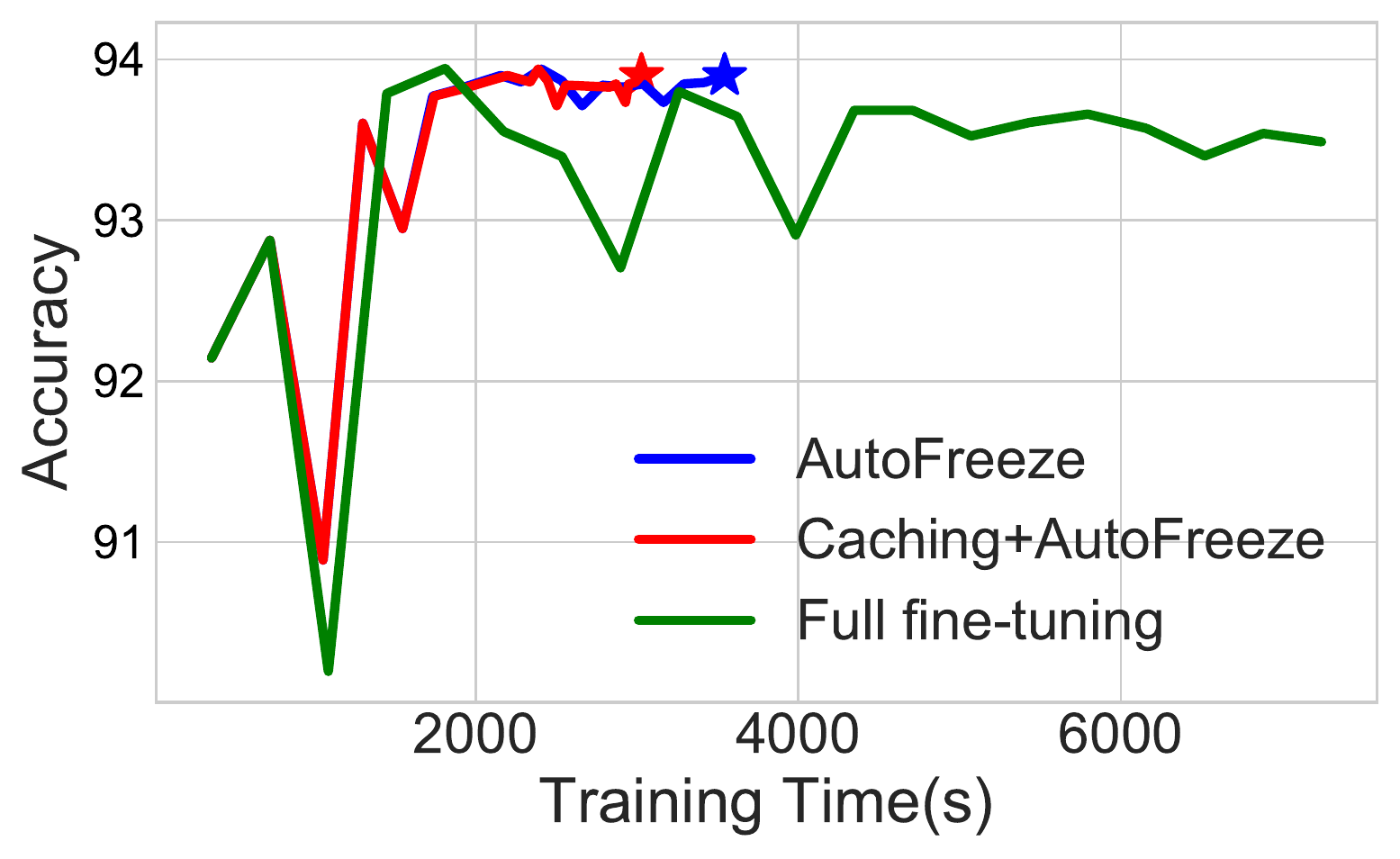}
    \end{subfigure}
    \begin{subfigure}[b]{0.3\textwidth}
    \includegraphics[width=\textwidth]{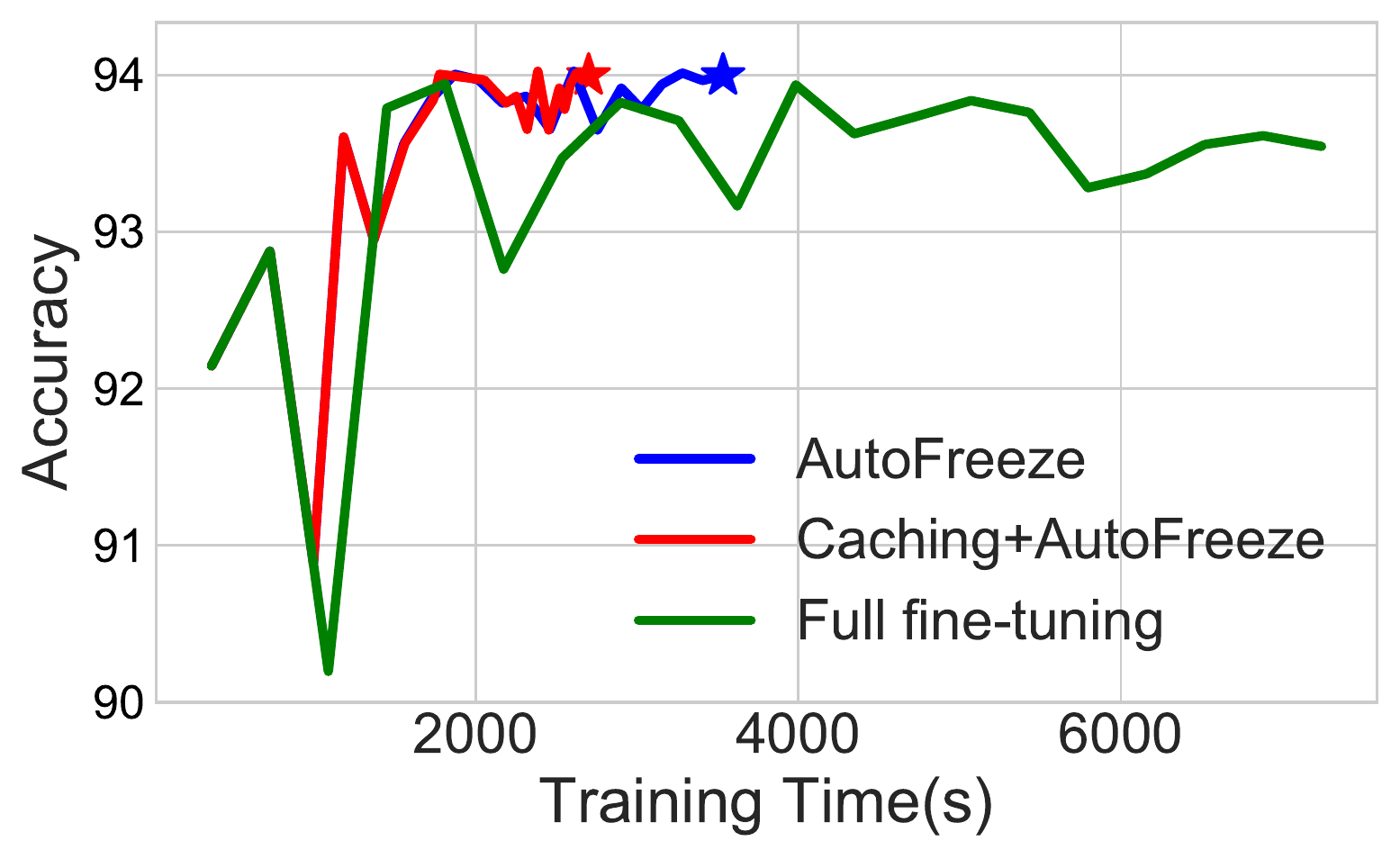}
    \end{subfigure}
        \begin{subfigure}[b]{0.3\textwidth}
    \includegraphics[width=\textwidth]{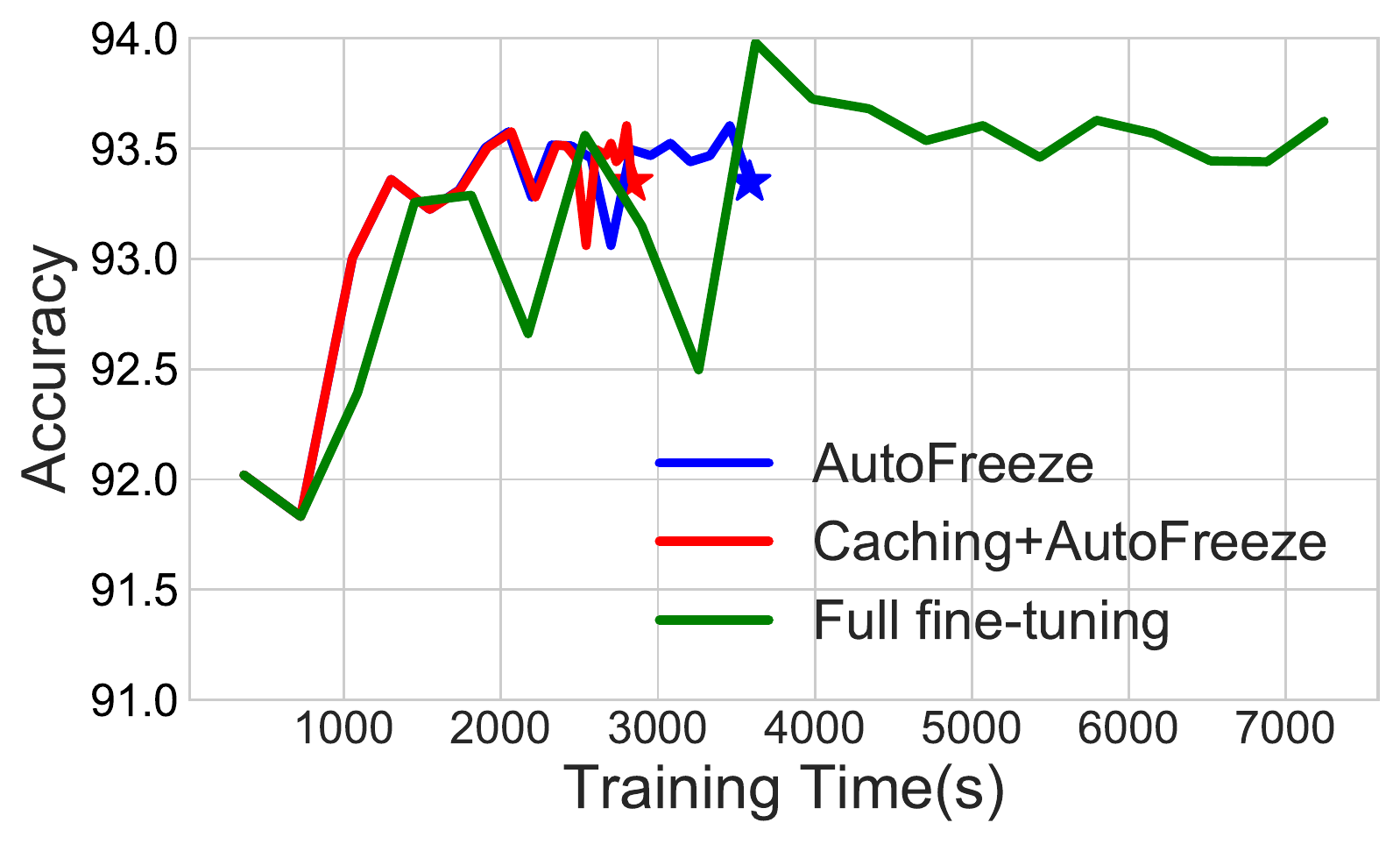}
    \end{subfigure}
    
    \caption{[\textbf{IMDb}] Test accuracy curve for each trial with respect to end-to-end training time for \sys{}, \sys{} with Caching turned on, and full fine-tuning. }
    \label{fig:imdb_curve1}
\end{figure*}

\begin{figure*}[ht]
    \centering
    \begin{subfigure}[b]{0.3\textwidth}
    \includegraphics[width=\textwidth]{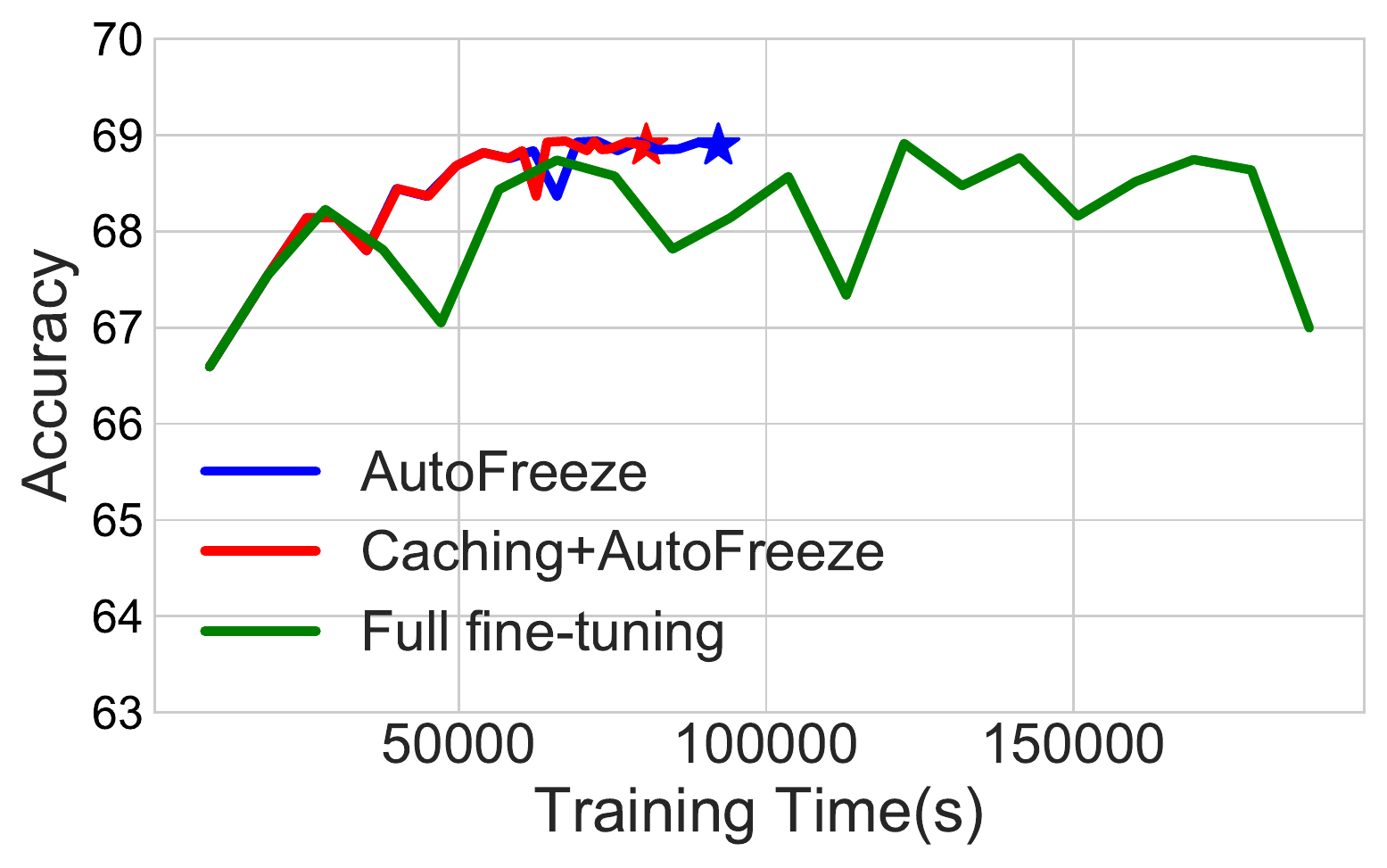}
    \end{subfigure}
    \begin{subfigure}[b]{0.3\textwidth}
    \includegraphics[width=\textwidth]{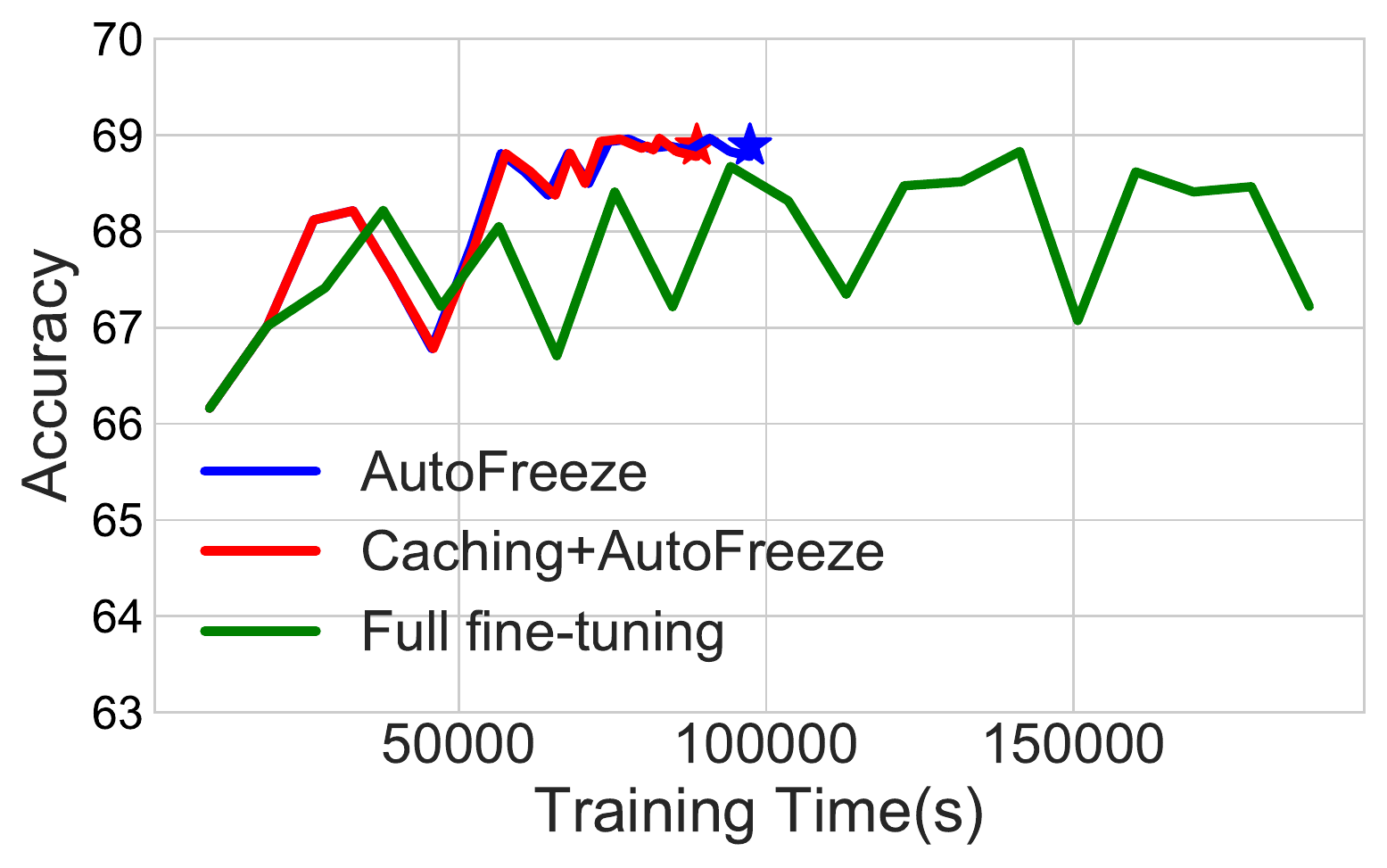}
    \end{subfigure}
    \begin{subfigure}[b]{0.3\textwidth}
    \includegraphics[width=\textwidth]{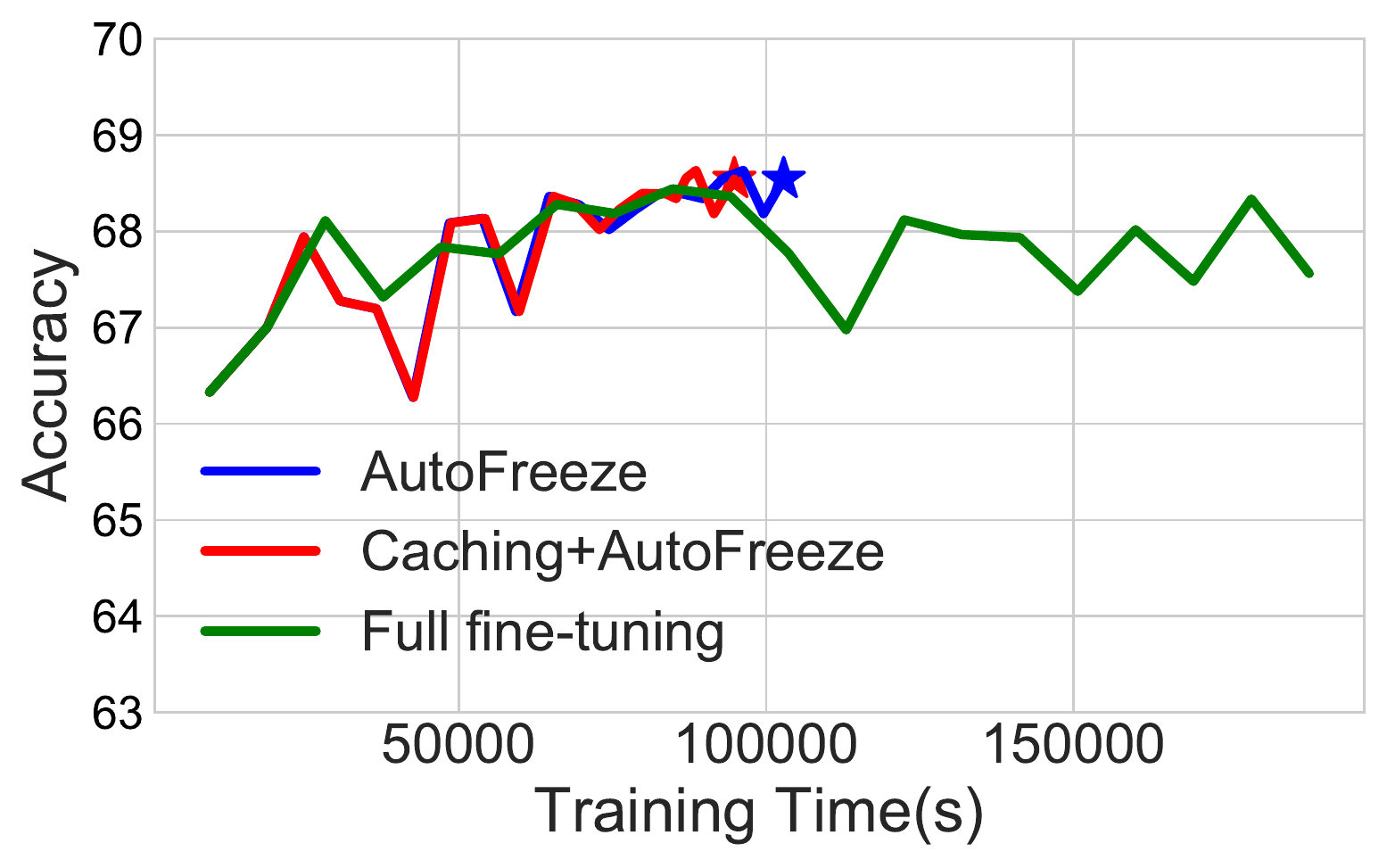}
    \end{subfigure}
    \caption{[\textbf{Yelp F.}] Test accuracy curve for each trial with respect to end-to-end training time for \sys{}, \sys{} with Caching turned on, and full fine-tuning. }
    \label{fig:yelp_curve}
\end{figure*}

\begin{figure*}[ht]
    \begin{subfigure}[b]{0.3\textwidth}
    \includegraphics[width=\textwidth]{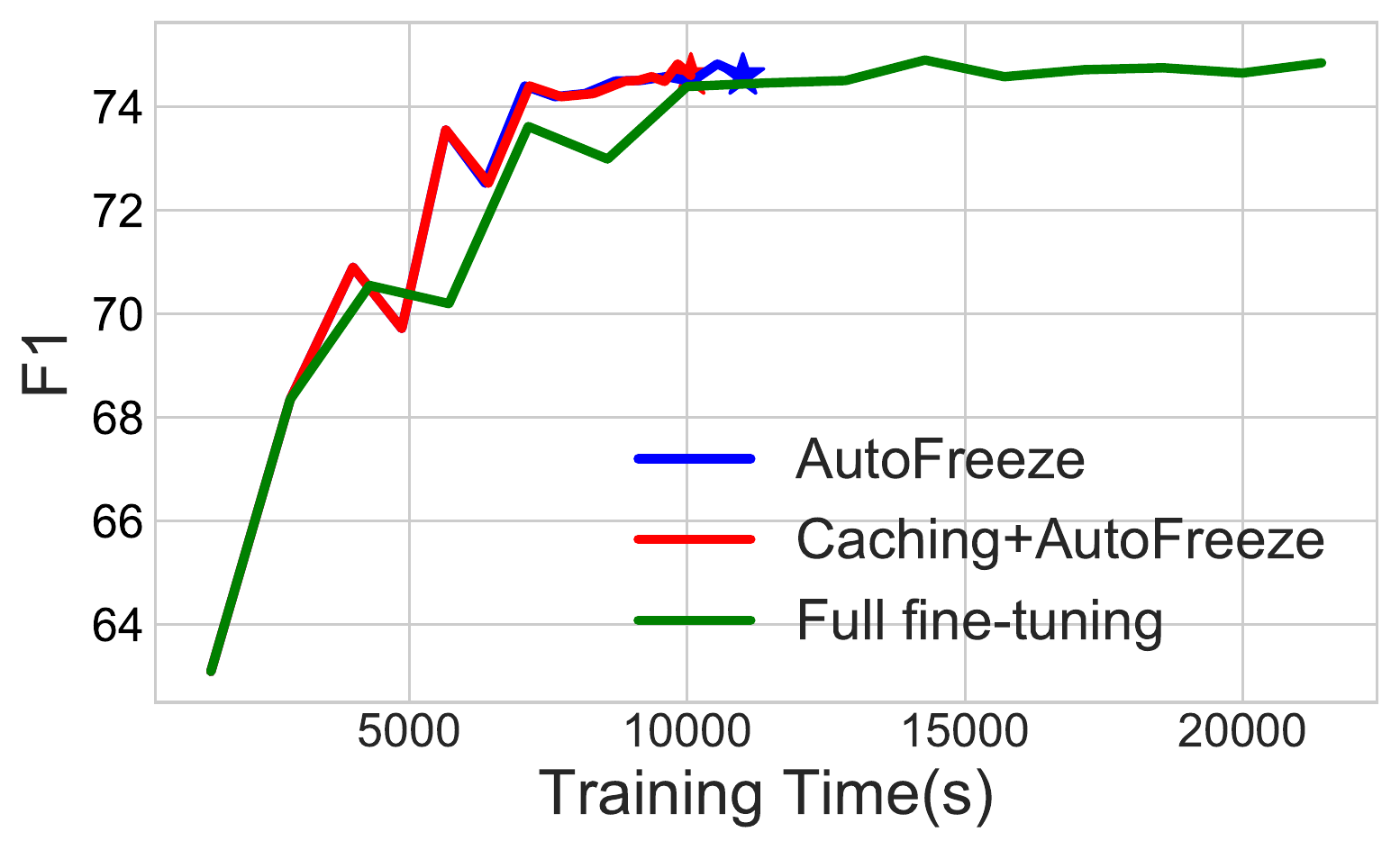}
    \end{subfigure}
    \begin{subfigure}[b]{0.3\textwidth}
    \includegraphics[width=\textwidth]{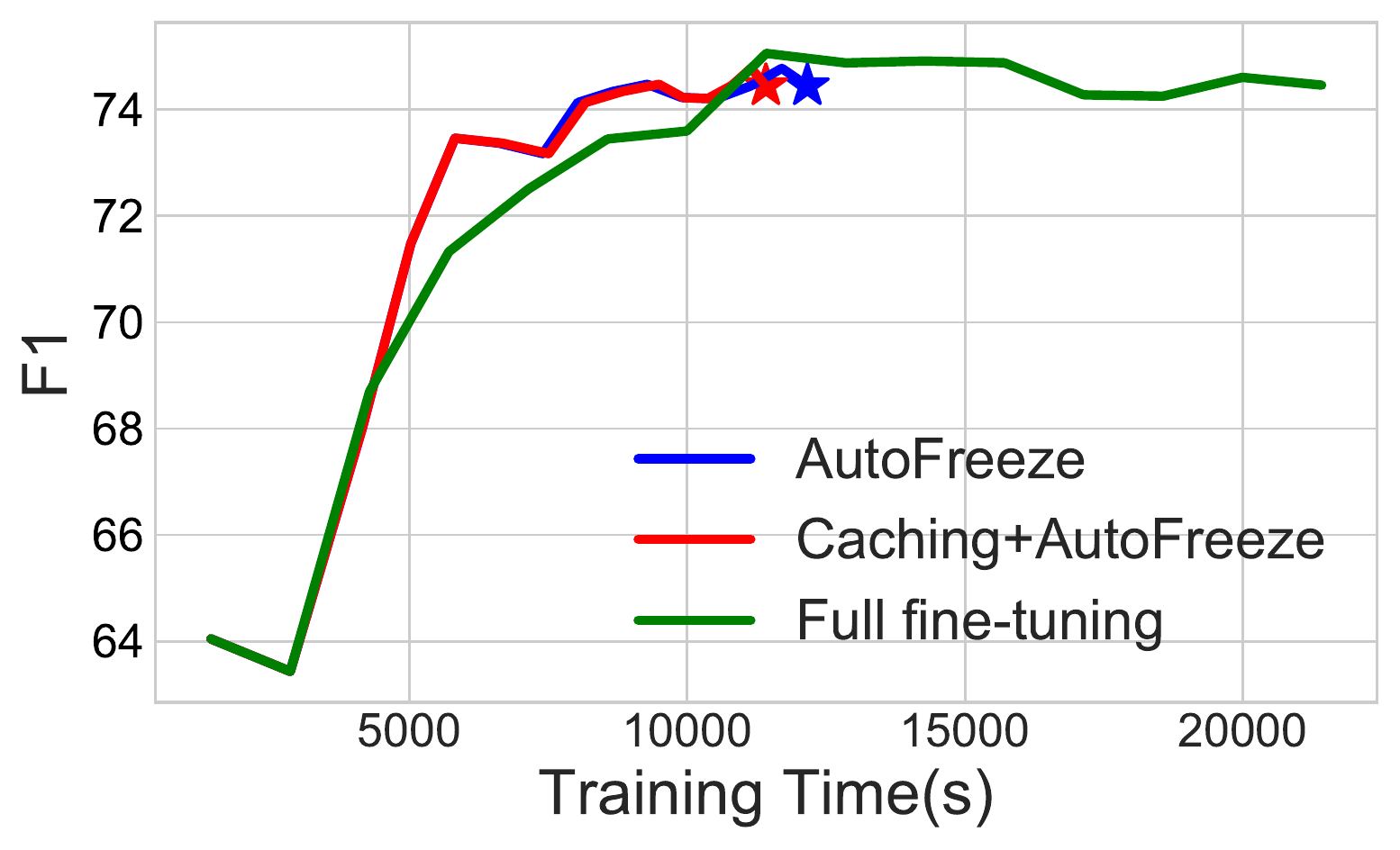}
    \end{subfigure}
        \begin{subfigure}[b]{0.3\textwidth}
    \includegraphics[width=\textwidth]{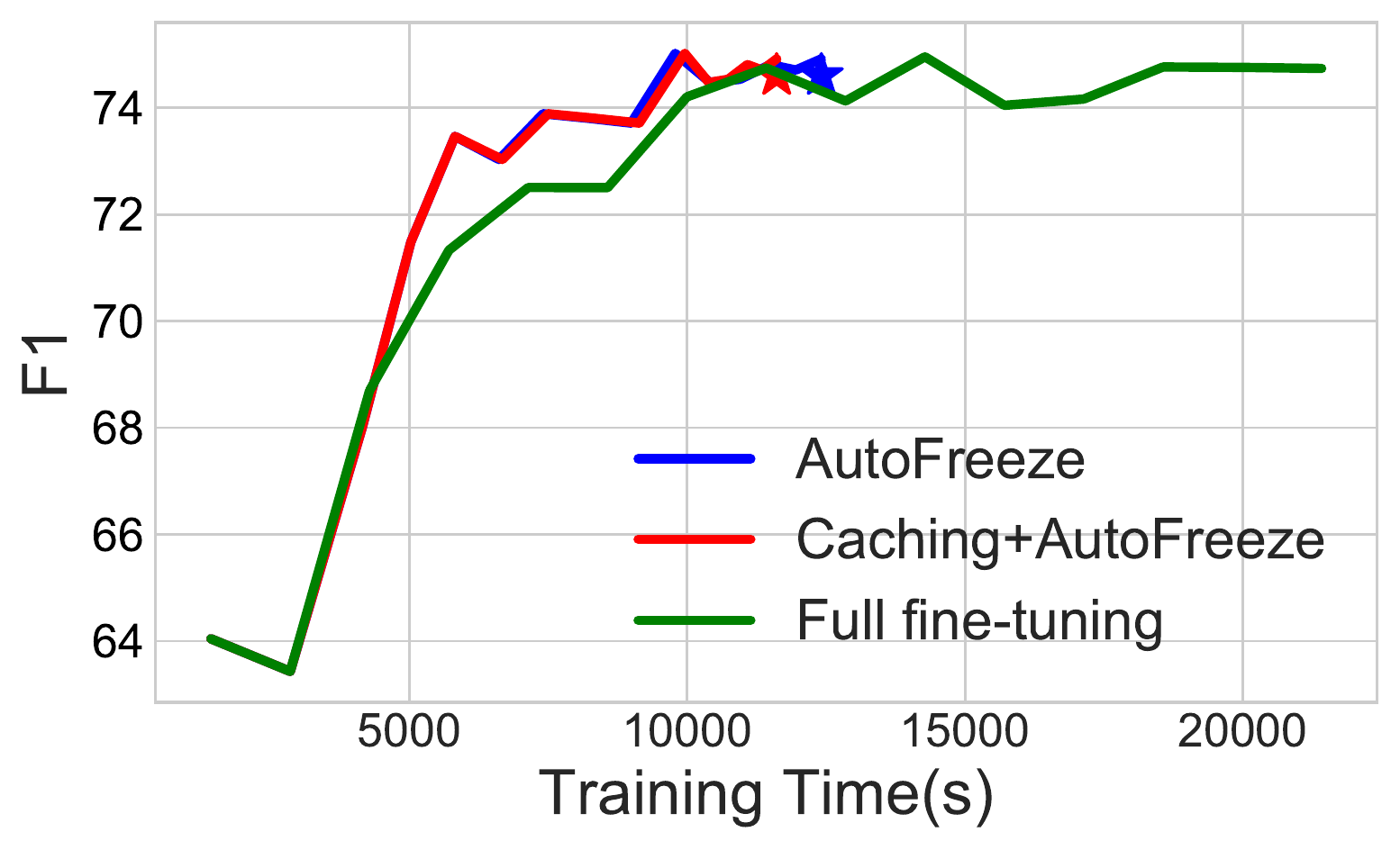}
    \end{subfigure}
    \caption{[\textbf{SQUAD2.0}] Dev F1 curve for each trial with respect to end-to-end training time for \sys{}, \sys{} with Caching turned on, and full fine-tuning. }
    \label{fig:squad_curve}
\end{figure*}

\begin{figure*}[ht]
    \centering
    \begin{subfigure}[b]{0.3\textwidth}
    \includegraphics[width=\textwidth]{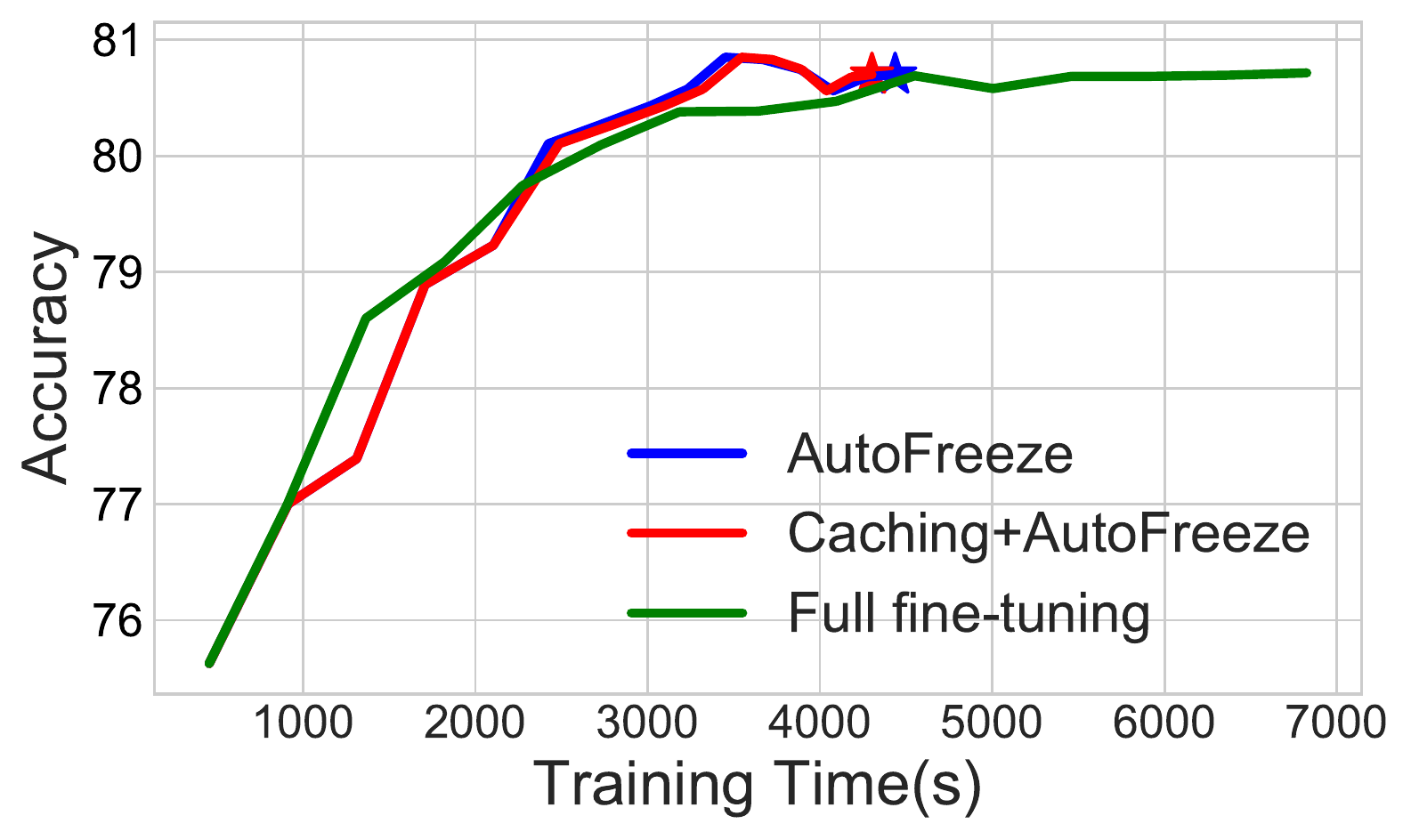}
    \end{subfigure}
    \begin{subfigure}[b]{0.3\textwidth}
    \includegraphics[width=\textwidth]{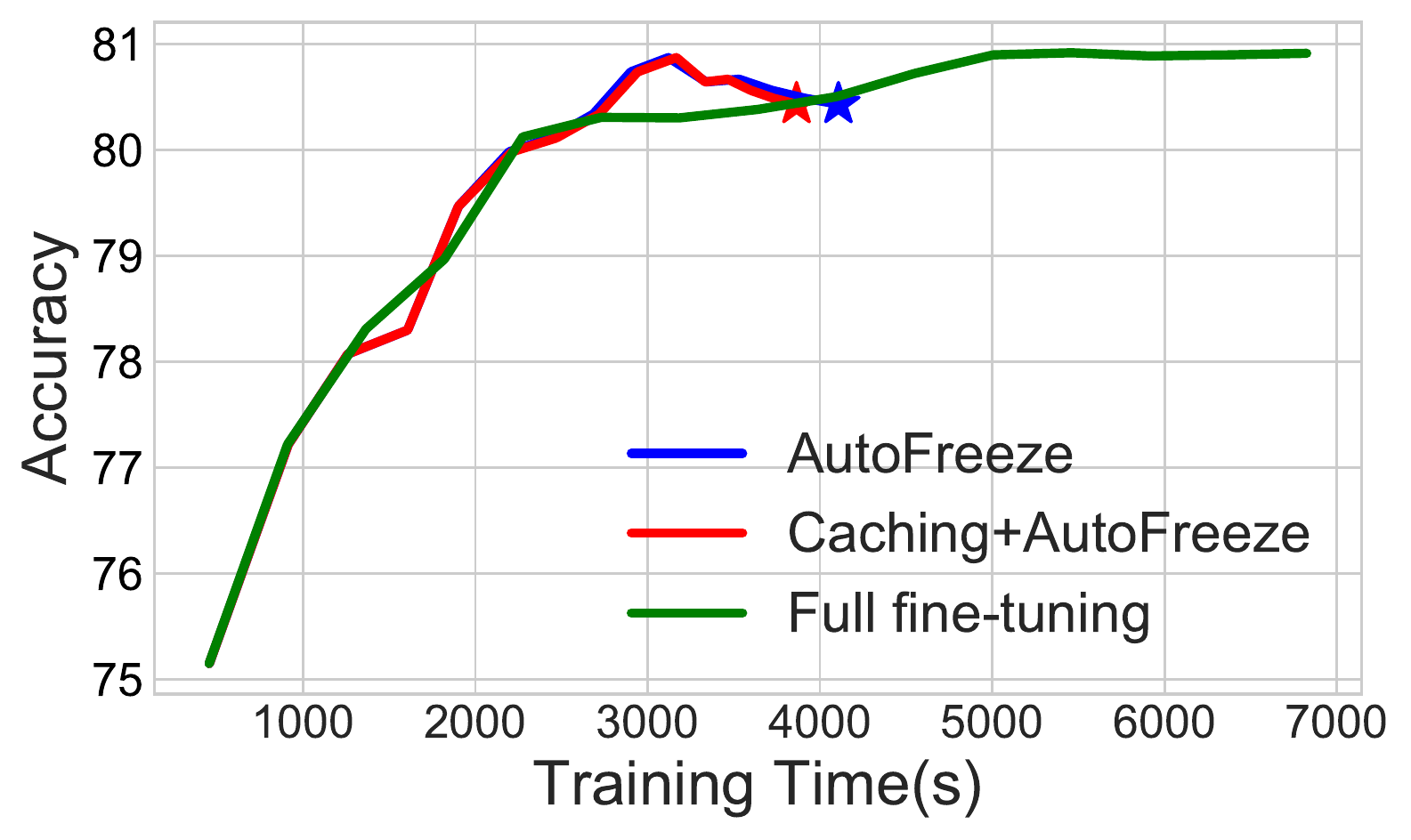}
    \end{subfigure}
        \begin{subfigure}[b]{0.3\textwidth}
    \includegraphics[width=\textwidth]{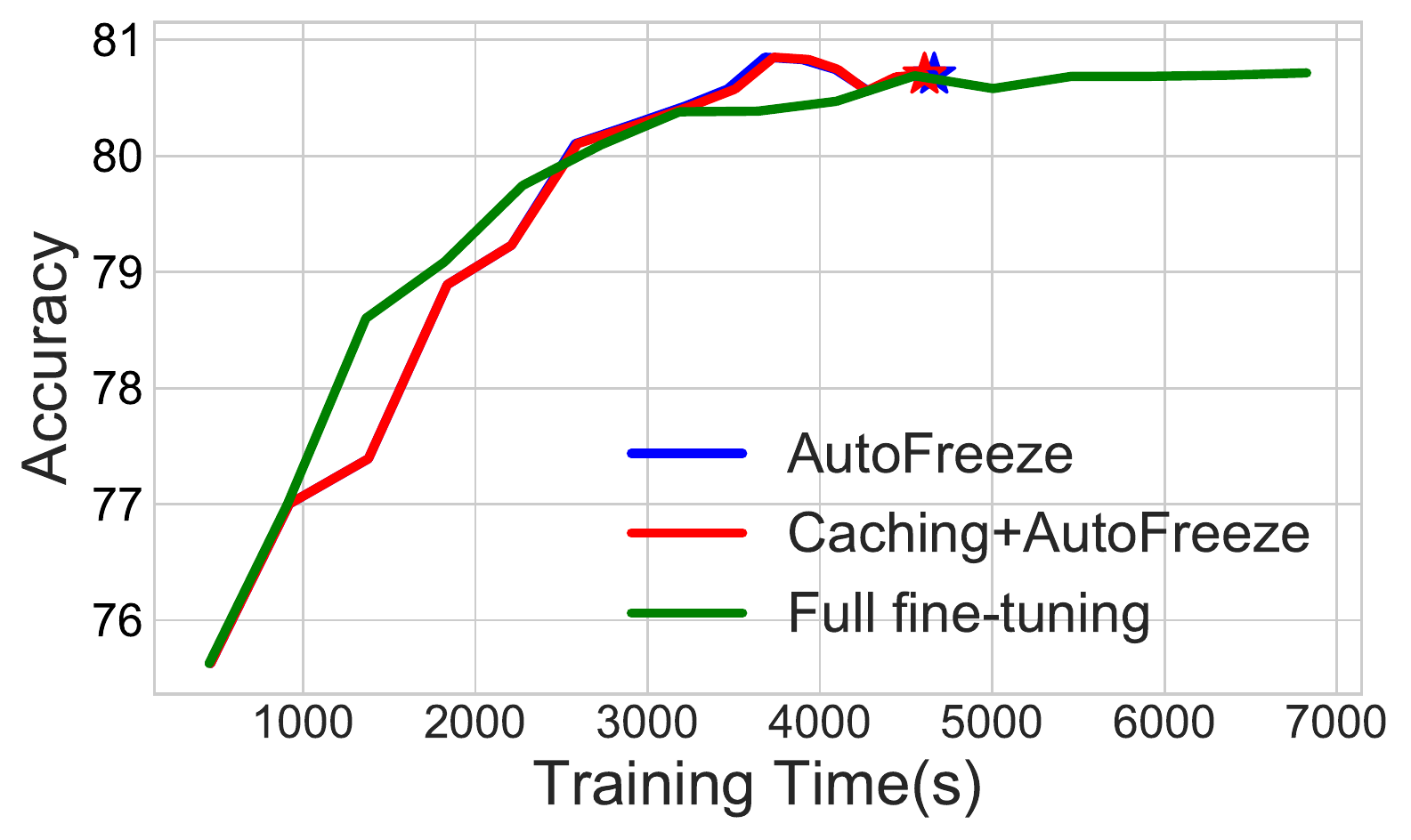}
    \end{subfigure}
    \caption{[\textbf{SWAG}] Dev accuracy curve for each trial with respect to end-to-end training time for \sys{}, \sys{} with Caching turned on, and full fine-tuning. }
    \label{fig:swag_curve}
\end{figure*}

We also include the test accuracy convergence curve with respect to time for each of the three repeated runs using stepped learning rate schedule for each dataset in Figures~\ref{fig:ag_curve1}, \ref{fig:sogou_curve1}, \ref{fig:imdb_curve1}, \ref{fig:yelp_curve}, \ref{fig:squad_curve}, and \ref{fig:swag_curve}.
We see that \sys{} and full fine-tuning achieve comparable max accuracy with an average end-to-end training speedup of 2.05$\times$, 1.55$\times$, 2.05$\times$, 1.94$\times$, 1.81$\times$, and 1.56$\times$ for AG News, Sogou News, IMDb, Yelp F., SQuAD2.0 and SWAG respectively. We can also see that the freezing speedup is on the same scale across different runs. We gain 2.55$\times$, 1.66$\times$, 2.55$\times$, 2.15$\times$, 1.95$\times$, and 1.64$\times$ more speedup on average when turning on the storage manager for AG News, Sogou News, IMDb, and Yelp F., SQuAD2.0 and SWAG respectively compared to full fine-tuning. As shown in Figure \ref{fig:sogou_curve1}(b), we do not have significant improvements when turning on the storage manager as \sys{} decides to start freezing layers from the third epoch for this set of experiments. Accordingly, we are only able to achieve speedup gains from the last epoch through caching.

\begin{figure*}[ht]
    \centering
    \begin{subfigure}[b]{0.3\textwidth}
    \includegraphics[width=\textwidth]{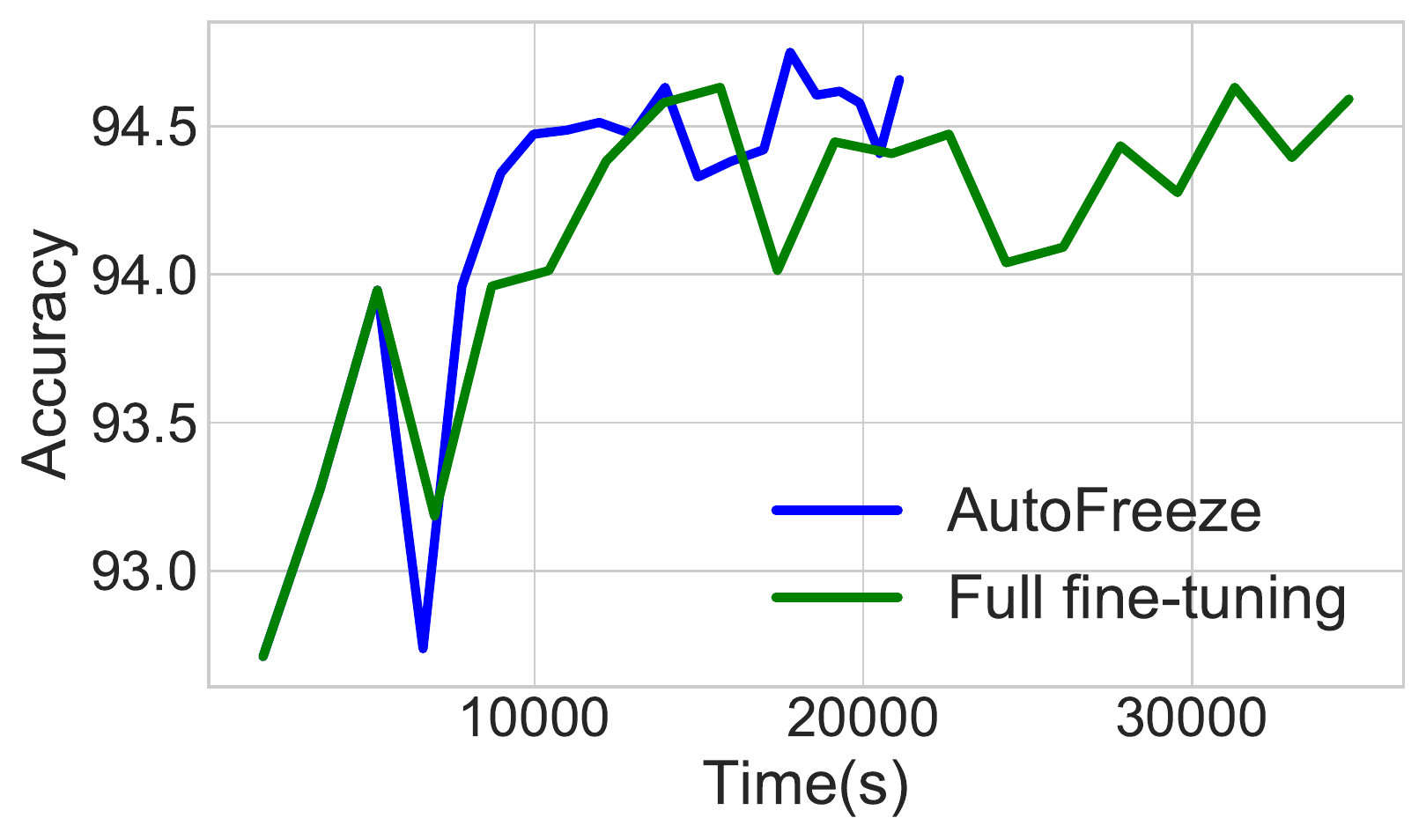}
    \subcaption{AG News}
    \label{fig:ag_constant}
    \end{subfigure}
    \begin{subfigure}[b]{0.3\textwidth}
    \includegraphics[width=\textwidth]{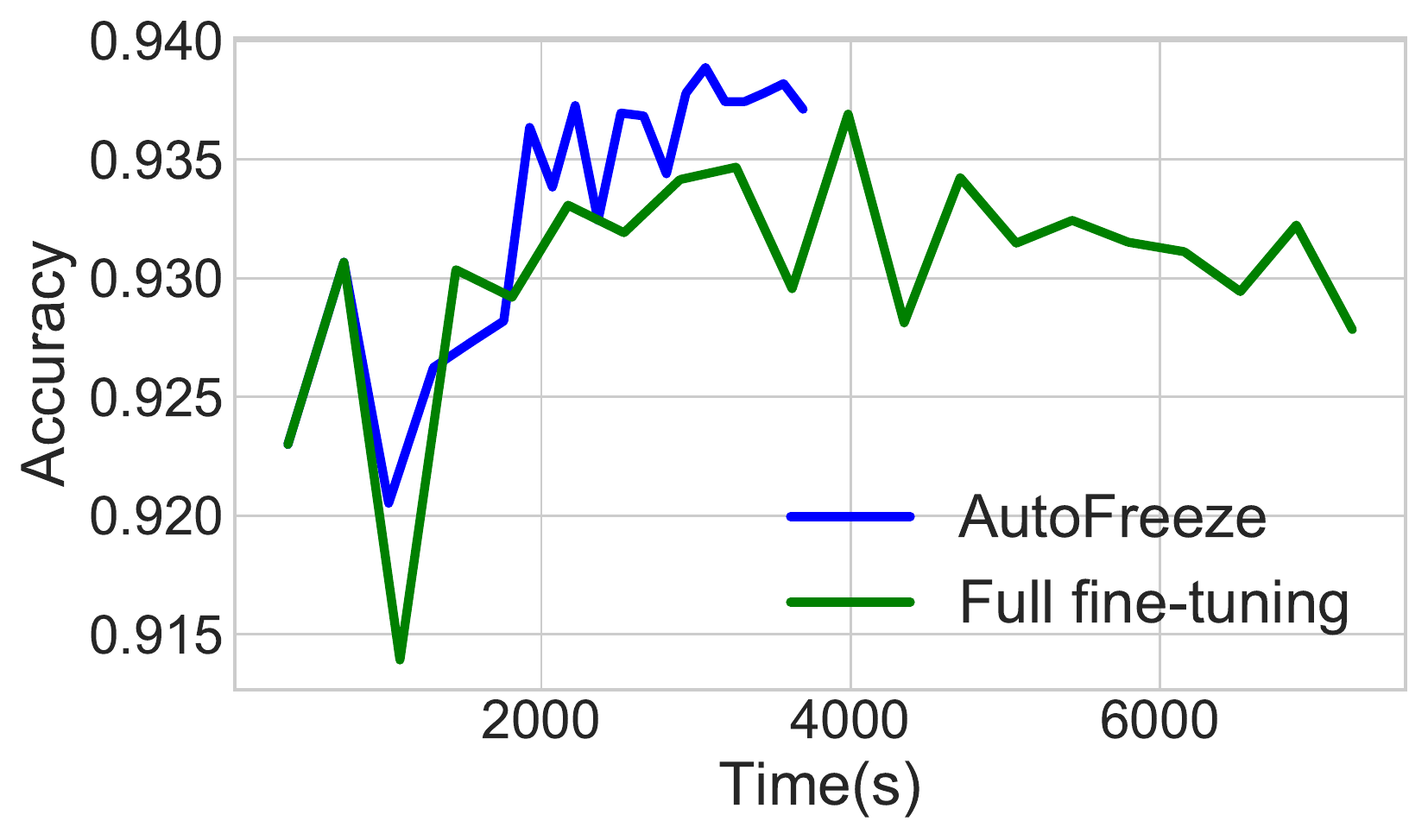}
    \subcaption{IMDb}
    \label{fig:imdb_constant}
    \end{subfigure}
    \begin{subfigure}[b]{0.3\textwidth}
    \includegraphics[width=\textwidth]{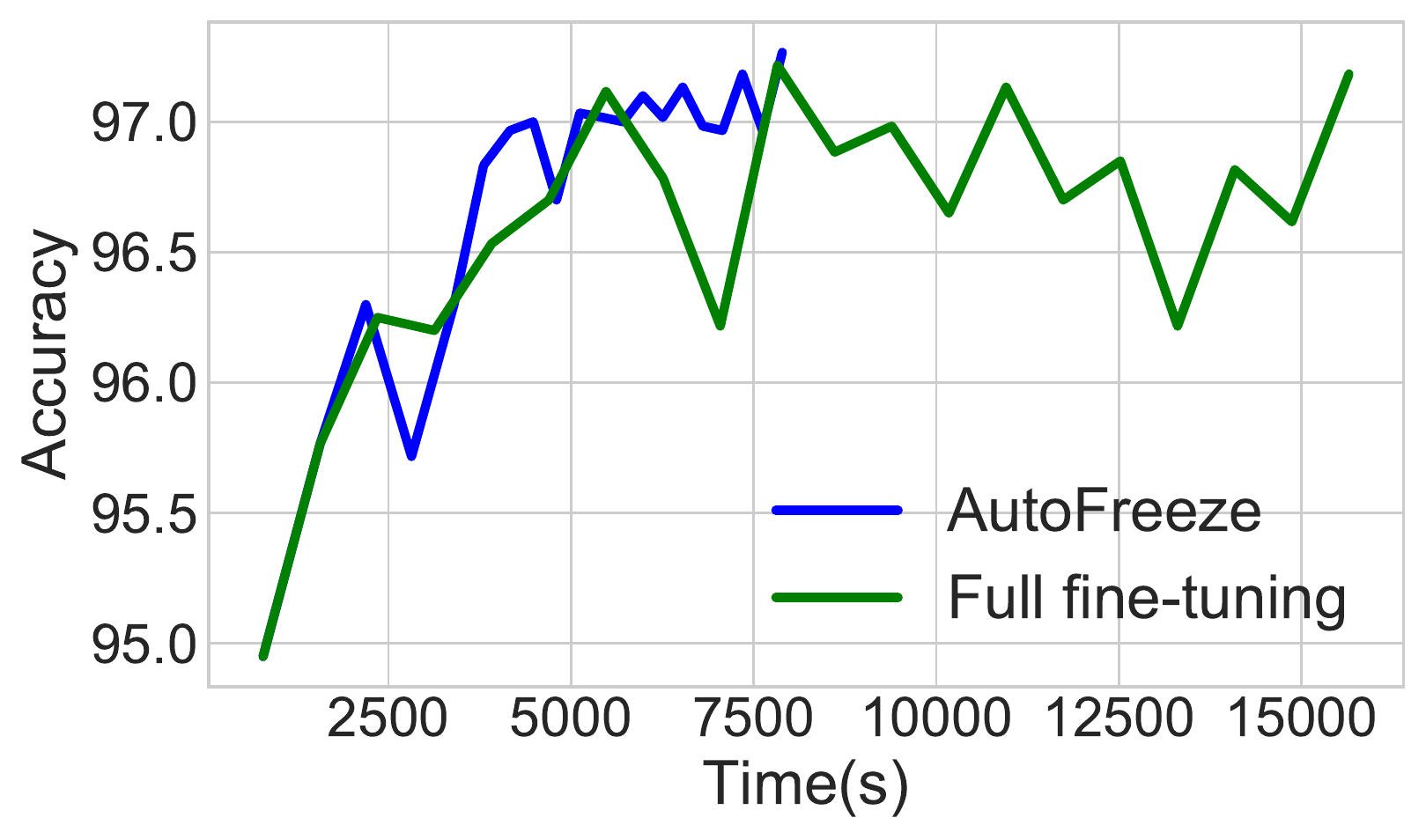}
    \subcaption{Sogou}
    \label{fig:sogou_contant}
    \end{subfigure}
    \caption{Test accuracy curve with respect to end-to-end training time for \sys{} and full fine-tuning when using \textbf{constant learning rate} schedule.
    }
\end{figure*}

\subsection{Constant learning rate schedule}
\label{sec:appendix_constant}
To show that \sys{} is effective for other learning rate schedules, we run \sys{} using constant learning rate schedule with learning rate of 1e-5. As shown in Figure \ref{fig:ag_constant}, \ref{fig:imdb_constant} and \ref{fig:sogou_contant}, \sys{} is able to achieve 1.65$\times$, 1.96$\times$, and 1.98$\times$ speedup for AG News, IMDb, and Sogou News with respect to end-to-end fine-tuning time without harming the model accuracy.

\subsection{Caching benefits for longer runs}
\label{subsec:caching_appendix}
\begin{figure}[!htb]
    \centering
    \includegraphics[width=0.4\textwidth]{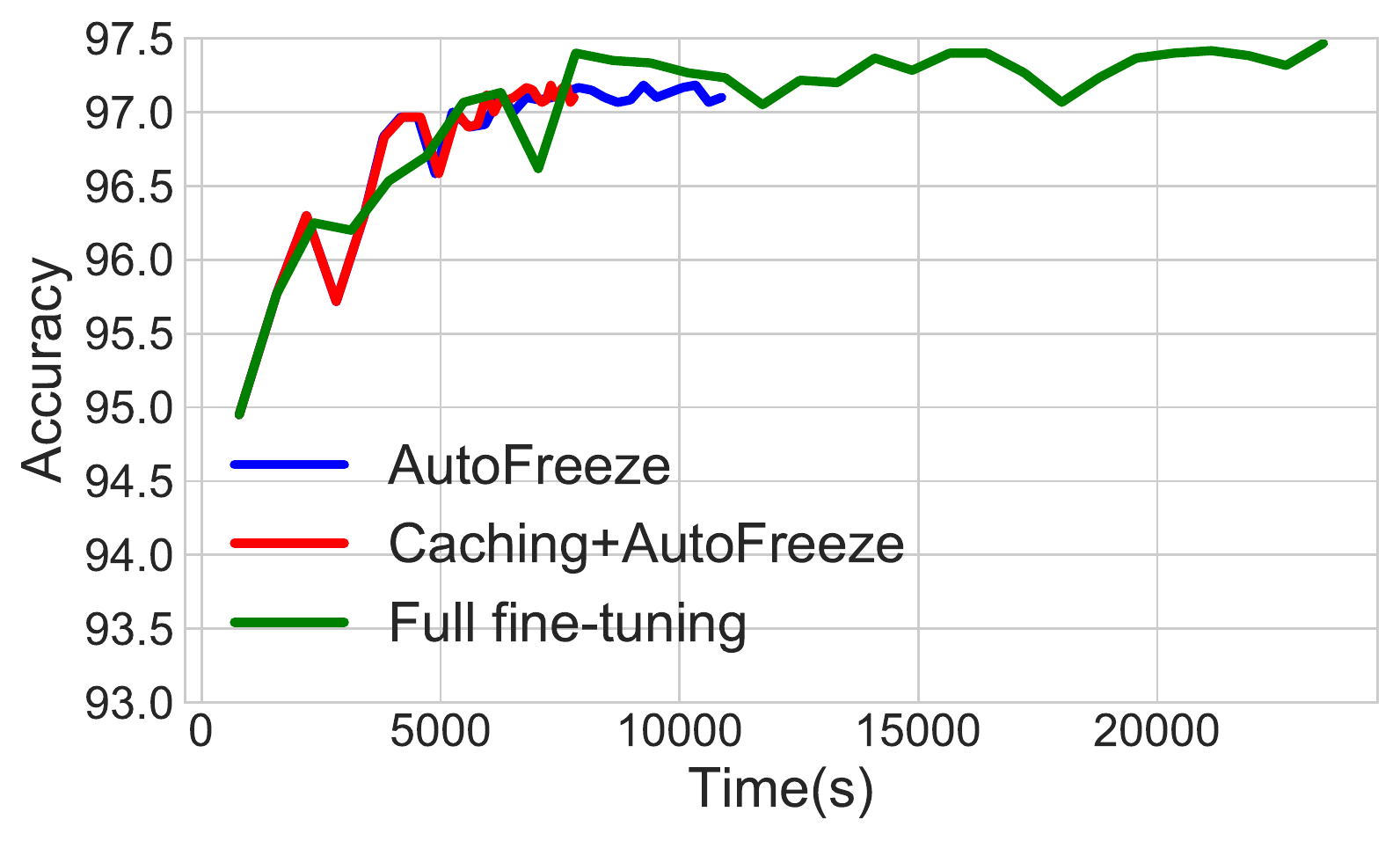}
        \caption{Benefits from \sys{} when the \textbf{Sogou} dataset is fine-tuned for more epochs (6 epochs). We see that caching can provide more benefits in this case.
        }
    \label{fig:long_epc}
\end{figure}
We next present the benefits brought by caching when we fine-tune ${\rm BERT_{BASE}}$ for more than four epochs. We fine-tune ${\rm BERT_{BASE}}$ on Sogou dataset for 6 epochs using stepped learning rate schedule with initial learning rate of 1e-5. As shown in Figure \ref{fig:long_epc}, \sys{} is able to achieve 2.16$\times$ speedup compared to the full fine-tuning. When the storage manager is turned on, we get 3.01$\times$ speedup compared with the baseline. The storage manager is able to get more significant speedup in this setup as \sys{} decides to freeze up to layer 9 at the end of the second epoch, thus saving most of the forward computation for future epochs. In general, we can save more forward computation time when we run the fine-tuning procedure for longer epochs. However, for the datasets we consider in this paper we use a maximum of four epochs. This is because, as reported in prior work~\cite{sun2019fine}, using more epochs doesn't lead to significant improvements in accuracy for these datasets.

\begin{table*}[!t]
\begin{small}

\begin{tabular}{@{}llllllll@{}}
\toprule
\multirow{2}{*}{\textbf{Dataset}}    & \multicolumn{3}{l}{\hfil \textbf{AutoFreeze}}                                                      & \multicolumn{3}{l}{\hfil \textbf{Full fine-tuning}}                                                & \multirow{2}{*}{\textbf{\begin{tabular}[c]{@{}l@{}}Training \\ Speedup\end{tabular}}} \\
                                     & \begin{tabular}[c]{@{}l@{}}Best \\ Iteration\end{tabular} & Accuracy & Training Time(s) & \begin{tabular}[c]{@{}l@{}}Best \\ Iteration\end{tabular} & Accuracy & Training Time(s) &                                                                                       \\ \midrule
\multirow{3}{*}{\textbf{AG News}}    & 80000                                                     & 94.66         & 18993            & 40000                                                     & 94.59         & 34559            & 1.82$\times$                                                                          \\
                                     & 28000                                                     & 94.68         & 15936            & 52000                                                     & 94.66         & 35114            & 2.20$\times$                                                                          \\
                                     & 80000                                                     & 94.66         & 16242            & 36000                                                     & 94.70         & 35058            & 2.16$\times$                                                                          \\ \midrule
\multirow{3}{*}{\textbf{Sogou News}} & 21600                                                     & 97.45         & 10795            & 28800                                                     & 97.38         & 15552            & 1.44$\times$                                                                          \\
                                     & 30600                                                     & 97.12         & 9462             & 28800                                                     & 97.32         & 15527            & 1.64$\times$                                                                          \\
                                     & 28800                                                     & 97.4          & 9866             & 28800                                                     & 97.48         & 15478            & 1.57$\times$                                                                          \\ \midrule
\multirow{3}{*}{\textbf{Yelp F.}}    & 389988                                                    & 68.96         & 97368            & 324990                                                    & 68.83         & 188892           & 1.94$\times$                                                                          \\
                                     & 389988                                                    & 68.63         & 102859           & 194994                                                    & 68.44         & 189207           & 1.84$\times$                                                                          \\
                                     & 303324                                                    & 68.94         & 92226            & 281658                                                    & 68.91         & 188957           & 2.05$\times$                                                                          \\ \midrule
\multirow{3}{*}{\textbf{IMDb}}       & 9163                                                      & 93.94         & 3543             & 4165                                                      & 93.944        & 7304             & 2.06$\times$                                                                          \\
                                     & 10829                                                     & 94.024        & 3584             & 4165                                                      & 93.944        & 7267             & 2.03$\times$                                                                          \\
                                     & 15827                                                     & 93.604        & 3512             & 8330                                                      & 93.98         & 7253             & 2.07$\times$                                                                          \\ \bottomrule
\end{tabular}
\caption{\small{\textbf{AutoFreeze Performance Evaluation (Classification tasks):} We report performance of \sys{} on 4 different datasets. Each experiment is repeated 3 times with different random seeds. We observes \sys{} leads to upto $2\times$ reduction in fine tuning time while reaching same accuracy as full fine tuning. }}
\label{tab:accuracy_num}
\end{small}
\end{table*}

\begin{table*}[]
\begin{small}
\begin{tabular}{@{}llllllll@{}}
\toprule
\hfil Dataset  & \multicolumn{3}{l}{\hfil \textbf{AutoFreeze}  }     & \multicolumn{3}{l}{\hfil \textbf{Full fine-tuning}}        & \multirow{2}{*}{\begin{tabular}[c]{@{}l@{}}Training \\ Speedup\end{tabular}} \\
                            & Best Iteration & Dev F1 & Training Time (s) & Best Iteration & Dev F1 & Training Time (s) &                                                                              \\ \midrule
\multirow{3}{*}{\textbf{SQUAD2.0}} & \hfil 42881          & \hfil 74.83  & \hfil 11002     & \hfil 29687    & \hfil 74.90  & \hfil 21419             & \hfil 1.94x                                                                        \\
                            & \hfil 29687          & \hfil 75.02  & \hfil 12163             & \hfil 29687          & \hfil 74.95  & \hfil 21532             & \hfil 1.77x                                                                        \\
                            & \hfil 42881          & \hfil 74.78  & \hfil 12414             & \hfil 23090          & \hfil 75.05  & \hfil 21512             & \hfil 1.73x                                                                        \\ \bottomrule
\end{tabular}
\caption{\textbf{AutoFreeze Performance Evaluation (Question Answering): }We report performance of AutoFreeze on SQuAD2.0. The experiment is repeated 3 times with different random seeds. }
\label{tab:squad_summ}
\end{small}
\end{table*}

\begin{table*}[]
\centering
\begin{tabular}{@{}llllllll@{}}

\toprule
\multirow{2}{*}{Dataset} & \multicolumn{3}{l}{\hfil AutoFreeze}                                                                                                                                              & \multicolumn{3}{l}{\hfil Full fine-tuning}                                                                                                                                         & \multirow{2}{*}{\begin{tabular}[c]{@{}l@{}}Training \\ Speedup\end{tabular}} \\
                         & \begin{tabular}[c]{@{}l@{}}Best\\ Iteration\end{tabular} & \begin{tabular}[c]{@{}l@{}}Dev \\ F1\end{tabular} & \begin{tabular}[c]{@{}l@{}}Training \\ Time (s)\end{tabular} & \begin{tabular}[c]{@{}l@{}}Best \\ Iteration\end{tabular} & \begin{tabular}[c]{@{}l@{}}Dev \\ F1\end{tabular} & \begin{tabular}[c]{@{}l@{}}Training \\ Time (s)\end{tabular} &                                                                              \\ \midrule
\multirow{3}{*}{SWAG}    & 4138                                                     & 80.85                                             & 4436                                                         & 6437                                                      & 80.72                                             & 6868                                                         & 1.55x                                                                        \\
                         & 6437                                                     & 81.02                                             & 4663                                                         & 6437                                                      & 80.89                                             & 6848                                                         & 1.47x                                                                        \\
                         & 4138                                                     & 80.88                                             & 4107                                                         & 5057                                                      & 80.92                                             & 6857                                                         & 1.66x                                                                        \\ \bottomrule
\end{tabular}

\caption{\textbf{AutoFreeze Performance Evaluation (Multiple Choice): }We report performance of AutoFreeze on the SWAG dataset. Each experiment is repeated 3 times with different random seeds. }
\label{tab:swag_summ}
\end{table*}

\clearpage

\clearpage

\end{document}